\documentclass[12pt,draftcls,onecolumn]{IEEEtran}
\usepackage{cite,subfig}

\ifCLASSINFOpdf
  \usepackage[pdftex]{graphicx}
  \graphicspath{{./figs/}}
  \DeclareGraphicsExtensions{.pdf}
\else
  \usepackage[dvips]{graphicx}
  \graphicspath{{./figs/}}
  \DeclareGraphicsExtensions{.eps}
\fi

\usepackage[cmex10]{amsmath}
{
      \newtheorem{assumption}{Assumption}
  }
\usepackage{amssymb,tabularx,multirow}
\usepackage{algorithm,algorithmic}
\usepackage{array}
\usepackage{subfloat,adjustbox}

\usepackage{fixltx2e,rotating,dblfloatfix}

\usepackage{hyperref}
\allowdisplaybreaks

\begin{document}
\title{Simultaneously sparse and low-rank abundance matrix estimation for hyperspectral image unmixing}
\author{Paris~V.~Giampouras,
	Konstantinos~E.~Themelis,
        Athanasios~A.~Rontogiannis,~\IEEEmembership{Member,~IEEE,}
        and~Konstantinos~D.~Koutroumbas%
\thanks{P. V. Giampouras, K. E. Themelis, A. A. Rontogiannis and K. D. Koutroumbas are with the Institute for Astronomy, Astrophysics, Space Applications and Remote Sensing (IAASARS), National Observatory of Athens, I. Metaxa \& Vas. Pavlou str., GR-15236, Penteli, Greece (e-mail: parisg@noa.gr; themelis@noa.gr; tronto@noa.gr; koutroum@noa.gr).}
}




\maketitle

\begin{abstract}
In a plethora of applications dealing with inverse problems, e.g. in image processing, social networks, compressive sensing, biological data processing etc., the signal of interest is known to be structured in several ways at the same time. This premise has recently guided the research to the innovative and meaningful idea of imposing  multiple constraints on the unknown parameters involved in the problem under study. For instance, when dealing with problems whose unknown parameters form sparse and low-rank matrices, the adoption of suitably combined constraints imposing sparsity and low-rankness, is expected to yield substantially enhanced estimation results. In this paper, we address the spectral unmixing problem in hyperspectral images. Specifically, two novel unmixing algorithms are introduced, in an attempt to exploit both spatial correlation and sparse representation of pixels lying in homogeneous regions of hyperspectral images. To this end, a novel mixed penalty term is first defined consisting of the sum of the weighted $\ell_1$ and the weighted nuclear norm of the abundance matrix corresponding to a small area of the image determined by a sliding square window. This penalty term is then used to regularize a conventional quadratic cost function and impose simultaneously sparsity and row-rankness on the abundance matrix. The resulting regularized cost function is minimized by a) an {\it incremental proximal sparse and low-rank} unmixing algorithm and b) an algorithm based on the {\it alternating minimization method of multipliers} (ADMM). The effectiveness of the proposed algorithms is illustrated in experiments conducted both on simulated and real data.
\end{abstract}

\begin{IEEEkeywords}
Semi-supervised spectral unmixing, hyperspectral images, simultaneously sparse and low-rank matrices, proximal methods, alternating direction method of multipliers (ADMM), abundance estimation 
\end{IEEEkeywords}
\ifCLASSOPTIONpeerreview
\begin{center} \bfseries EDICS Category: \end{center}
\IEEEpeerreviewmaketitle
\fi


\section{Introduction}
\label{sec:intro}
Spectral unmixing (SU) of hyperspectral images (HSIs) has attracted considerable attention in recent years both in research and applications. SU can be considered as the process of a) identifying the spectral signatures of the materials ({\it endmembers}) whose mixing generates the (so called) {\it mixed} pixels of an HSI and b) deriving their corresponding fractions ({\it abundances}) in the formation of each HSI pixel, \cite{ma2013signal}. The latter constitute the so called {\it abundance vector} of the pixel. This two step procedure has given rise to a plethora of methods tackling either one or both these two tasks. Diverse statistical and geometrical approaches have been lately put forward in literature addressing the first step, commonly known as endmembers' extraction (e.g. \cite{nascimento2005vertex,li2008minimum}). On the other hand, there have been many research works that assume that the spectral signatures of the endmembers are available and focus on the abundance estimation task. Algorithms that fall into this class, need to make a fundamental assumption concerning the inherent mixing process that generates the spectral signatures of the HSI pixels. 

In view of the latter, the linear mixing model (LMM) holds a dominant position being widely adopted in numerous state-of-the-art unmixing algorithms (see e.g. \cite{ma2013signal} and the references therein). More specifically, these algorithms are based on the premise that the pixels' spectral signatures are generated by a linear combination of  endmembers' spectra contained in a predefined set, usually termed as {\it endmembers' dictionary}. Abundance estimation is henceforth treated as a linear regression problem. LMM has prevailed over other models, due to its simplicity and mathematical tractability. Physical considerations that naturally arise, impose various constraints on the unmixing problem. In this context, the so-called {\it abundance nonnegativity} and  the {\it abundance sum-to-one} constraints are usually adopted. That said, unmixing can be viewed as a constrained linear regression problem.

In an attempt to achieve better abundance estimation results, recent novel ideas promote the incorporation of further prior knowledge in the unmixing problem. In light of this, several methods bring into play the {\it sparsity} assumption, \cite{themelis2010semi,bioucas2010alternating,iordache2011sparse,themelis2012novel}. Its adoption is justified by the fact that only a few of the available endmembers participate in the formation of a given pixel, especially in the case of large size endmembers' dictionaries. Put it in other terms, it is envisaged that pixels' spectral signatures accept sparse representations with respect to a given endmembers' dictionary. Furthermore, one could also say that the abundance vectors corresponding to the pixels of HSIs are deemed having only a few non-zero values. Practically speaking, sparsity is imposed on abundances by means of $\ell_1$ norm regularization, \cite{themelis2010semi,bioucas2010alternating,iordache2011sparse} when a deterministic approach is followed. On the other hand, in Bayesian schemes appropriate sparsity inducing priors are adopted for the abundance vectors, \cite{themelis2012novel,themelis2013fast}. {\it Spatial correlation} is another constraint that has recently been incorporated in the unmixing process, offering stimulating results, \cite{eches2011enhancing,quabundance,iordache2014collaborative}. In that vein, the additional information that exists in homogeneous regions of HSIs is subject to exploitation. Actually, in such regions, there is a high degree of correlation among the spectral signatures of neighboring pixels. It is hence anticipated that there should also be correlation among the abundance vectors corresponding to these pixels. This has led to the development of novel unmixing schemes, whereby the information provided by the neighboring pixels is taken into account in the  abundance estimation of each single pixel. 

In this spirit, a collaborative deterministic scheme, termed CLSUnSAL, was recently proposed in \cite{iordache2014collaborative}, which uses a wealth of information stemming from all the pixels of the examined HSI. CLSUnSAL adopts dictionaries consisting of a large amount of endmembers. Then it assumes that spatial correlation translates into abundance vectors sharing the same support set i.e., presenting a similar sparsity pattern. Thus, the matrix whose columns are the abundance vectors of all HSI pixels (called {\it abundance matrix}) should meaningfully be of a joint-sparse structure \footnote{A joint-sparse Bayesian unmixing scheme has been also presented in \cite{whispers2014}.}. To impose joint-sparsity, CLSUnSAL applies a $\ell_{2,1}$ norm on the sought abundance matrix, which is then used to penalize a suitably defined quadratic cost function. Minimization of the resulting regularized cost function is performed by an alternating direction method of multipliers (ADMM), \cite{boyd2011distributed}. A similar perspective is followed in \cite{quabundance}, however in a ``localized'' fashion. Specifically, \cite{quabundance} proposes the use of a $3\times3$ square window that slides all over the image. The abundance vector of the central pixel is then inferred, by taking into account the spectral signatures of the adjacent pixels contained in the window. Based on this idea, two algorithms are derived. First the MMV-ADMM, which in a similar to CLSUnSAL fashion, seeks joint-sparse abundance matrices utilizing the $\ell_{2,1}$ norm, and second the LRR algorithm that promotes a low-rank structure on the abundance matrix. Actually, the LRR algorithm presents an alternative way of modelling the spatial correlation among neighboring pixels. That is, it assumes that the correlation among pixels' spectral signatures is reflected as linear dependence among their corresponding abundance vectors. Apparently, the matrix formed by these abundance vectors should be of low rank. That said, a nuclear norm is imposed on the abundance matrix, and a properly adapted augmented Lagrangian cost function is minimized in an alternating minimization fashion.

In this paper, we introduce a novel idea for performing abundance estimation in HSIs under the LMM, that simultaneously takes spatial correlation and sparsity into consideration. Similarly to \cite{quabundance}, we utilize a $\kappa\times \kappa$ square sliding window with $\kappa$ odd, and we consider the spectral signatures of adjacent pixels lying in it. Departing from the usual paradigm, we propose to seek for $\kappa^2$-column abundance matrices that are {\it simultaneously sparse and low-rank}. SU is thus formulated  as a {\it sparse reduced-rank regression} problem, \cite{chen2012sparse}. As stated earlier, low-rankness arises naturally in abundance matrices corresponding to relatively homogeneous regions, due to the linear dependence of the respective abundance vectors. At the same time, sparsity is a reasonable hypothesis that still holds independently, as explained above, within each individual abundance vector. Broadly speaking, imposing  multiple structures on the same mathematical object when dealing with inverse problems is a strategy still in its very infancy in signal processing and machine learning literature, \cite{savalle2012estimation,oymak2012simultaneously,golbabaee2012compressed,richard2014}. The aforementioned sparsity and low-rank constraints give rise to a mixed penalty term that regularizes a least squares fitting function through the weighted $\ell_1$ norm and the weighted trace norm of the abundance matrices, respectively. In order to minimize the cost function, two novel iterative algorithms are proposed, namely the {\it incremental proximal sparse low-rank} unmixing algorithm (IPSpLRU), inspired by \cite{bertsekas2011incremental}, and the {\it alternating direction sparse and low-rank} unmixing algorithm (ADSpLRU). As implied by their names, IPSpLRU is based on proximal operators of the individual terms that compose the cost function. On the other hand, ADSpLRU is an ADMM based approach properly adapted to our problem formulation. The proposed algorithms are compared with state-of-the-art unmixing techniques and their effectiveness is demonstrated via extensive simulated and real-data experiments.

\emph{Notation}: Matrices are represented as boldface uppercase letters, e.g., ${\mathbf X}$, and, column vectors as boldface lowercase letters, e.g., ${\mathbf x}$, while the $i$-th component of vector $\mathbf x$ is denoted by $x_i$ and the $ij$-th element of matrix $\mathbf X$ by $x_{ij}$. Moreover, $^T$ denotes transposition, $\mathbf{I}_N$ is the $N\times N$ identity matrix and $\mathbf{0}$ is a zero matrix with respective dimensions, $\mathbf{1}$ denotes the all ones vector, $\mathrm{rank}(\mathbf{X})$ is the rank of $\mathbf{X}$, $\mathrm{tr}[ \mathbf{X} ]$ denotes the trace of matrix $\mathbf{X}$, $\mathrm{diag}(\mathbf{x})$ is a diagonal matrix with the elements of vector $\mathbf{x}$ on its diagonal, 
$\sigma_i(\mathbf{X})$ is the $i$th largest singular value of $\mathbf{X}$,  $\|\cdot\|_2$ is the standard $\ell_2$ (Euclidean) vector norm, $\|\mathbf{X}\|_{\ast} = \mathrm{Tr}(\sqrt{\mathbf{X}^T \mathbf{X}}) = \sum_{k=1}^{rank(\mathbf{X})}\sigma_i(\mathbf{X}) $, denotes the nuclear norm (or trace norm), $\|\mathbf{X}\|_1 = \sum_i\sum_j |x_{ij}|$ is the sum of the absolute values of all entries of
$\mathbf{X}$ (called the $\ell_1$ norm), $\|\mathbf{X}\|_F = \sqrt{\sum_i\sum_j x_{ij}^2}$, stands for the Frobenius norm. $\mathcal N(\cdot)$ denotes the Gaussian distribution.  
Also, $\mathbb{R}^k$ stands for the $k$-dimensional Euclidean space and $\mathbb{R}_+^k$ denotes the $k$-dimensional non-negative orthant. The matrix inequality   $\mathbf{X}\geq\mathbf{Y}$ declares element-wise operation and $\odot$ stands for component-wise multiplication between matrices of the same size.
\begin{figure}
\centering
\includegraphics[width=0.7\linewidth]{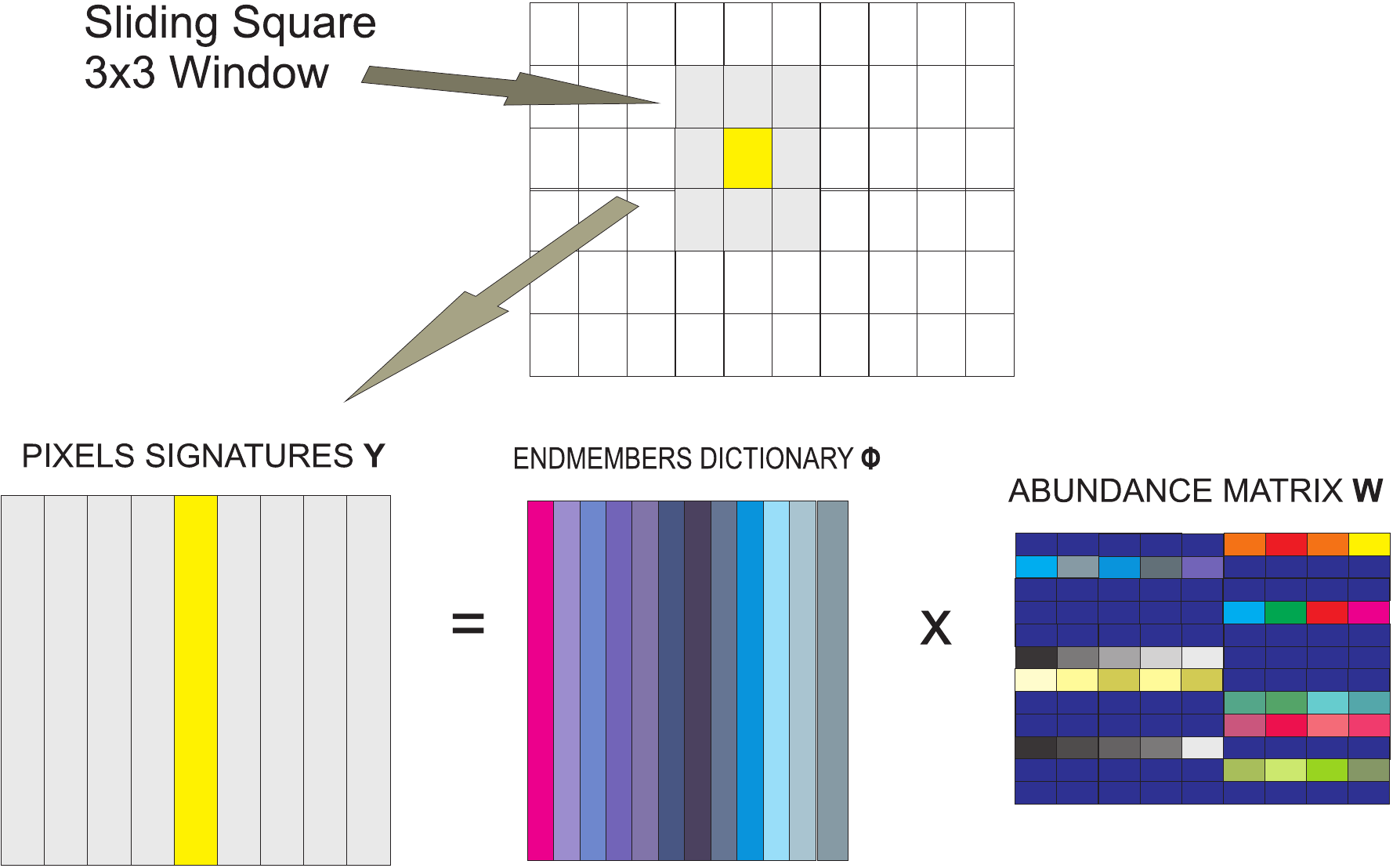}
\caption{Graphical illustration of the sliding window approach of our unmixing algorithms. Abundance matrix is considered sparse and low-rank (in this example rank = 2). Blue cells in matrix $\mathbf{W}$ represent zero values.}
\label{windowing_approach}
\end{figure}

\section{Problem Formulation}


We consider an $L$-spectral band hyperspectral image, with each of its pixels being composed of $N$ endmembers. Let $\boldsymbol \Phi =  [\boldsymbol \phi_1, \boldsymbol \phi_2, \dots, \boldsymbol \phi_N]$ stand for the $L\times N$ endmembers' dictionary, where $\boldsymbol \phi_i \in \mathbb{R}_+^L, i = 1 , 2, \dots, N$, is the spectral signature of the $i$th endmember. 
Consider also a small sliding square window that contains $K$ adjacent pixels ($K = \kappa\times \kappa$), with 
the measurement spectra ${\mathbf y}_k, k = 1, 2, \dots, K$, that are assumed to share the same endmember matrix $\boldsymbol \Phi$, as shown graphically in Fig. \ref{windowing_approach} for $K=9$. In matrix notation, let ${\mathbf Y} = [{\mathbf y}_1, {\mathbf y}_2, \dots, {\mathbf y}_K] $ be the ${L\times K}$ matrix containing the spectra of the $K$ pixels in the window as its columns. Utilizing the linear mixing model (LMM), the mixing process can be described by the equation
\begin{align}
{\mathbf Y} = {\mathbf \Phi}{\mathbf W} + {\mathbf E},
\label{multiplepixelsmodel}
\end{align}
where ${\mathbf W}\in {\mathbb R}_+^{N\times K}$ is the abundance matrix whose columns are the $N$-dimensional abundance vectors of the corresponding $K$ pixels, and $\mathbf{E}\in {\mathbb{R}}^{L\times K}$ is an i.i.d., zero-mean Gaussian noise matrix.
Due to physical considerations, the abundance coefficients in $\mathbf{W}$ should satisfy two constraints, namely, the \emph{abundance nonnegativity} and the \emph{abundance sum-to-one} constraints, \cite{2002_keshava}, i.e., 
\begin{align}
{\mathbf W} \geq \mathbf{0} ,\text{  and   } \mathbf{1}^T\mathbf{W} = \mathbf{1}^T.
\end{align}
Nevertheless, in the following we relax the sum-to-one constraint based on the reasoning presented in \cite{iordache2011sparse}. 
That said, the general problem considered in this paper is the following: ``given the spectral measurements $\mathbf{Y}$ and the endmember matrix $\boldsymbol \Phi$, estimate the abundance matrix $\mathbf{W}$ subject to the nonnegativity constraint''. This is a typical inverse problem, which has been addressed via many methods in the signal processing literature. However, the efficacy of the proposed approach lies on the exploitation of the intrinsic structural characteristics of $\mathbf{W}$, i.e., sparsity and low-rankness. To this end, we impose concurrently two naturally justified {\it structural constraints} on the abundance matrix $\mathbf{W}$, that promote \emph{low-rankness} and \emph{sparsity}.

\emph{Low-rankness property:} A logical consideration is that all pixels belonging to the same window are correlated, i.e., they are composed of the same materials, although maybe in different proportions. This property suggests that the abundance matrix $\mathbf{W}$ to be estimated has linearly dependent columns and thus is either low-rank, or it can be well-approximated by a low-rank  matrix. In the bibliography, low-rank matrix estimation techniques have recently emerged as powerful estimation tools, e.g., \cite{negahban2011, babacaan12, bach2008, quabundance}. These estimators are mainly based on regularization by the nuclear norm of $\mathbf{W}$. A similar regularization is also adopted in this paper in order to impose the low-rank constraint. 


\emph{Sparsity property:} Another typical assumption is that only a small portion of the $N$ endmembers will be present in the spatial area marked by the $\kappa \times \kappa$  shifting window. In other words, it is safe to assume that the abundance matrix $\mathbf{W}$ has a \emph{sparse representation} in terms of the endmember matrix dictionary $\boldsymbol \Phi$. This motivates the use of a sparsity-cognizant estimator for the abundance matrix $\mathbf{W}$, which is envisaged to produce more robust unmixing results. It should be noted that sparsity has already been successfully exploited  in many spectral unmixing algorithms, e.g., \cite{themelis2012novel, themelis2013fast, sunspi, bioucas2010alternating, iordache2011sparse, iordache2014collaborative}.

It is worth mentioning that the sparsity of $\mathbf{W}$ does by no means invalidate its low-rankness. On the contrary, both structural hypotheses on $\mathbf{W}$ are assumed to hold \emph{simultaneously}, although low-rankness implicitly imposes some kind of structure on sparsity. So far, reports in the spectral unmixing literature explore either the sparsity, e.g. \cite{themelis2012novel,sunspi}, or the low-rankness property of $\mathbf{W}$, e.g. \cite{quabundance}. This is the first time, to the best of our knowledge, that spectral unmixing is formulated as a simultaneously sparse and low-rank matrix estimation problem. That is, we seek a matrix $\mathbf{W}\geq \mathbf{0}$ that, apart from fitting the data well in the least squares sense, it has minimum rank and only a few positive elements. 
To achieve this, we define the following optimization problem, 
\begin{align}
(\mathrm{P1}): \ \hat{\mathbf{W}} = \underset{\mathbf{W}\in\mathbb{R}_+^{N\times K}}{\text{argmin  }}  \left \{ \frac{1}{2}\|{\mathbf Y} - \mathbf{\Phi}{\mathbf W}\|_{F}^2 + \gamma \|\mathbf{W}\|_1 + \tau  \|\mathbf{W}\|_{\ast} \right \} 
\label{simultsplr}
\end{align}
where $\gamma,\tau \geq 0 $ are parameters that control the trade-off between the sparsity and rank regularization terms and the data fidelity term. Being parametrized, $(\mathrm{P1})$ becomes flexible enough to impose either one of the two structures on $\mathbf{W}$. For example, by setting $\gamma=0$, $(\mathrm{P1})$ results in searching for a matrix that is of low-rank structure. Accordingly, setting $\tau=0$ is tantamount to searching for a sparse matrix. The flexibility of the proposed model provides certainly an advantage over either low-rank or sparse estimation methods, as it will also be demonstrated later, in the experimental results section. 

It is also worth pointing out that ($\mathrm{P1}$)  involves the convex surrogates of the zero norm $\|\mathbf{W}\|_0$ and $\mathrm{rank}(\mathbf{W})$, i.e., the $\ell_1$ and the nuclear norm, respectively. In an attempt to promote further the robustness and consistency of the proposed estimator, we propose to use {\it weighted} $\ell_1$ and nuclear norms in $(\mathrm{P1})$. Such an approach is expected to enhance the sparsity on the individual elements $w_{ij}$ and the singular values $\sigma_i(\mathbf{W})$, e.g. \cite{zou2006adaptive,candes2008enhancing, lu2014generalized,kim13}. These weighted norms are defined as
\begin{align}
 \| \mathbf{A} \odot \mathbf{W}\|_1 & = \sum_{i=1}^{N}\sum_{j=1}^{K} a_{ij} |w_{ij}|,  \label{eq:weightedmatrixnorm}\\
\|\mathbf{W}\|_{\mathbf{b},\ast} & = \sum_{i=1}^{\mathrm{rank}(\mathbf{W})} b_i \sigma_i(\mathbf{W}), \label{eq:weightedtracenorm}
\end{align}
where $a_{ij}$ and $b_i$ are nonnegative weighting coefficients (i.e. $a_{ij}\geq 0 $ and $b_i \geq 0$). Utilizing (\ref{eq:weightedmatrixnorm}) and (\ref{eq:weightedtracenorm}), the proposed optimization problem is rewritten as 
\begin{align}
(\mathrm{P2}): \ \hat{\mathbf{W}}= \underset{\mathbf{W}\in\mathbb{R}_+^{N\times K}}{\text{argmin  }}  \left \{ \frac{1}{2}\|{\mathbf Y} - \mathbf{\Phi}{\mathbf W}\|_{F}^2 + \gamma \|\mathbf{A} \odot \mathbf{W}\|_1 + \tau  \|\mathbf{W}\|_{\mathbf{b},\ast} \right \} .
\label{eq:simultsplradaptive}
\end{align}
To the best of our knowledge, such a formulation, has not been used before as a regularizer for promoting simultaneously sparsity and low-rankness. In the following, problem (P2) will be studied under the following assumption. 
\begin{assumption}
For the weighting coefficients $b_i$ of the nuclear norm it holds that $b_i = b$, $i=1,2,\ldots,K$.
\end{assumption}
Under Assumption 1, the nuclear norm is convex \cite{gu2014weighted,lu2014generalized}, while the weighted $\ell_1$ norm is always convex for nonnegative $\mathbf{A}$. Thus, the overall cost function of  (P2) is convex. Alternative options for the selection of parameters $\mathbf{A}$ and $\mathbf{b}$ are discussed in section III-C. Although convex, $(\mathrm{P2})$ under Assumption 1 is a nontrivial problem to solve, due to the non-differentiable form of the $\ell_1$ and nuclear norm regularizers, \cite{bach2011convex}. 
In the following, we suitably explore two standard convex optimization tools to tackle this problem; an incremental proximal method and an alternating direction method of multipliers (ADMM) based technique.
\section{The Proposed algorithms}
\label{sec:proposedalgorithms}
In this section we present two algorithms to address the non-smooth, constrained, convex optimization problem in $(\mathrm{P2})$. The first one is an incremental proximal algorithm, recently presented and analyzed in \cite{bertsekas2011incremental}, which makes use of the proximal operators of all the terms appearing in $(\mathrm{P2})$, while the second exploits the splitting strategy of the ADMM philosophy, \cite{boyd2011distributed}. 

\subsection{Incremental proximal descent sparse and low-rank unmixing algorithm}
\label{sec:proximal}

Let us first recall that the proximal operator of a function $f(\cdot)$ is defined as \cite{2011combettes,parikh2013proximal}, 
\begin{align}
\mathrm{prox}_{\lambda f(\cdot)}(\boldsymbol{\upsilon}) =  \mathrm{arg}\underset{\mathbf{x}}{\mathrm{min  }} \left ( f(\mathbf{x}) +  \frac{1}{2\lambda} \|\mathbf{x}-\boldsymbol\upsilon\|_2^2 \right),
\label{def:proximal}
\end{align}
where $\boldsymbol \upsilon \in \mathbb{R}^n$ and $\mathbf{x} \in \mathrm{dom}(f)$, the domain of $f$. In \cite{bertsekas2011incremental} the following minimization problem is considered
\begin{align}
\underset{\mathbf{x} \in {\cal X}}{\text{min}}\sum_{i=1}^{m}f_i(\mathbf{x})
\label{eq:minber}
\end{align}
where $f_i(\mathbf{x}), i=1,2,\ldots,m$ are convex functions and ${\cal X}\subseteq \mathbb{R}^n$ is a closed convex set. One version of the algorithm proposed in \cite{bertsekas2011incremental} to solve this problem is the following. The proximal operators of all $f_i$'s are first derived and then a sequential scheme is defined, in which the proximal operator of $f_i(\mathbf{x})$ is evaluated at the point provided by its predecessor (the proximal operator of $f_{i-1}(\mathbf{x})$), for $i=2,3,\ldots,m$. This procedure is repeated in a cyclic manner at each iteration of the algorithm\footnote{Instead of sequential, a randomized evaluation of the proximals of $f_i$'s could be also employed, \cite{bertsekas2011incremental}.}. A convergence and rate of convergence analysis of this incremental proximal scheme is also given in \cite{bertsekas2011incremental}. After this short introduction, we may observe that $(\mathrm{P2})$ in (\ref{eq:simultsplradaptive}) with $\mathbf{b}=b\mathbf{1}$ has exactly the same form with the minimization problem in (\ref{eq:minber}), with respect to $\mathbf{W}$. Embedding the nonnegativity to the cost function in (\ref{eq:simultsplradaptive}) we obtain the following regularized quadratic loss function, 
\begin{align}
\mathcal{L}_1(\mathbf{W}) &=  \frac{1}{2}\|{\mathbf Y} - \mathbf{\Phi}{\mathbf W}\|_{F}^2 +  \gamma \|\mathbf{A} \odot \mathbf{W}\|_1 +  \tau \|\mathbf{W}\|_{\mathbf{b},\ast}  +  {\cal I}_{\mathbb{R}_{+}}({\mathbf W}),
\label{eq:proxproblem}
\end{align}
where the nonnegativity constraint is now replaced by the (convex) indicator function ${\cal I}_{\mathbb{R}_{+}}({\mathbf W})$, which is zero when all $w_{ij} \geq 0, i = 1, 2, \dots, N, j = 1, 2, \dots, K$, and $+\infty$ if at least one $w_{ij}$ is negative. Typically, we wish to minimize $\mathcal{L}_1(\mathbf{W})$ with respect to $\mathbf{W}$. Notice that $\mathcal{L}_1(\mathbf{W})$ is the sum of four convex functions and the incremental proximal algorithm of \cite{bertsekas2011incremental} can be applied directly in our problem.  Next, the proximal operators of all four convex functions are obtained. Starting with the least squares fitting term, we readily get
\begin{align}
\mathrm{prox}_{\lambda \frac{1}{2} \|\mathbf{Y} - \mathbf{\Phi} \cdot\|_F^{2}}(\mathbf{W}) = (\mathbf{\Phi}^T\mathbf{\Phi} + \lambda^{-1}\mathbf{I}_N)^{-1}(\mathbf{\Phi}^T\mathbf{Y}+ \lambda^{-1}\mathbf{W})
\label{eq:proximal_frobenius}
\end{align}
Before we give the proximal operators for the next three terms, some necessary definitions are in order. First, we define the soft-thresholding operator on matrix $\mathbf{W}=[w]_{ij}$ as 
\begin{align}
\mathrm{SHR}_\mathbf{\Delta} (\mathbf{W}) = \mathrm{sign}(\mathbf{W}) \ \mathrm{max}(\mathbf{0},|\mathbf{W}|-\mathbf{\Delta}) ,
\label{eq:soft_thr_op}
\end{align}
where  $\mathbf{\Delta} = [\delta]_{ij}$ is the matrix that contains thresholding parameters.
Note that the soft-thresholding in (\ref{eq:soft_thr_op}) is performed in an element-wise manner, i.e., $\mathrm{SHR}_{\delta_{ij}}(w_{ij}) = \mathrm{sign}(w_{ij}) \ \mathrm{max}(0,|x_{ij}|-\delta_{ij})$. Notably, when we apply the soft-thresholding operator on a diagonal matrix, we shrink only the elements belonging to its diagonal. These elements are assumed to be shrinked by thresholding parameters contained in a vector. With this in mind, we  define the singular value thresholding operation by 
\begin{align}
 \mathrm{SVT}&_{\boldsymbol \delta}(\mathbf{W}) = \mathbf{U} \, \mathrm{SHR}_{\boldsymbol \delta}( \mathbf{\Sigma}) \, \mathbf{V}^T \nonumber
\end{align}
where $\mathbf{W} = \mathbf{U}\mathbf{\Sigma}\mathbf{V}^T$ is the singular value decomposition (SVD) of $\mathbf{W}$, and $\boldsymbol{\delta}$ is the vector whose entries are the thresholding parameters that reduce the corresponding diagonal elements of matrix $\mathbf{\Sigma}$. Finally, we define the projection operator on the set of nonnegative real numbers,
\begin{align}
 \Pi_{\mathbb{R}_+}(v) = \mathrm{arg}\underset{x \in \mathbb{R}_+}{\mathrm{min}}|x - v| = \left\{
     \begin{array}{lr}
       0, & v < 0 \\
       v, & v \geq 0 
     \end{array}
   \right.,
\end{align}
which can also be applied to matrices in an element-wise manner. 

Utilizing the above definitions, we can compute the proximal operators for all regularizing convex functions in (\ref{eq:proxproblem}). Specifically, $\mathrm{prox}_{\gamma \|\mathbf{A} \odot \cdot \|_1}(\mathbf{W})$ is computed by soft-thresholding matrix $\mathbf{W}$ with $\gamma\mathbf{A}$ as follows, 
\begin{align}
\mathrm{prox}_{\gamma \|\mathbf{A} \odot \cdot\|_1}(\mathbf{W}) = \mathrm{SHR}_{\gamma \mathbf{A} }(\mathbf{W}).
\label{eq:proximal_ell_1}
\end{align}
Similarly, the proximal operator of the nuclear norm can be expressed via a soft thresholding operation on the singular values of  $\mathbf{W}$, i.e., 
\begin{align}
& \mathrm{prox}_{{\tau}\|\cdot\|_{\mathbf{b},\ast}}(\mathbf{W})  = \mathrm{SVT}_{\tau \mathbf{b}}(\mathbf{W}) .
\label{eq:proximal_nuclear}
\end{align}
Moreover, the computation of $\mathrm{prox}_{{\cal I}_{\mathbb{R}_{+}}(\cdot)}({\mathbf W})$ reduces to a projection operation, i.e.,
\begin{align}
\mathrm{prox}_{{\cal I}_{\mathbb{R}_+}(\cdot)}(\mathbf{W}) = \Pi_{\mathbb{R}_+}(\mathbf{W}).
\label{eq:proximal_nonnegative}
\end{align}
%
The proposed incremental proximal sparse and low-rank unmixing algorithm (IPSpLRU) iterates among the proximal operators  (\ref{eq:proximal_frobenius}), (\ref{eq:proximal_ell_1}), (\ref{eq:proximal_nuclear})  and (\ref{eq:proximal_nonnegative}) in a cyclic order until convergence, \cite{bertsekas2011incremental}.  IPSpLRU is summarized in Algorithm \ref{tab:ipsplru}. Note that in order to retain the convexity of the composite functions, the weighting parameters $\mathbf{A}$ and $\mathbf{b}$ are initialized and kept fixed during the execution of the algorithm. The issue of dynamic selection of these parameters is discussed in Section \ref{sec:selectionparameters}.  
\begin{algorithm}[t]
\caption{The proposed IPSpLRU algorithm}
\begin{algorithmic}
\STATE Inputs  $\mathbf{Y}$, $\boldsymbol\Phi$  
\STATE Select parameters  $\mathbf{A}, \mathbf{b},\lambda,\tau,\gamma$
\STATE Set $t=0$, $\mathbf{R} = (\mathbf{\Phi}^T\mathbf{\Phi}+\lambda^{-1}\mathbf{I}_N)^{-1}$, $\mathbf{P} = \mathbf{\Phi}^T\mathbf{Y}$, $\mathbf{Q} = \mathbf{R}\mathbf{P}$
\STATE Initialize $\mathbf{W}^0$
\REPEAT
\STATE $\mathbf{W}^{t} = \mathbf{Q}+\lambda^{-1}\mathbf{R}\mathbf{W}^t$
\STATE $\mathbf{W}^{t} = \mathrm{prox}_{\gamma \|\mathbf{A} \odot  \cdot\|_1}(\mathbf{W}^t)$
\STATE $\mathbf{W}^{t} = \mathrm{prox}_{{\tau}\|\cdot\|_{\mathbf{b},\ast}}(\mathbf{W}^t)$
\STATE $\mathbf{W}^{t+1} = \mathrm{prox}_{{\cal I}_{\mathbb{R}_+}(\cdot)}(\mathbf{W}^t)$
\UNTIL{convergence}
\STATE Output : Abundance matrix $\hat{\mathbf{W}} = \mathbf{W}^{t+1}$
\end{algorithmic}
\label{tab:ipsplru}
\end{algorithm}
Concerning the computational complexity of IPSpLRU, the most complex step is the SVD of the abundance matrix $\mathbf{W}^t$, which takes place at each iteration and is of the order of $\mathcal{O}(KN^2 + K^3)$, \cite{golub2012matrix}. Note that matrices $\mathbf{R} = (\mathbf{\Phi}^T\mathbf{\Phi}+\lambda^{-1}\mathbf{I}_N)^{-1}$, $\mathbf{P} = \mathbf{\Phi}^T\mathbf{Y}$ and $\mathbf{Q} = \mathbf{R}\mathbf{P}$ are computed only once at the initialization stage and thus the first step of the algorithm just requires a fast  matrix-by-matrix multiplication. The algorithm converges rapidly and terminates when either the following stopping criterion is satisfied, 
\begin{align}
\frac{||\mathbf{W}^{t+1}-\mathbf{W}^{t}||_{F}^{2}}{||\mathbf{W}^{t}||_{F}^{2}} < \delta
\label{eq:sc_prox}
\end{align}
where $\delta$ is a predefined threshold value, or a preset maximum number of iterations is reached. In the following section we present an alternative approach to solve the same problem by employing a primal-dual ADMM type technique.  
%
%
\subsection{Alternating direction method of multipliers for sparse and low rank unmixing }
\label{sec:admmsparseandlowrank}

In this section, we develop an instance of the alternating direction method of multipliers that solves the abundance matrix estimation problem $(\mathrm{P2})$. To proceed, we utilize the auxiliary matrix variables $\boldsymbol \Omega_1, \boldsymbol \Omega_2, \boldsymbol \Omega_3$ and $\boldsymbol \Omega_4$ of proper dimensions (similar to \cite{iordache2014collaborative,sunspi}), and reformulate the original problem $(\mathrm{P2})$ into its equivalent ADMM form, \cite{boyd2011distributed}, 
\begin{align}
(\mathrm{P3}): \ \underset{\boldsymbol \Omega_1, \boldsymbol \Omega_2, \boldsymbol \Omega_3, \boldsymbol \Omega_4}{\text{min  }} \left \{\frac{1}{2}\|\boldsymbol \Omega_1 - \mathbf{Y}\|_{F}^2 +  \gamma \| \mathbf{A} \odot \boldsymbol \Omega_2\|_1 + \tau\|\boldsymbol \Omega_3\|_{\mathbf{b},\ast} +  {\cal I}_{\mathbb{R}_{+}}({\mathbf \Omega}_4) \right \} \label{eq:admm_form} \\
\mathrm{s.t.} \ \boldsymbol \Omega_1 - \mathbf{\Phi}\mathbf{W} = \mathbf{0}, \ \boldsymbol \Omega_2 - \mathbf{W}=\mathbf{0}, \ \boldsymbol \Omega_3 - \mathbf{W}=\mathbf{0}, \ \boldsymbol \Omega_4 - \mathbf{W}=\mathbf{0} \nonumber
\end{align} 
Based on $\mathrm{(P3)}$, the following augmented Lagrangian function is defined,
\begin{align}  
\mathcal{L}_2(&\mathbf{W}, \boldsymbol \Omega_1, \boldsymbol \Omega_2, \boldsymbol \Omega_3, \boldsymbol \Omega_4) =  \frac{1}{2}\|\boldsymbol \Omega_1 - \mathbf{Y}\|_{F}^2 +  \gamma \| \mathbf{A} \odot \boldsymbol \Omega_2\|_1 + \tau\|\boldsymbol \Omega_3\|_{\mathbf{b},\ast} +  {\cal I}_{\mathbb{R}_{+}}({\mathbf \Omega}_4)  \nonumber \\ 
& + \mathrm{tr} \left [ \mathbf{D}_{1}^T\left( \boldsymbol \Omega_1 - \boldsymbol \Phi \mathbf{W} \right) \right ] + \mathrm{tr} \left [ \mathbf{D}_{2}^T\left(\boldsymbol \Omega_2 - \mathbf{W}\right) \right ]  + \mathrm{tr} \left [ \mathbf{D}_{3}^T\left(\boldsymbol \Omega_3 - \mathbf{ W}\right) \right ] + \mathrm{tr} \left [ \mathbf{D}_{4}^T\left( \boldsymbol \Omega_4 - \mathbf{W}\right) \right ] \nonumber\\
& +\frac{\mu}{2} \left(\|\mathbf{\Phi W} - \boldsymbol \Omega_1\|_{F}^2 + \|\mathbf{W} - \boldsymbol \Omega_2\|_{F}^2 + \|\mathbf{W} - \boldsymbol \Omega_3\|_{F}^2  + \|\mathbf{W} - \boldsymbol \Omega_4 \|_{F}^2 \right)
\label{eq:auglagr1}
\end{align}
where the $L\times K$ matrix $\mathbf{D}_1$, and the $N \times K$ matrices $\mathbf{D}_2, \mathbf{D}_3, \mathbf{D}_4$ are the Lagrange multipliers and $\mu > 0$ is a positive penalty parameter. Note that again the nonnegative weights $\mathbf{A}$ and $\mathbf{b}$ are considered to be constant and assumption 1 also holds here. Let 
\begin{align}
\mathbf{\Omega} = \left[
\begin{tabular}{c}
$\boldsymbol \Omega_1$ \\
$\mathbf{\Omega}_2$ \\
$\mathbf{\Omega}_3$\\
$\mathbf{\Omega}_4 $
\end{tabular}
\right], \ 
\mathbf{G} = \left[
\begin{tabular}{c}
$\boldsymbol \Phi$ \\
$\mathbf{I}_N$ \\
$\mathbf{I}_N$\\
$\mathbf{I}_N $
\end{tabular}
\right],
\text{ and  } \mathbf{B} =  \left[
\begin{array}{c c c c}
 - \mathbf{I}_L & \mathbf{0} & \mathbf{0} & \mathbf{0} \\
 \mathbf{0} & -\mathbf{I}_N  & \mathbf{0} & \mathbf{0} \\
\mathbf{0} & \mathbf{0} &- \mathbf{I}_N & \mathbf{0}  \\
\mathbf{0} & \mathbf{0} & \mathbf{0} & - \mathbf{I}_N
\end{array}
 \right].
\end{align} 
Then (\ref{eq:auglagr1}) can be written in an equivalent form as 
\begin{align}
\mathcal{L}_{3}(\mathbf{W},\boldsymbol \Omega, \boldsymbol \Lambda) & =  \frac{1}{2}\|\boldsymbol \Omega_1 - \mathbf{Y}\|_{F}^2 +  \gamma \| \mathbf{A} \odot \boldsymbol \Omega_2\|_1 + \tau \|\boldsymbol \Omega_3\|_{\mathbf{b},\ast} \nonumber \\  
&+  {\cal I}_{\mathbb{R}_{+}}(\boldsymbol \Omega_4)  + \frac{\mu}{2}\|\mathbf{GW} + \mathbf{B} \boldsymbol \Omega - \mathbf{\Lambda}\|_{F}^2,
\label{eq:auglagr2}
\end{align}
where $\mathbf{\Lambda} = \left[\mathbf{\Lambda}^T_1 \; \mathbf{\Lambda}^T_2 \; \mathbf{\Lambda}^T_3 \; \mathbf{\Lambda}^T_4 \right]^T, \boldsymbol \Lambda_i = (1/\mu) \mathbf{D}_i, i = 1, \dots, 4$, contains the scaled Lagrange multipliers. Having expressed the augmented Lagrangian function as in (\ref{eq:auglagr2}), the ADMM proceeds by minimizing $\mathcal{L}_{3}(\mathbf{W},\boldsymbol \Omega, \boldsymbol \Lambda)$ sequentially, each time with respect to a single matrix variable, keeping the remaining variables at their latest values. The dual variables (Lagrange multipliers) are also updated via a gradient ascend step at the end of each alternating minimization cycle. 

To elaborate further on the steps of the ADMM, the optimization with respect to $\mathbf{W}$ gives
\begin{align}
\mathbf{W}^{t+1} & = \mathrm{arg}\underset{\mathbf{W}}{\mathrm{min}} \ \mathcal{L}_3(\mathbf{W}, \boldsymbol \Omega^t,\boldsymbol \Lambda^t) \nonumber \\
& = \left(\mathbf{\Phi}^T\mathbf{\Phi} + 3\mathbf{I}_N\right)^{-1} [\mathbf{\Phi}^T(\boldsymbol \Omega_{1}^t+\mathbf{\Lambda}_{1}^t)  + \boldsymbol \Omega_{2}^t+\mathbf{\Lambda}_{2}^t+ \boldsymbol \Omega_{3}^t + \mathbf{\Lambda}_{3}^t + \boldsymbol \Omega_{4}^t + \mathbf{\Lambda}_{4}^t ].
\label{eq:Wupd}
\end{align}
Next, the optimization with respect to $\boldsymbol \Omega_1$ is performed as 
\begin{align}
{\boldsymbol \Omega}_{1}^{t+1} &= \mathrm{arg}\underset{\boldsymbol \Omega_1}{\mathrm{min}} \ \mathcal{L}_3(\mathbf{W}^{t+1}, \boldsymbol \Omega, \boldsymbol \Lambda^t) \nonumber \\ 
&= \frac{1}{1 + \mu}\left(\mathbf{Y} + \mu\left(\mathbf{\Phi}\mathbf{W}^{t+1} - \mathbf{\Lambda}_{1}^t\right) \right).
\label{eq:V1upd}
\end{align}
The remaining auxiliary variables $\boldsymbol \Omega_2,\boldsymbol \Omega_3$, and $\boldsymbol \Omega_4$ are involved in non-differentiable norms, namely, the weighted $\ell_1$ norm, the weighted nuclear norm, and the indicator function, respectively. In this regard, the minimization task with respect to these variables resolves to computing some of the proximity operators that we introduced in the previous section. 
Minimizing (\ref{eq:auglagr2}) with respect to $\boldsymbol \Omega_{2}$ yields
\begin{align}
 \boldsymbol \Omega_{2}^{t+1}  &=  \mathrm{arg}\underset{\boldsymbol \Omega_2}{\mathrm{min}} \ \mathcal{L}_3(\mathbf{W}^{t+1}, \boldsymbol \Omega, \boldsymbol \Lambda^t)  \nonumber \\
& = \mathrm{SHR}_{\gamma \mathbf{A} }(\mathbf{W}^{t+1} - \boldsymbol \Lambda_2^t).
\label{eq:V2upd}
\end{align}
In the same vein, $\boldsymbol \Omega_3$ is computed by a shrinkage operation, 
\begin{align}
\boldsymbol \Omega_{3}^{t+1} &= \mathrm{arg}\underset{\boldsymbol \Omega_3}{\mathrm{min}} \ \mathcal{L}_3(\mathbf{W}^{t+1}, \boldsymbol \Omega, \boldsymbol \Lambda^t)  \nonumber \\
& = \mathrm{SVT}_{\tau \mathbf{b} }(\mathbf{W}^{t+1} - \boldsymbol \Lambda_3^t).
\label{eq:V3upd}
\end{align}
Next, for the auxiliary variable $\boldsymbol \Omega_4$,  a projection onto the nonnegative orthant is required,
\begin{align} 
\boldsymbol \Omega_{4}^{t+1} & =  \mathrm{arg}\underset{\boldsymbol \Omega_4}{\mathrm{min}} \ \mathcal{L}_3(\mathbf{W}^{t+1}, \boldsymbol \Omega, \boldsymbol \Lambda^t)  \nonumber \\
 & = \Pi_{\mathbb{R}_+}(\mathbf{W}^{t+1} - \boldsymbol \Lambda_4^t).
\label{eq:V4upd}
\end{align}

At the final step of the proposed method, the scaled Lagrange multipliers in $\mathbf{\Lambda}$ are sequentially updated by performing gradient ascent on the dual problem \cite{boyd2011distributed}, as follows, 
\begin{align}
\mathbf{\Lambda}_{1}^{t+1} &= \mathbf{\Lambda}_{1}^{t} - \mathbf{\Phi W}^{t+1} + \boldsymbol \Omega_{1}^{t+1} \nonumber \\
\mathbf{\Lambda}_{i}^{t+1} &= \mathbf{\Lambda}_{i}^{t} - \mathbf{W}^{t+1} + \boldsymbol \Omega_{i}^{t+1}, i = 2,3,4
\label{eq:Lupd}
\end{align}
The proposed algorithm, termed the alternating direction sparse and low-rank unmixing algorithm (ADSpLRU), is summarized in Algorithm \ref{SPLR-ADMM}. An iteration of ADSpLRU consists of the update steps given in (\ref{eq:Wupd}), (\ref{eq:V1upd}), (\ref{eq:V2upd}), (\ref{eq:V3upd}), (\ref{eq:V4upd}), and (\ref{eq:Lupd}). Its computational complexity is $\mathcal{O}(KLN + KN^2)$ per iteration, slightly higher than that of IPSpLRU, since it usually holds $L>N$. However, as verified by the simulations of the next section, ADSpLRU converges a little faster than IPSpLRU to a slightly lower steady-state error\footnote{The reason for this may be that ADSpLRU manipulates the whole cost function at each step, while IPSpLRU splits the cost function in a number of convex terms and treats each term individually at every step of the algorithm.}, while its convergence is also guaranteed as explained in \cite{eckstein1992douglas}.
\begin{algorithm}[t]
\caption{The proposed ADSpLRU algorithm}
\label{SPLR-ADMM}
\begin{algorithmic}
\STATE Inputs  $\mathbf{Y}$, $\boldsymbol\Phi$  
\STATE Select parameters  $\mathbf{A}, \mathbf{b}, \mu,\tau,\gamma $
\STATE Set $t=0$,\ $\mathbf{R} = \left(\mathbf{\Phi}^T\mathbf{\Phi} + 3\mathbf{I}_N\right)^{-1}$
\STATE Initialize  $\mathbf{W}^0,\boldsymbol \Omega^0,\mathbf{\Lambda}^0$
\REPEAT
\STATE $\mathbf{W}^{t+1} = \mathbf{R} [\mathbf{\Phi}^T(\boldsymbol \Omega_{1}^t+\mathbf{\Lambda}_{1}^t) + \boldsymbol \Omega_{2}^t +\mathbf{\Lambda}_{2}^t + \boldsymbol \Omega_{3}^t + \mathbf{\Lambda}_{3}^t + \boldsymbol \Omega_{4}^t + \mathbf{\Lambda}_{4}^t ]$ 
\STATE $\boldsymbol \Omega_{1}^{t+1} = 1 \slash (1 + \mu)\left(\mathbf{Y} + \mu\left(\mathbf{\Phi}\mathbf{W}^{t+1} - \mathbf{\Lambda}_{1}^{t}\right) \right) $
\STATE $\boldsymbol \Omega_{2}^{t+1} = \mathrm{SHR}_{\gamma \mathbf{A} }(\mathbf{W}^{t+1} - \boldsymbol \Lambda_2^t) $
\STATE $\boldsymbol \Omega_{3}^{t+1} = \mathrm{SVT}_{\tau \mathbf{b} }(\mathbf{W}^{t+1} - \boldsymbol \Lambda_3^t)$
\STATE $\boldsymbol \Omega_{4}^{t+1} = \Pi_{\mathbb{R}_+}(\mathbf{W}^{t+1} - \boldsymbol \Lambda_4^t)$
\STATE $\mathbf{\Lambda}_{1}^{t+1} = \mathbf{\Lambda}_{1}^{t} - \mathbf{\Phi W}^{t+1} + \boldsymbol \Omega_{1}^{t+1}$
\STATE $\mathbf{\Lambda}_{i}^{t+1} = \mathbf{\Lambda}_{i}^{t} - \mathbf{W}^{t+1} + \boldsymbol \Omega_{i}^{t+1}$, $i=2,3,4$
\UNTIL{convergence}
\STATE Output : Abundance matrix $\hat{\mathbf{W}} = \mathbf{W}^{t+1}$
\end{algorithmic}
\end{algorithm}
Note that all functions that form the objective function $\mathcal{L}_1(\mathbf{W})$ in (\ref{eq:proxproblem}), are closed, proper and convex. Since matrix $\mathbf{G}$ has full column rank, the convergence conditions defined in \cite{eckstein1992douglas} are met and if an optimal solution exists, ADSpLRU converges, for any $\mu>0$. This in turn implies that for the primal and dual residuals $\mathbf{r}^{t}$, $\mathbf{d}^{t}$  given by
\begin{align}
&\mathbf{r}^{t} = \mathbf{G}\mathbf{W}^{t} + \mathbf{B}\mathbf{\Omega}^{t},  \nonumber\\
&\mathbf{d}^{t} = \mu \mathbf{G}^\top \mathbf{B}\left(\mathbf{\Omega}^{t} - \mathbf{\Omega}^{t-1}\right) \nonumber
\end{align}
it holds that, $\mathbf{r}^{t}\rightarrow 0$  and $\mathbf{d}^{t}\rightarrow 0$, respectively, as $t\rightarrow \infty$. In this work, ADSpLRU stops when either the following termination criterion 
\begin{align}
\|\mathbf{r}^{t}\|_2 \leq \zeta \ \text{and} \ \|\mathbf{d}^{t}\|_2 \leq \zeta
\end{align}
holds for the primal and dual residuals, where $\zeta = \sqrt{(3N+L)K}\zeta^{rel}$,\cite{boyd2011distributed}, (the relative tolerance $\zeta^{rel} > 0$ takes its value depending on the application, and in our experimental study has been empirically determined to $10^{-4}$), or the maximum number of iterations is reached. 
\subsection{Selection of weighting coefficients and regularization parameters}
\label{sec:selectionparameters}
As mentioned previously, in both IPSpLRU and ADSpLRU the weighting coefficients $\mathbf{A}$ and $\mathbf{b}$ are predetermined, satisfy certain constraints and remain constant during the execution of the algorithms. As is widely known, \cite{zou2006adaptive,candes2008enhancing,gu2014weighted}, a proper selection of these parameters is quite crucial as for the accuracy of the estimations. In view of this, two potential choices are a) to select the weighting coefficients based on the least squares estimate $\mathbf{W}^{LS}$ of $\mathbf{W}$ i.e.,  
\begin{align}
a_{ij} =  \left(\frac{1}{w_{ij}^{LS} + \epsilon }\right) \text{    and  } b_{i} =  \left(\frac{1}{\sigma_{i}(\mathbf{W}^{LS}) + \epsilon} \right),
\label{eq:lsweights}
\end{align}
where $\epsilon= 10^{-16}$ is a small constant added to avoid singularities and b) to update them at each iteration $t$ of the algorithms based on the current estimate $\mathbf{W}^t$ of $\mathbf{W}$ i.e.,
\begin{align}
a_{ij}^t =  \left(\frac{1}{w_{ij}^{t} + \epsilon }\right) \text{    and  } b_{i}^t =  \left(\frac{1}{\sigma_{i}(\mathbf{W}^{t}) + \epsilon} \right),
\label{eq:dynweights}
\end{align}

It should be noted, that both these two options render the minimization problem $(\mathrm{P2})$ nonconvex, since the weighted nuclear norm is known to be convex only if the weights $b_i, \ i=1,2,\dots,K$ are nonnegative and non-ascending, \cite{gu2014weighted,Dong2014}. Additionally, the reweighting norms minimization problem is known to be inherently nonconvex, \cite{candes2008enhancing}, while its theoretical convergence analysis for these cases is difficult to be established. Nevertheless, numerous research works advocate the positive impact of these nonconvex weighted norms on the performance of constrained estimation tasks \cite{Dong2014,candes2008enhancing,gu2014weighted,lu2014generalized}. Along this line of thought, the algorithms presented in the previous section are modified by adopting the reweighting scheme given by (\ref{eq:dynweights}). As verified in our empirical study presented in the next section, such an option enhances to a large degree the effectiveness of the proposed algorithms, while no numerical issues have been encountered in our experiments. 

As far as the remaining parameters is concerned, $\lambda$ and $\mu$, which control the convergence behavior of IPSpLRU and ADSpLRU, respectively, take positive values, with $\mu$ close to zero and $\lambda$ on the order of 1. In all our experiments we fixed $\mu=0.01$ and $\lambda = 0.5$. On the other hand, the low-rank and sparsity promoting parameters $\tau$ and $\gamma$ are chosen via fine-tuning, as is commonly done in relevant deterministic schemes. This is so because the optimal set of these parameters depends on the unknown in advance particular structure of the sought abundance matrix, an issue which is further explained in the next section.  
\section{Experimental Results}
This section unravels the performance characteristics of the proposed IPSpLRU and ADSpLRU algorithms via experiments conducted both on simulated and real data. We compare our techniques with three well-known state-of-the-art unmixing algorithms, namely, the nonnegative constraint sparse unmixing by variable splitting and augmented Lagrangian algorithm (CSUnSAL), \cite{bioucas2010alternating},  the recently reported nonnegative constraint joint-sparse method (MMV-ADMM), \cite{quabundance}, and finally the fast Bayesian inference iterative conditional expectations (BiICE) unmixing algorithm, \cite{themelis2013fast}. The computational complexity (in terms of the number of multiplications) of all the tested algorithms is given in Table \ref{complexity_table}. As shown in the Table, the spatial correlation-aware algorithms namely, IPSpLRU, ADSpLRU and MMV-ADMM,  present higher complexity since the information from $K$ pixels is used for the unmixng of a single pixel. Moreover, it is noticed that among the two proposed algorithms, IPSpLRU has lower computational complexity than ADSpLRU  per iteration, resulting from its more simplistic incremental approach.

In what follows, we first refer to the parameters' setting established for all the involved algorithms, and the  performance evaluation metrics that are utilized in the experimental procedure. To corroborate the effectiveness and robustness of the proposed algorithms we execute five different types of synthetic data experiments whose detailed description is given below. Finally, we empirically compare the abundance maps as revealed by all examined algorithms, when applied on a real hyperspectral image.  
\begin{table*}
\centering
\caption{Computational complexity per pixel and iteration}
\begin{tabular}{| c | c | c | c | c | c |}
\hline 
Algorithm & IPSpLRU & ADSpLRU & CSUnSAl \cite{bioucas2010alternating} & MMV-ADMM \cite{quabundance} & BiICE \cite{themelis2013fast} \\\hline
Computational complexity & $\mathcal{O}(KN^2 + K^3)$ &  $\mathcal{O}(KN^2 + KLN)$  & $\mathcal{O}(N^2)$ & $\mathcal{O}(KN^2 + KLN)$  & $\mathcal{O}(N^2)$ \\\hline
\end{tabular}
\label{complexity_table}
\end{table*}
\subsection{Setting of Parameters and Performance Evaluation Criteria}\label{subsecA}
For simplicity reasons,  we use  $\gamma$ for the sparsity imposing parameter in all tested algorithms (except BiICE which has no regularization parameters, \cite{themelis2012novel}), $\mu$ for the Lagrange multiplier regularization parameter of the ADMM-type techniques and $\lambda$ for the relevant to $\mu$ regularization  parameter of IPSpLRU. Additionally, the low-rank promoting parameter of the proposed algorithms, is denoted by $\tau$. Parameters $\tau$ and $\gamma$ are fine tuned with 10 different values, as shown in Table \ref{parameters_setting}. On the other hand, the Lagrange multiplier regularization parameter $\mu$ and the regularization parameter $\lambda$ of IPSpLRU, which influence to a less extend the efficiency of the corresponding algorithms, are set to a fixed value. 
\begin{table*}[!t]
\renewcommand{\arraystretch}{1.3}
\caption{Parameters Setting}
\label{table_params}
\centering
\begin{tabular}{|c||c | c |c| c|}
\hline \hline
{\bf{Algorithm}}& $\tau$ (rank regularization parameter) & $\gamma$ (sparsity regularization parameter) & $\mu$ & $\lambda$ \\\cline{2-4}
\hline\hline
 IPSpLRU  & $\{0,10^{-10},10^{-9},\dots,10^{-1}\}$&$\{0,10^{-10},10^{-9},\dots,10^{-1}\}$ &  Not applicable & $0.5$ \\\hline
 ADSpLRU & $\{0,10^{-10},10^{-9},\dots,10^{-1}\}$ & $\{0,10^{-10},10^{-9},\dots,10^{-1}\}$ & $10^{-2}$ & Not applicable\\\hline
 CSUnSAL& Not applicable & $\{0,10^{-10},10^{-9},\dots,10^{-1}\}$  & $10^{-2}$ & Not applicable\\\hline
  MMV-ADMM & Not applicable & $\{0,10^{-10},10^{-9},\dots,10^{-1}\}$ & $10^{-2}$ & Not applicable\\\hline
\hline
\end{tabular}
\label{parameters_setting}
\end{table*}
In order to assess the performance of the proposed algorithms and the competing ones, for the experiments conducted on synthetic data we consider two metrics. First, the root mean square error ($\mathrm{RMSE}$),
\begin{align}
\mathrm{RMSE} = \sqrt{\frac{1}{Nn}\sum_{i=1}^{n}\|\hat{\mathbf{w}}_i - \mathbf{w}_{i}\|_2},
\end{align}
where $\hat{\mathbf{w}}_i$ and $\mathbf{w}_{i}$ represent the estimated and actual abundance vectors of the $i$-th pixel respectively, $n$ is the total number of the pixels in the image under study, and $N$, as mentioned in previous sections, stands for the number of endmembers. The second metric, is the signal-to-reconstruction error ($\mathrm{SRE}$),\cite{iordache2011sparse}, which reflects the ratio between the power of the signal and the power of the estimation error, and is given by the following formula
\begin{align}
\mathrm{SRE} = 10 \mathrm{log}_{10}\left(\frac{\frac{1}{n}\sum_{i=1}^{n} \|\hat{\mathbf{w}}_i\|_{2}^2 }{\frac{1}{n}\sum_{i=1}^{n} \|\hat{\mathbf{w}}_i - \mathbf{w}_{i}\|_{2}^2 }\right).
\end{align}
Of great importance is to notice that for the sliding window-based algorithms, the abundance vectors $\mathbf{w}_i$'s and their estimates $\hat{\mathbf{w}}_i$'s coincide with the central column vectors of the corresponding abundance matrices $\mathbf{W}_i$'s and their estimates $\hat{\mathbf{W}}_i$'s, as becomes clear from Fig. \ref{windowing_approach}. 
\subsection{Experiments on Simulated Datacubes}
In the sequel, $N$ endmembers are  randomly selected from the USGS library  $\mathbf{Q}\in\mathbb{R}_+^{224\times498}$, \cite{2007_usgs}, so as to form our endmembers' dictionary  $\mathbf{\Phi}$. Their reflectance values correspond to $L = 224$ spectral bands, uniformly distributed in the interval $0.4 - 2.5 \mu m$. The LMM of eq. (1) is then utilized for generating spectral signatures subject to given, different in each experiment, abundance matrices $\mathbf{W}$'s. 
\subsubsection{Reweighting coefficients efficiency and convergence behavior of IPSpLRU and ADSpLRU}
Herein, we aspire to demonstrate the merits emerging from the utilization of reweighting of $\mathbf{A}$ and $\mathbf{b}$ from (\ref{eq:dynweights}), on the estimation performance of the proposed algorithms. In light of this, we consider a rank $3$ and sparsity level $10\%$ (i.e., $10\%$ of its entries are nonzero) abundance matrix corresponding to $N=50$ endmembers, $K=9$ pixels. Then, $K=9$ spectral signatures are generated according to the LMM and contaminated by Gaussian noise such that SNR = 30dB.

\begin{figure*}
\centering
\includegraphics[width=0.7\linewidth,height=0.6\linewidth]{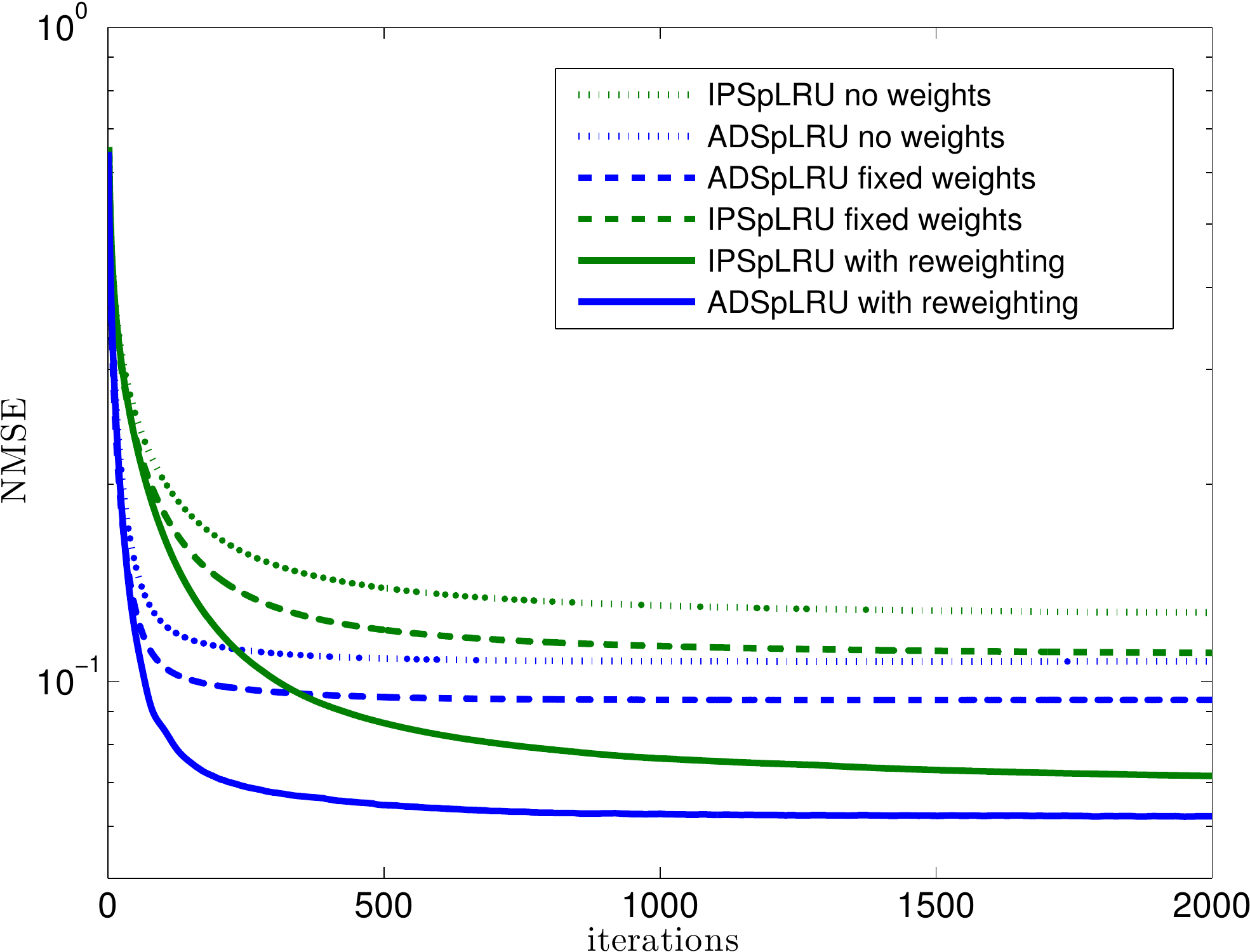} 
\caption{Convergence curves of IPSpLRU and ADSpLRU for a) updating weighting coefficients b) fixed weighting coefficients and c) no weighting coefficients.}
\label{convergence}
\end{figure*}

For $p=100$ realizations, Fig. \ref{convergence} depicts the normalized mean squared estimation error (NMSE) (defined as $\mathrm{NMSE}(t)=\frac{1}{p}\sum_{i=1}^{p}\frac{\|\hat{\mathbf{W}}^t_i - \mathbf{W}_i \|^2_{F}}{\|\mathbf{W}_{i}\|^2_{F}}$, where $\hat{\mathbf{W}}^t_i$ and $\mathbf{W}_i$ are the estimated at the $t$-th iteration and the true matrices respectively, of the $i$-th realization) as it evolves over 2000 iterations. Three different cases are investigated, corresponding to: a) updating weighting coefficients from (\ref{eq:dynweights}) b) keeping fixed the weighting coefficients  based on (\ref{eq:lsweights})  and c) no weighting coefficients i.e. the weighted norms degenerate to their non-weighted  versions by setting $\mathbf{b}=\mathbf{1}$ and $\mathbf{A} = [\mathbf{1},\mathbf{1},\ldots \mathbf{1}]$. As it is clearly evident in Fig. \ref{convergence}, both IPSpLRU and ADSpLRU achieve remarkably higher estimation accuracy in terms of NMSE, when using reweighting as compared to the case that fixed or no weights are employed. It is thus empirically verified that the enhanced efficiency of the reweighted $\ell_1$ and nuclear norms, emphatically advocated in \cite{zou2006adaptive,lu2014generalized,candes2008enhancing}, is retained when using the sum of these two norms. The price to be paid is that such an option might increase numerical risks, since the problem is rendered non-convex and (yet) no theoretical convergence analysis has been established. Nevertheless, it is worthy to mention that, despite the fact that convergence is not theoretically guaranteed, in all our experiments both IPSpLRU and ADSpLRU exhibited a very robust behavior in their convergence process. 

It is also noticed that ADSpLRU needs less iterations to converge as compared to IPSpLRU and it converges to a slightly lower NMSE. This results from the inherent nature of the two proposed algorithms, as explained before. Interestingly, the faster convergence rate of ADSpLRU with reweighting comes at the price of its higher per iteration computational complexity as compared to that of IPSpLRU.
\subsubsection{A toy example}
\label{expa}
\begin{figure*}
\centering
\begin{tabular}{c | c}
 \subfloat[$\mathbf{W}$, Ground truth]{\includegraphics[width=0.22\textwidth,height=0.35\textwidth]{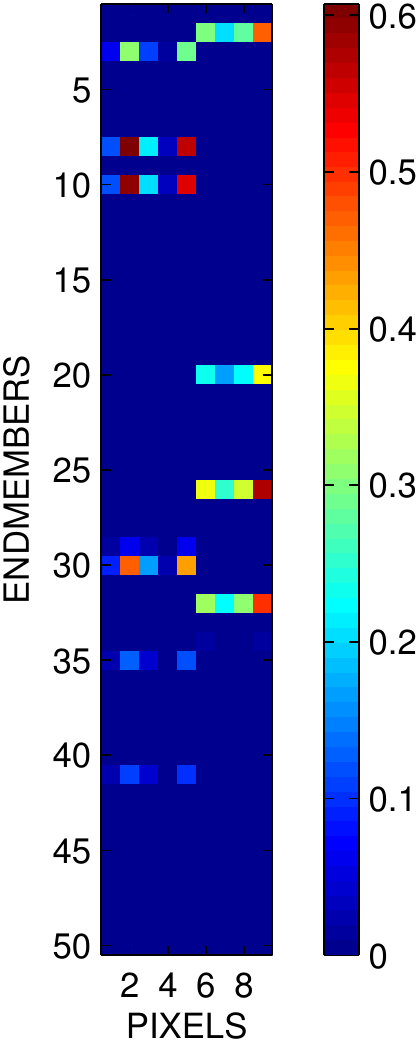} \label{Wgt}} \vspace{-5cm}&
\begin{tabular}{c c | c c |}
\subfloat[$\hat{\mathbf{W}}$, IPSpLRU]{\includegraphics[width=0.18\textwidth,height=0.26\textwidth]{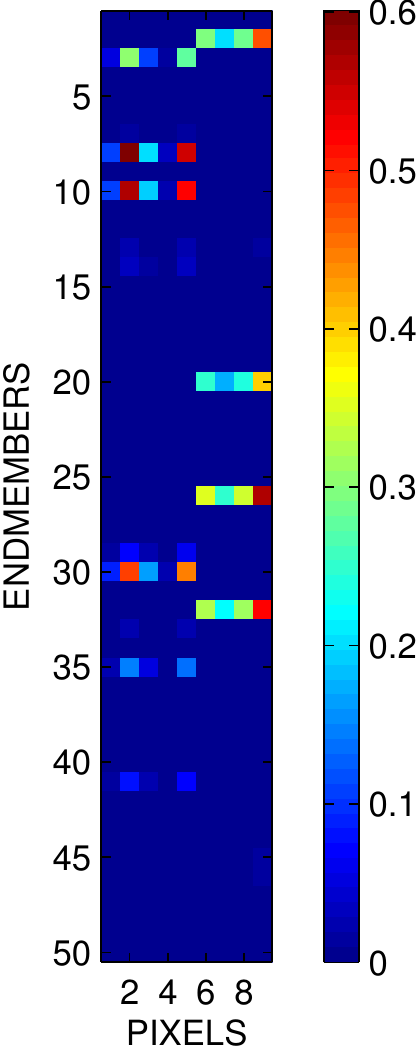}\label{wipsplru}} 
   & \subfloat[residual, IPSpLRU]{\includegraphics[width=0.18\textwidth,height=0.26\textwidth]{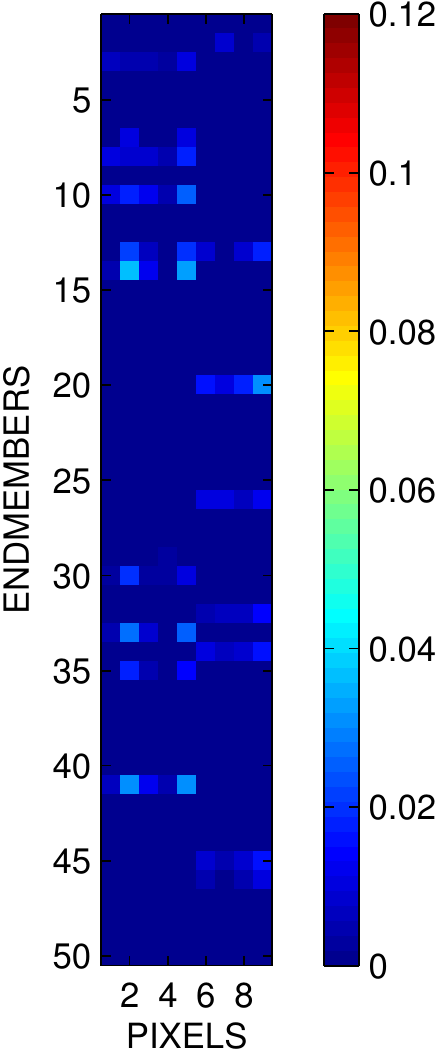}\label{ipsp}}  
      & \subfloat[$\hat{\mathbf{W}}$, ADSpLRU]{\includegraphics[width=0.18\textwidth,height=0.26\textwidth]{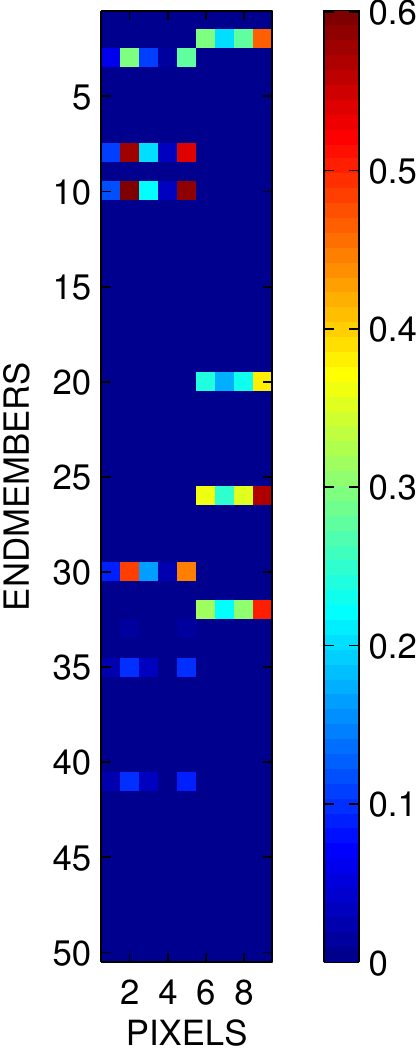}\label{iplr}}
       & \subfloat[residual, ADSpLRU]{\includegraphics[width=0.19\textwidth,height=0.26\textwidth]{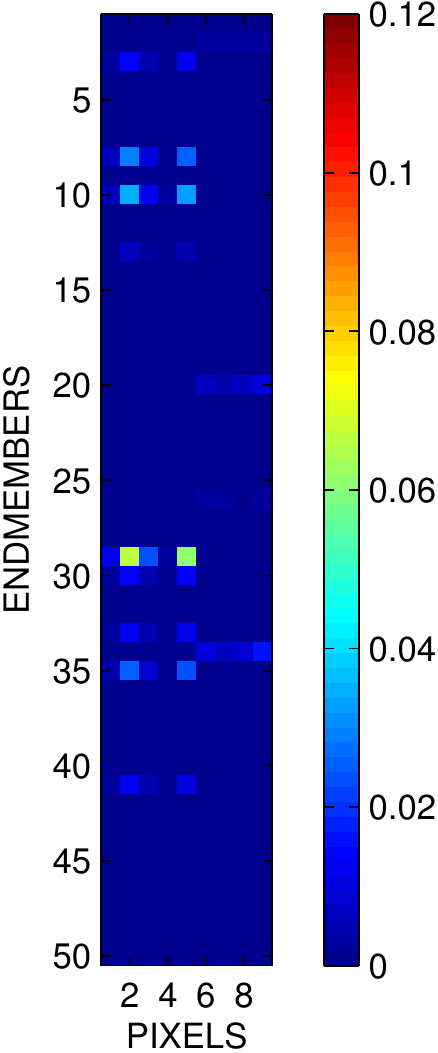}\label{ipsp}}  
      \\
 \subfloat[$\hat{\mathbf{W}}$, IPLRU]{\includegraphics[width=0.18\textwidth,height=0.26\textwidth]{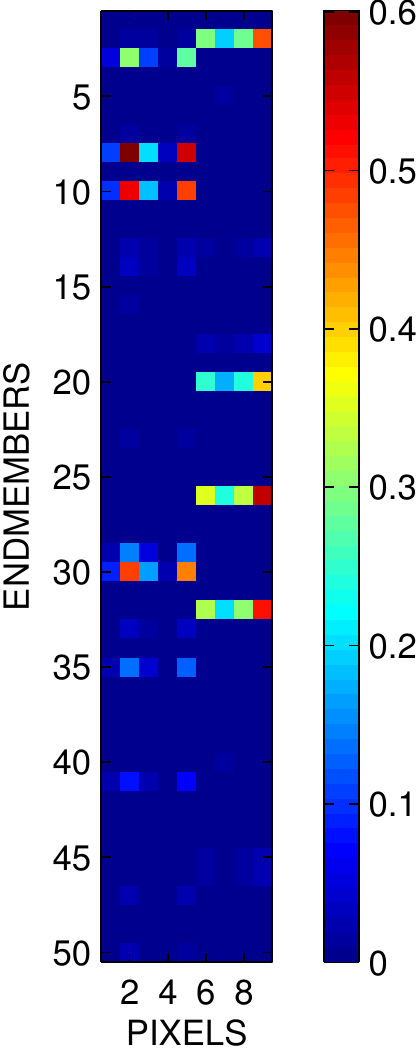}\label{adsplr}} 
   & \subfloat[residual, IPLRU]{\includegraphics[width=0.18\textwidth,height=0.26\textwidth]{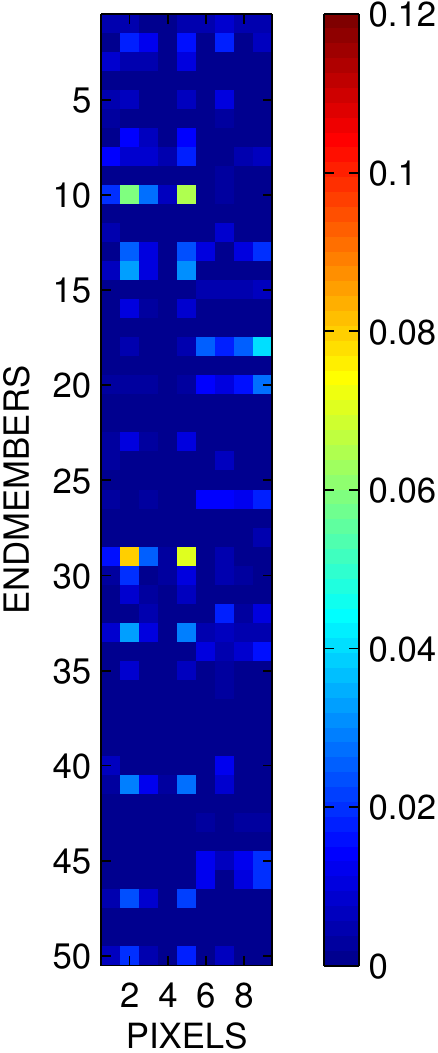}\label{adsp}}
      &   \subfloat[$\hat{\mathbf{W}}$, ADLRU]{\includegraphics[width=0.18\textwidth,height=0.26\textwidth]{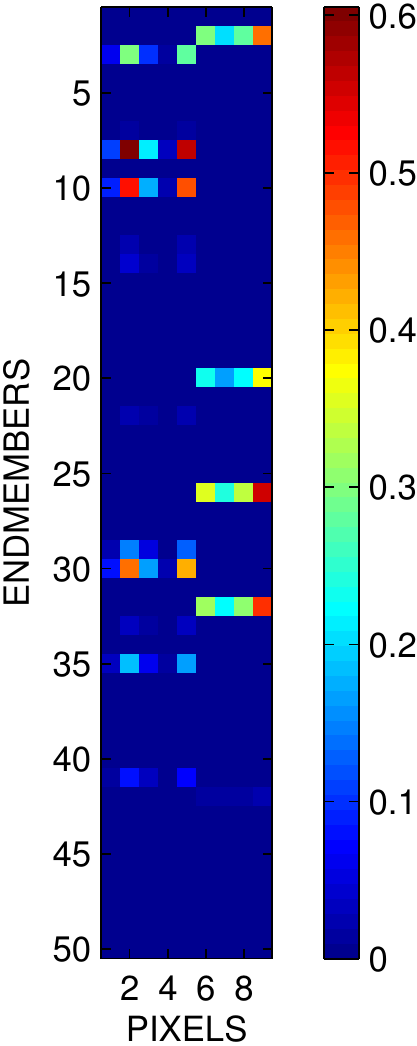}\label{adlr}} 
        & \subfloat[residual, ADLRU]{\includegraphics[width=0.18\textwidth,height=0.26\textwidth]{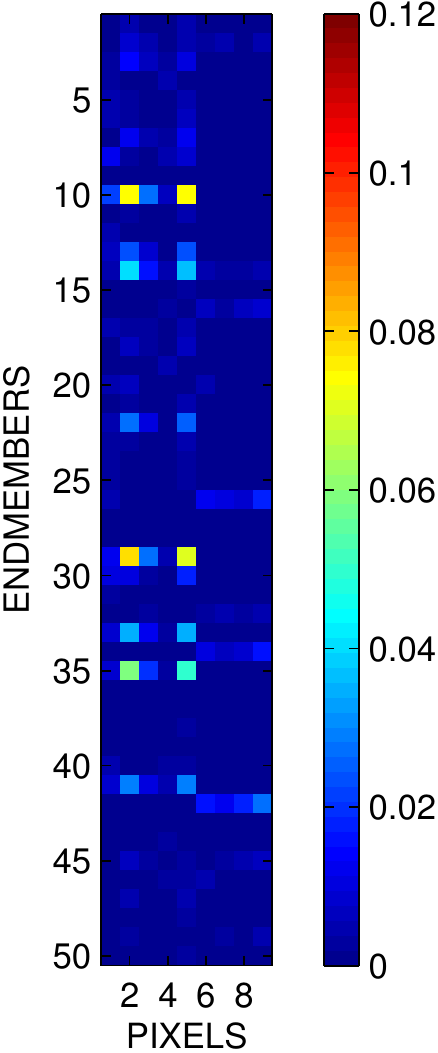}\label{iplr}}  
        \\
      \subfloat[$\hat{\mathbf{W}}$, IPSpU]{\includegraphics[width=0.18\textwidth,height=0.26\textwidth]{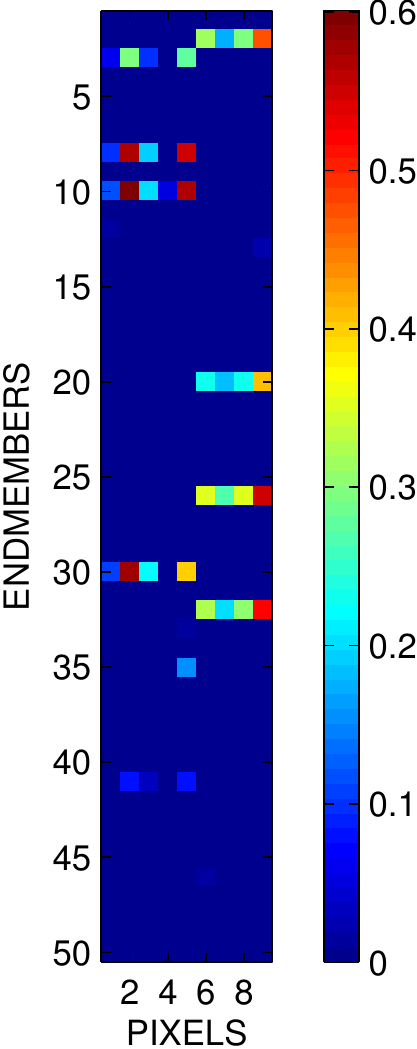}\label{adsplr}} 
   & \subfloat[residual, IPSpU]{\includegraphics[width=0.18\textwidth,height=0.26\textwidth]{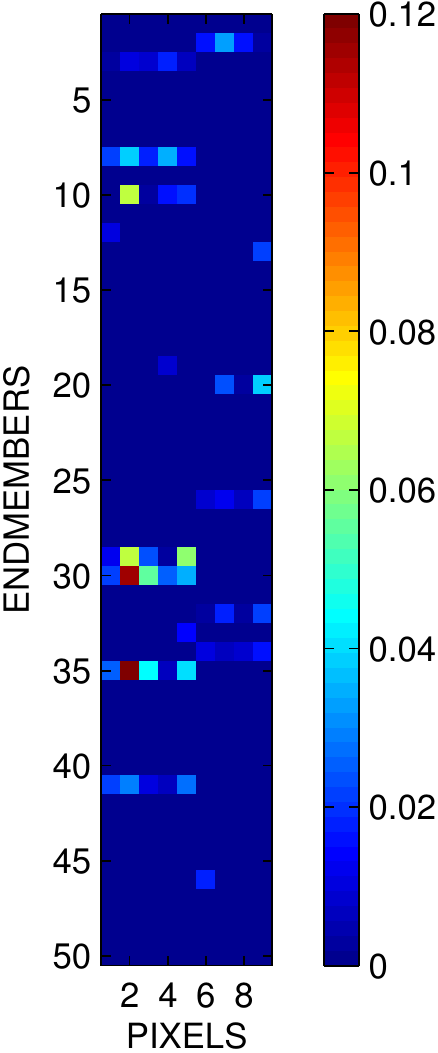}\label{adsp}}
      &   \subfloat[$\hat{\mathbf{W}}$, ADSpU]{\includegraphics[width=0.18\textwidth,height=0.26\textwidth]{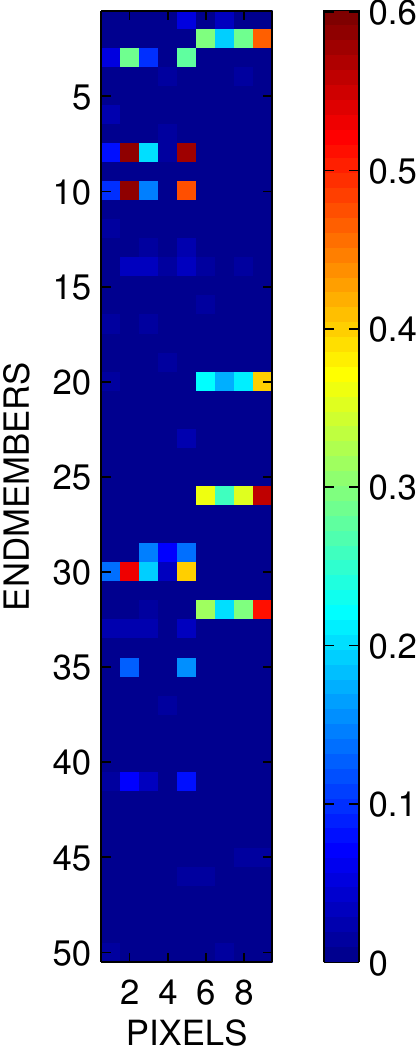}\label{adlr}} 
        & \subfloat[residual, ADSpU]{\includegraphics[width=0.18\textwidth,height=0.26\textwidth]{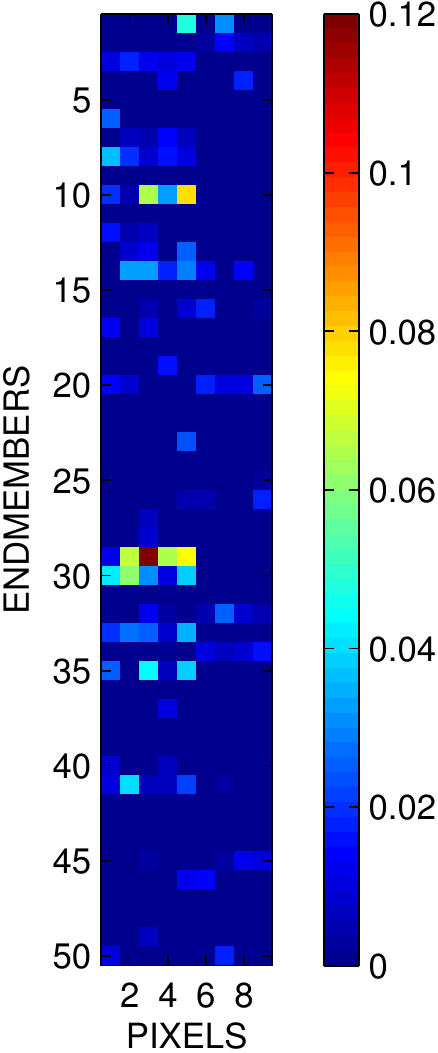}\label{iplr}}  
\end{tabular} \\ 
\resizebox{0.22\textwidth}{!}{
{\begin{tabular}{||c||c ||}
\hline
\bfseries Algorithm & \bfseries RMSE \\
\hline\hline
IPSpLRU & \bfseries 0.0056  \\\hline
IPSpU (sparse only) & 0.0121   \\\hline
IPLRU (low-rank only) & 0.0098  \\\hline
ADSpLRU &\bfseries 0.0061  \\\hline
ADSpU (sparse only) & 0.0130 \\\hline
ADLRU (low-rank only) & 0.0102 \\\hline
\hline
\end{tabular}}}& 
\end{tabular}
\vspace*{2cm}
\caption{Sparse and low-rank algorithms versus their sparse only and low-rank only counterparts.}
\label{fig:exp1a}
\end{figure*}

In this experiment our goal is to highlight the significance of the approach followed in this work, i.e., the simultaneous incorporation of sparsity and low-rankness on the abundance estimation problem. 
To this end, we initially derive the single prior counterparts of our algorithms. We first focus on the low-rankness assumption, thus the sparsity imposing norm is ignored ($\gamma=0$). IPSpLRU  and ADSpLRU are then reduced to their modified versions, namely, IPLRU and ADLRU respectively. As implied by their names, the aforementioned methods  allow exclusively for the low-rank assumption. Similarly, IPSpU and ADSpU  are formed by accounting solely for sparsity. That said, IPSpU and ADSpU emerge after dropping the low-rank prior constraint ($\tau=0$). Next, we generate a $N\times K$ (where  $N=50 \text{ and } K=9$) simultaneously sparse and low-rank abundance matrix ${\mathbf W}$ of rank 2 with sparsity level 20\%, which is graphically illustrated in Fig. \ref{Wgt}. Using this $\mathbf{W}$ we generate the $L \times K$ observations matrix $\mathbf{Y}$ via the LMM in Eq. (\ref{multiplepixelsmodel}), where the noise matrix $\mathbf{E}$ is Gaussian i.i.d. and SNR=35dB. 

Fig. \ref{fig:exp1a} shows the merits of the proposed IPSpLRU and ADSpLRU algorithms. Specifically, it appears that the concurrent exploitation of sparsity and low-rankness leads to significantly more accurate abundance matrix estimates, as compared to their single constraint counterparts, namely, IPLRU, IPSpU and ADLRU, ADSpU respectively. This is clearly seen in terms of the RMSE, as well as from a careful visual inspection of both the recovered abundance matrices and their residuals  with the true abundace matrix (i.e. $|\hat{\mathbf{W}} - \mathbf{W}|$), depicted in pair from Fig. \ref{wipsplru} - Fig. \ref{iplr}.
%
\begin{figure*}[!t]
\centering
\begin{tabular}{c c c  c}
\subfloat[\small{sparsity level=100\%, rank=1, IPSpLRU}]{\includegraphics[width=0.25\textwidth,height=0.2\textwidth]{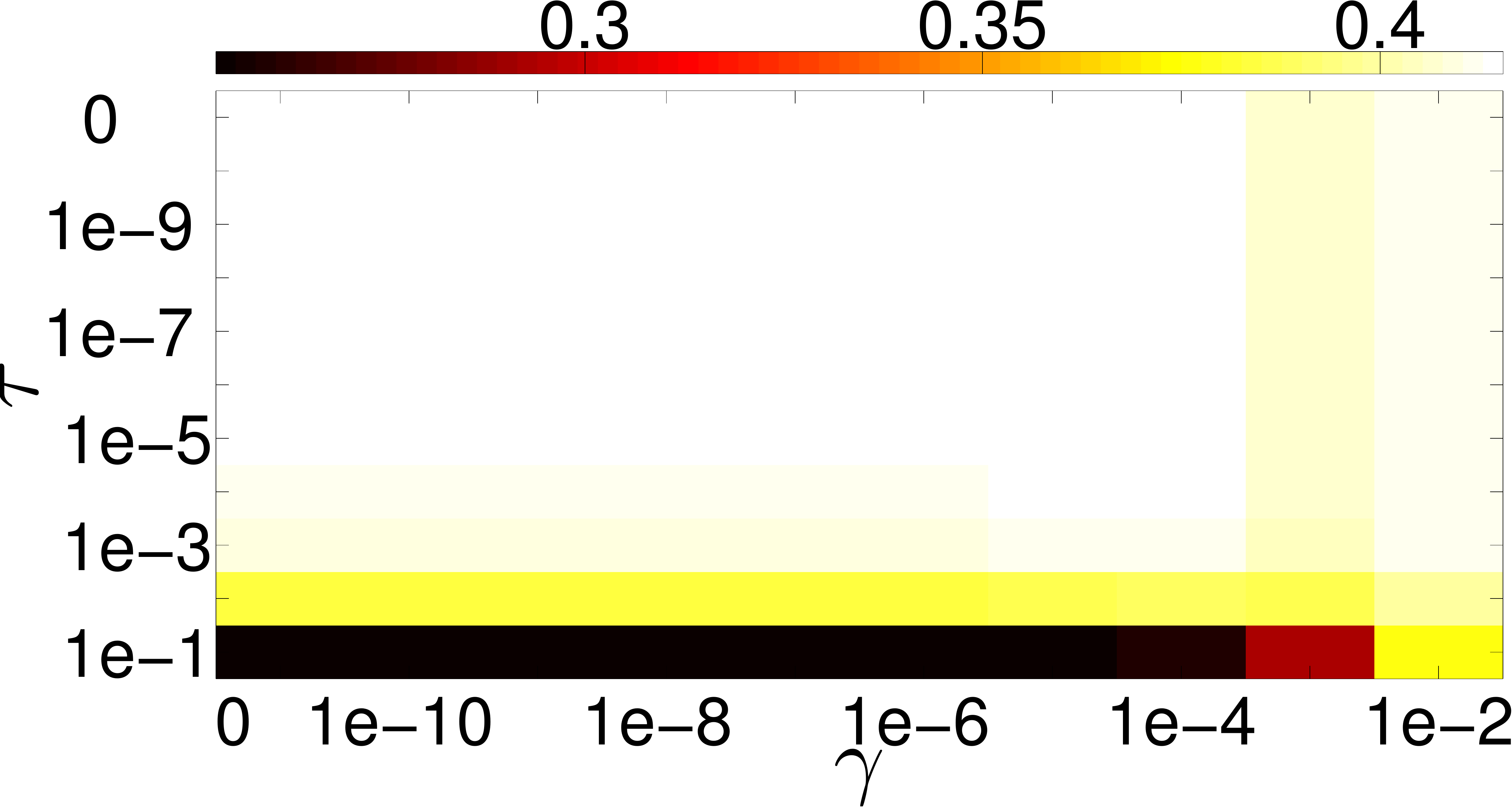}\label{4a}} 
   & \subfloat[\small{sparsity level=10\%, rank=4, IPSpLRU}]{\includegraphics[width=0.25\textwidth,height=0.2\textwidth]{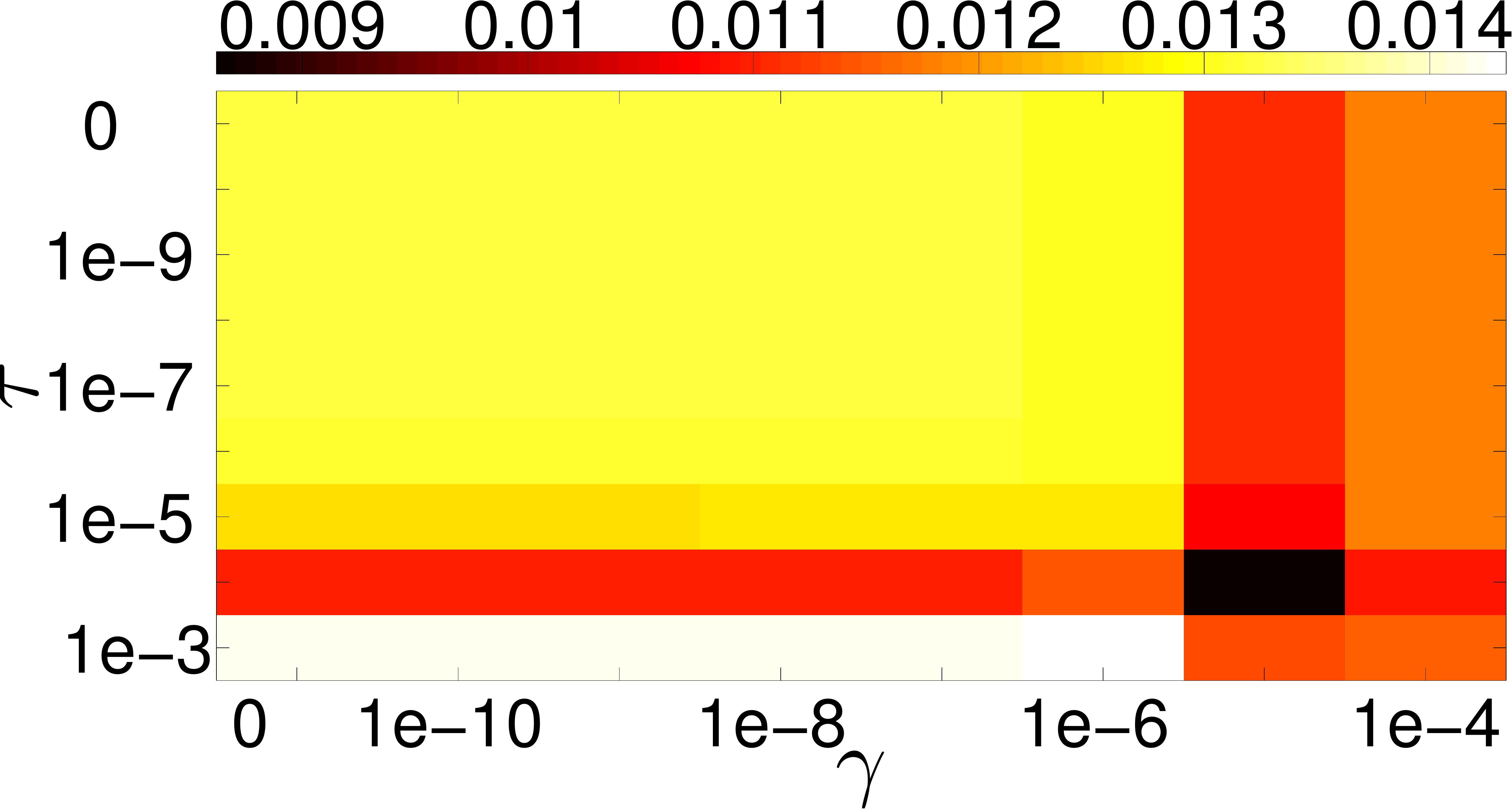}\label{4b}}
    &  \subfloat[\small{sparsity level=20\%, rank=5, IPSpLRU}]{\includegraphics[width=0.25\textwidth,height=0.2\textwidth]{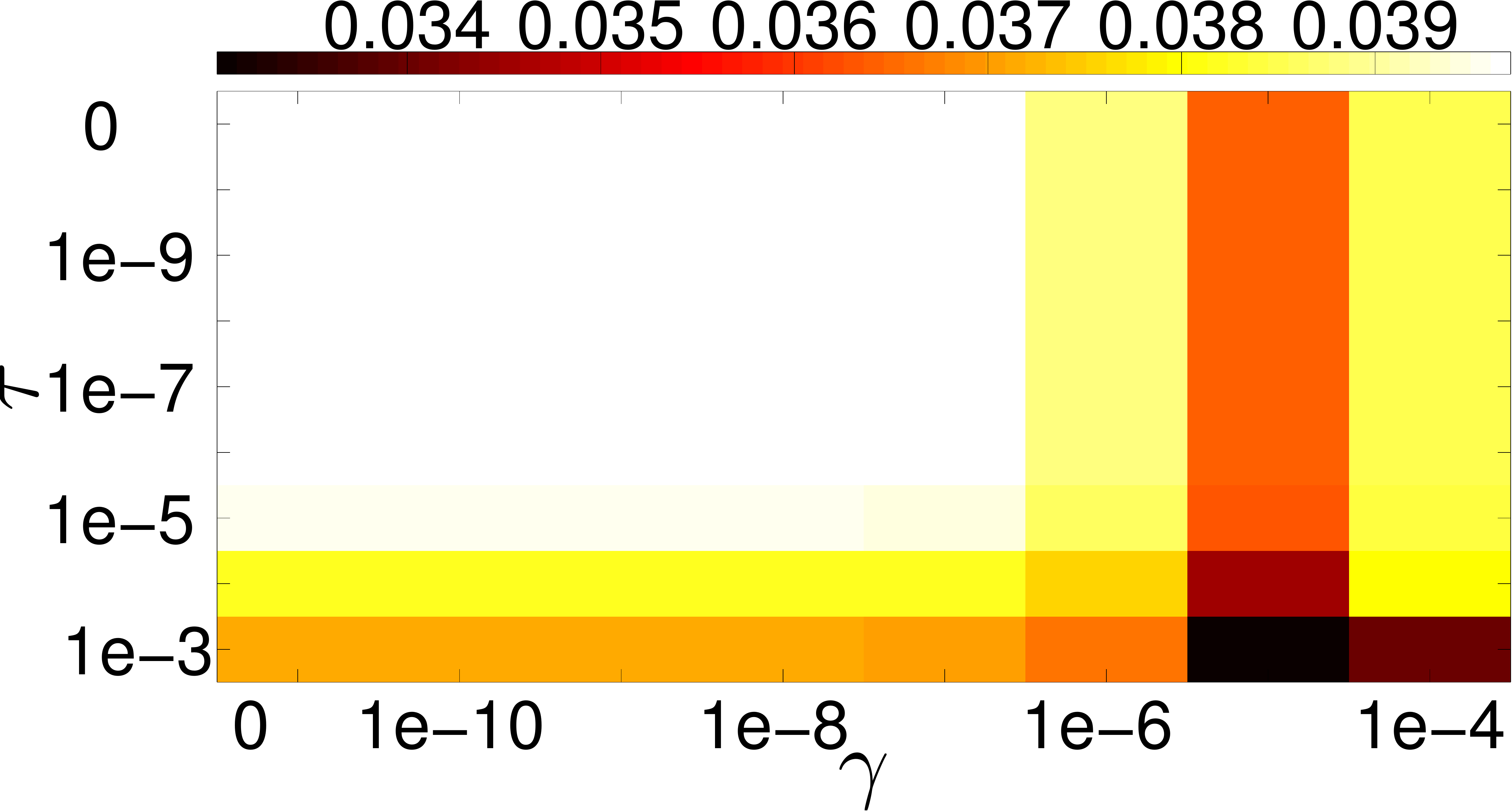}\label{4c}}
       &   \subfloat[\small{sparsity level=10\%, rank=9, IPSpLRU}]{\includegraphics[width=0.25\textwidth,height=0.2\textwidth]{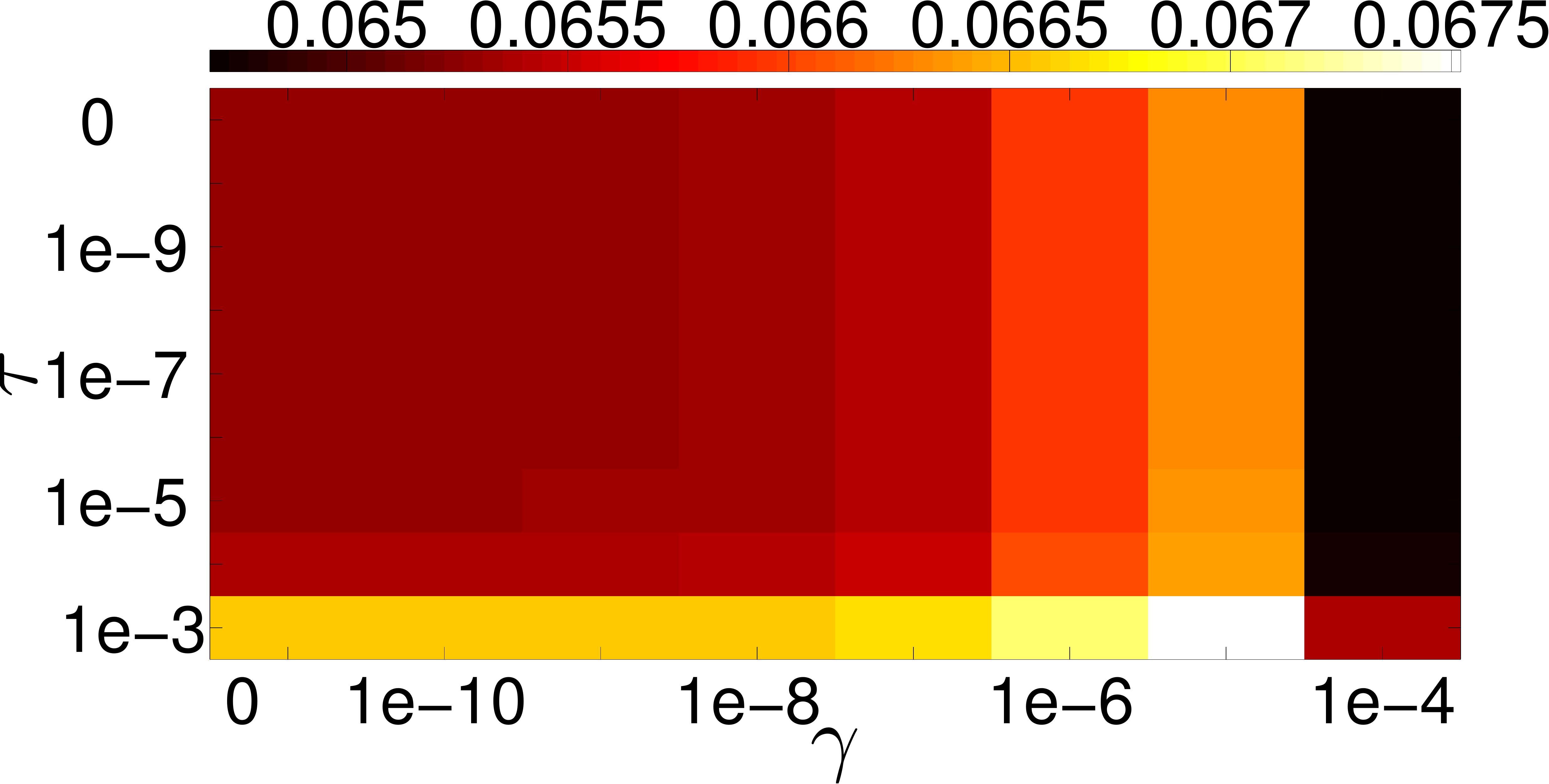}\label{4d}} \\
\subfloat[\small{sparsity level=100\%, rank=1, ADSpLRU}]{\includegraphics[width=0.25\textwidth,height=0.2\textwidth]{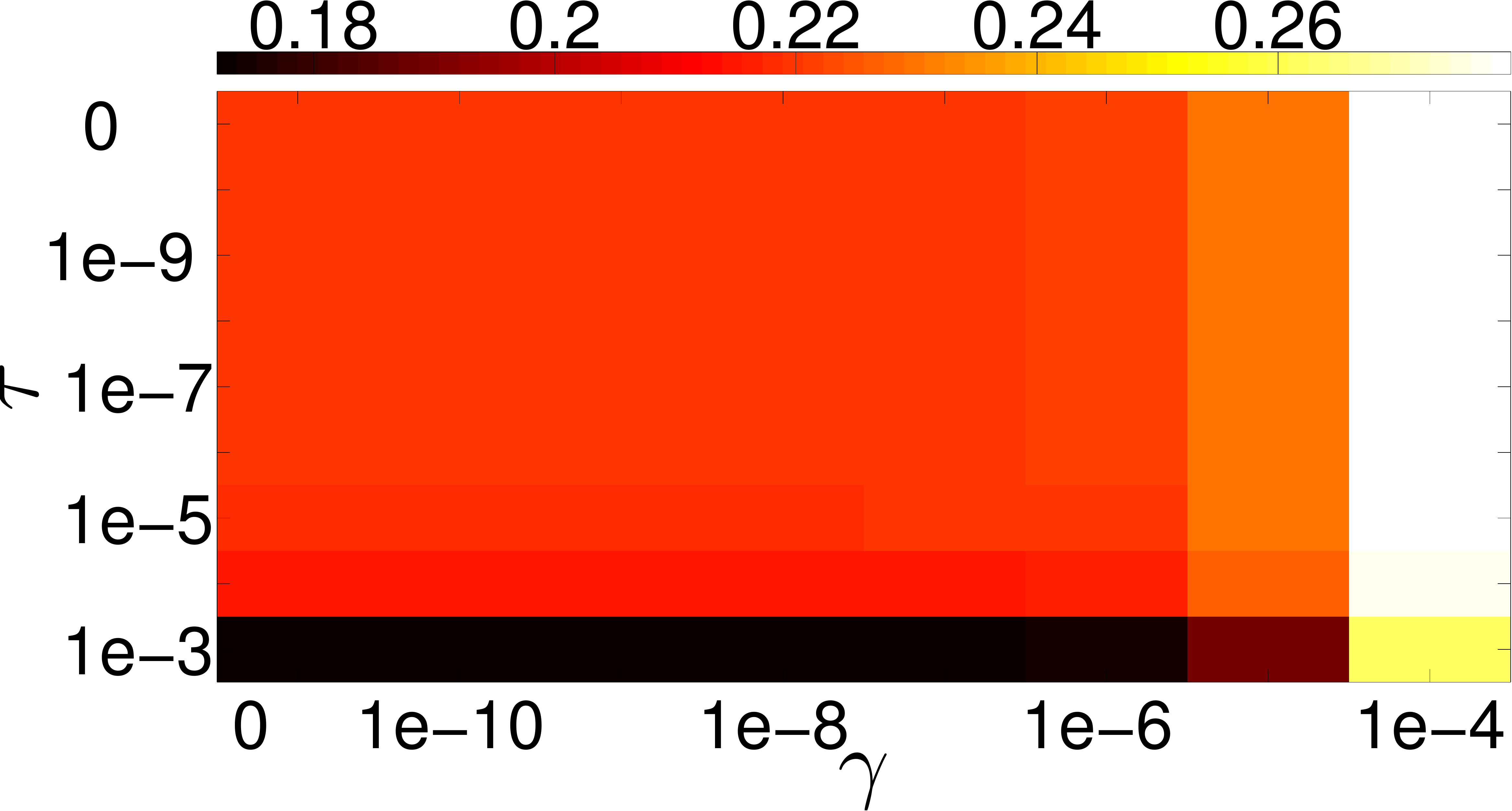}\label{4e}} 
   & \subfloat[\small{sparsity level=10\%, rank=4, ADSpLRU}]{\includegraphics[width=0.25\textwidth,height=0.2\textwidth]{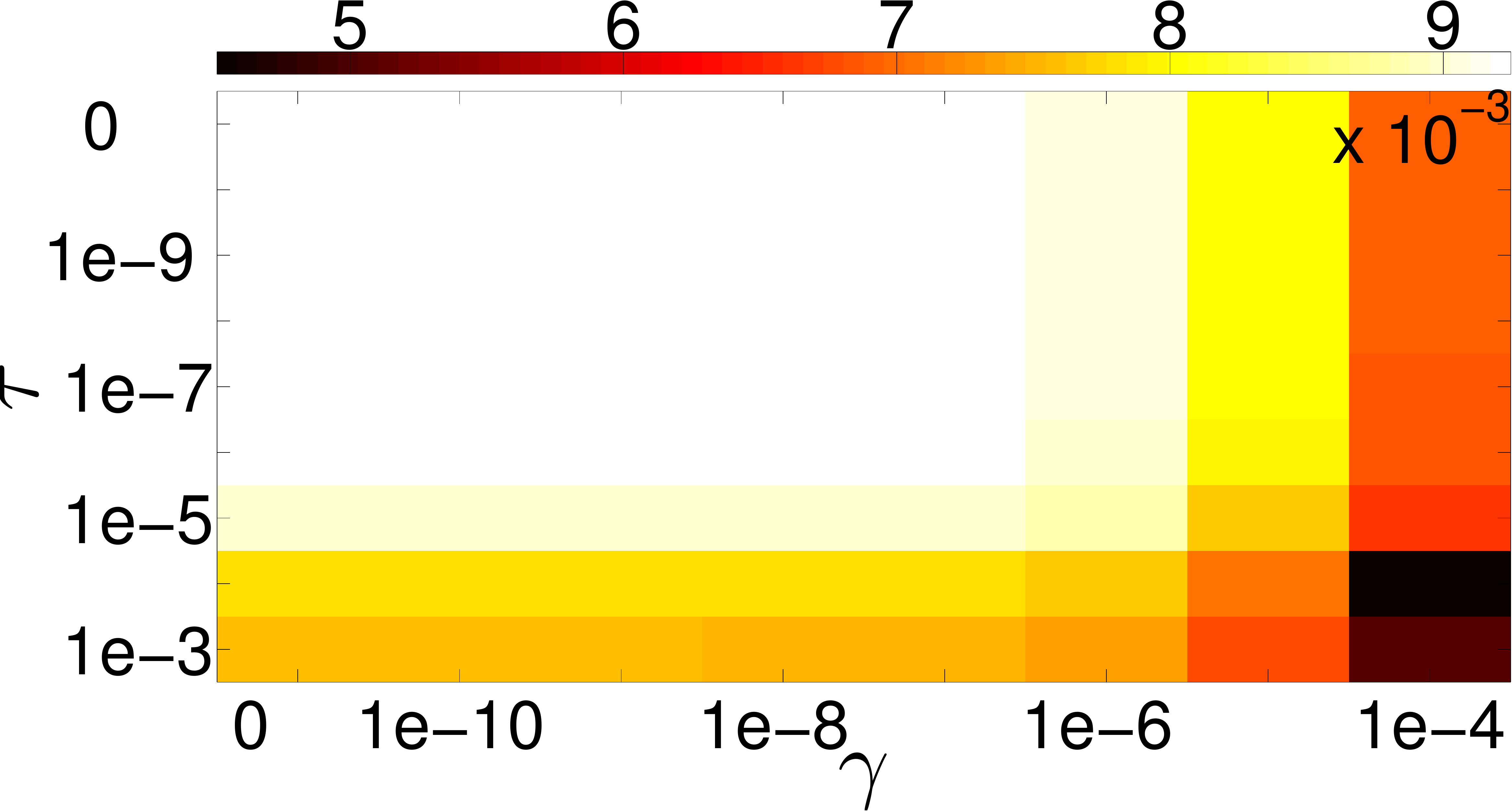}\label{4f}}
    &  \subfloat[\small{sparsity level=20\%, rank=5, ADSpLRU}]{\includegraphics[width=0.25\textwidth,height=0.2\textwidth]{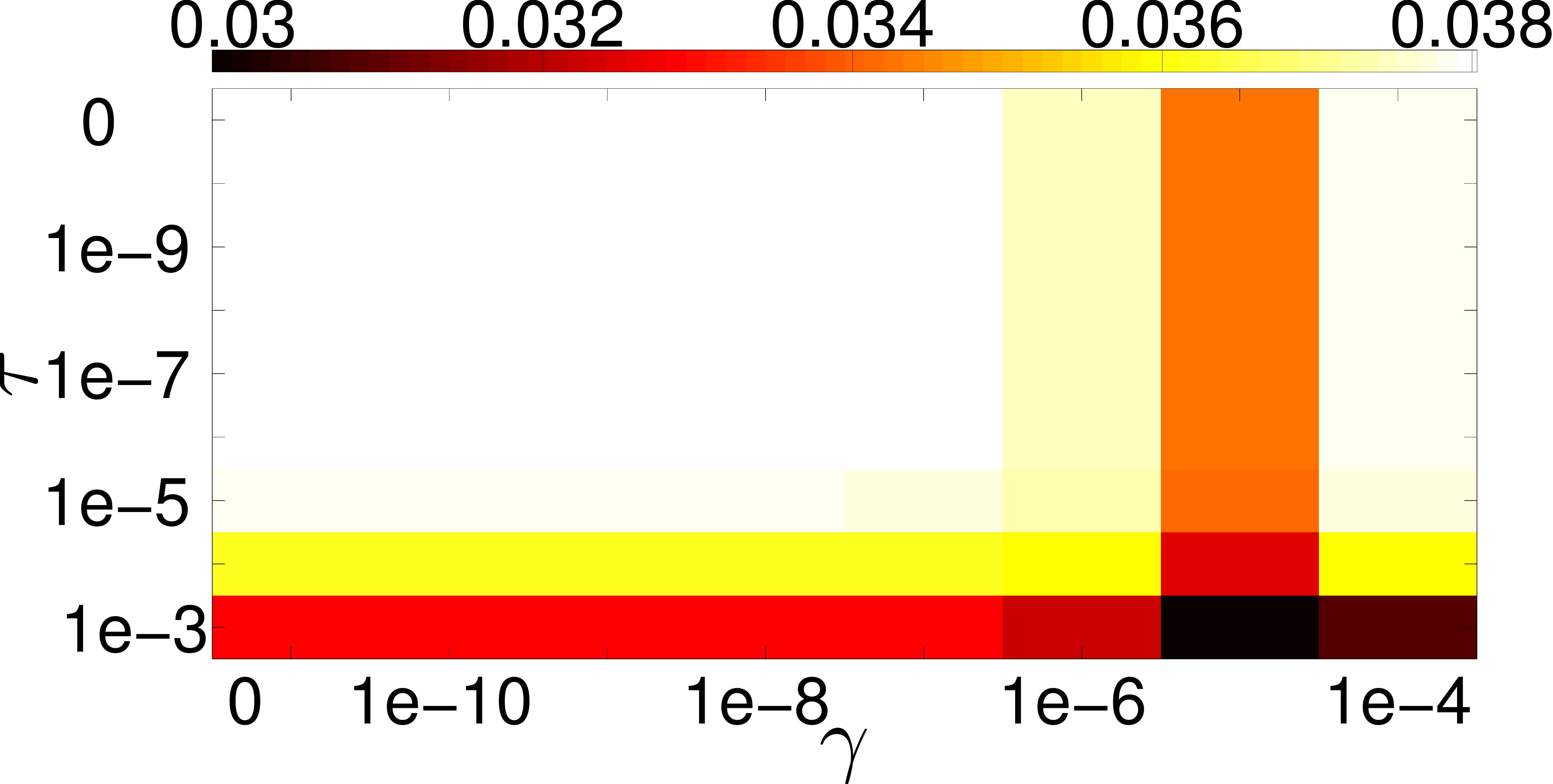}\label{4g}}
       &   \subfloat[\small{sparsity level=10\%, rank=9, ADSpLRU}]{\includegraphics[width=0.25\textwidth,height=0.2\textwidth]{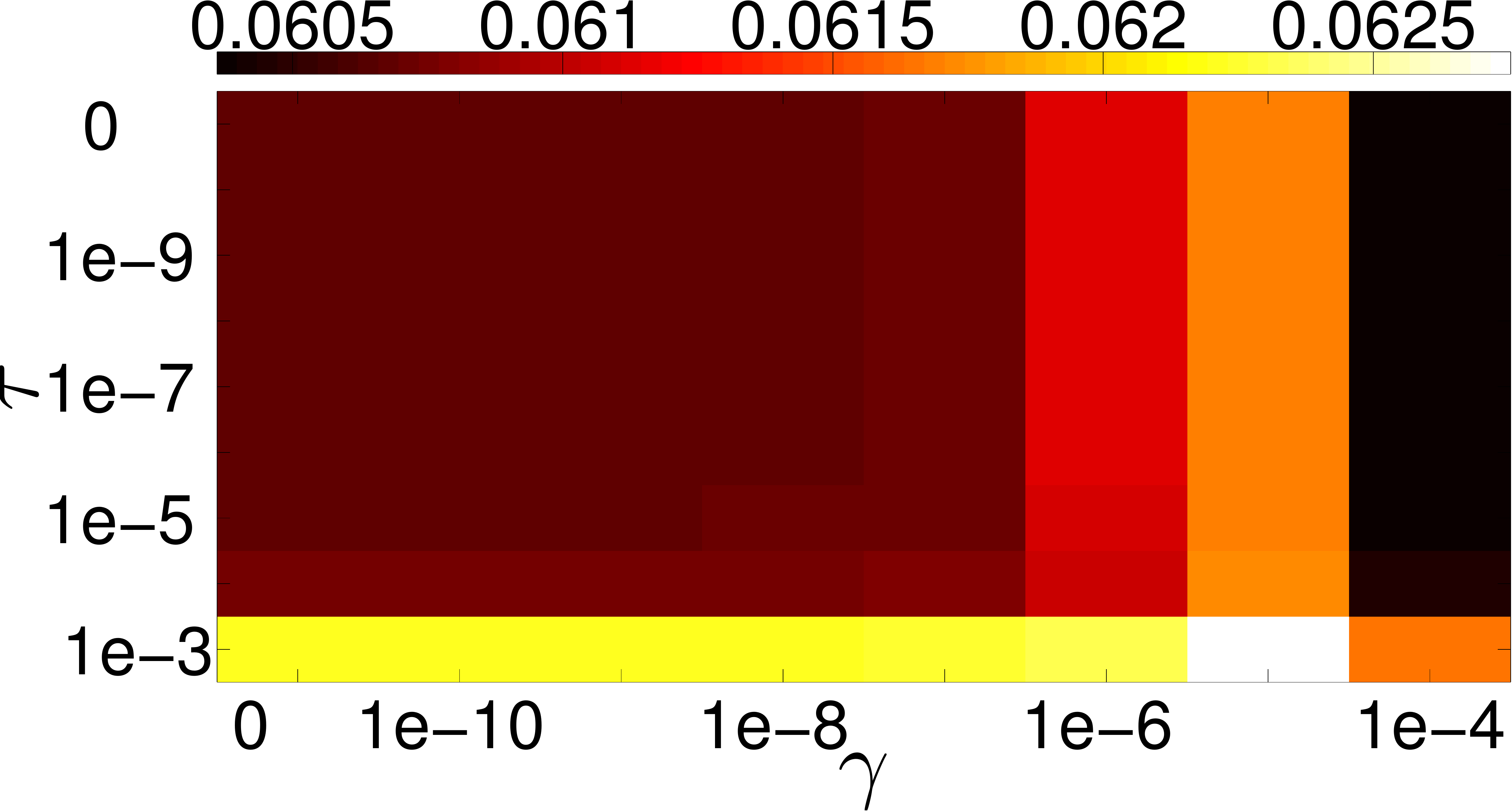}\label{4h}} 
\end{tabular}
\caption{RMSE as a function of the low-rankness and the sparsity regularization parameters $\tau$ and $\gamma$, respectively.}
\label{fig:exp1b}
\end{figure*}
\subsubsection{The key role of the parameters $\gamma,\tau$}
As explained earlier, parameters $\gamma \text{ and } \tau$  control the imposition of sparsity and low-rankness, respectively, on the abundance matrix $\mathbf{W}$. Herein, we unveil the dependency of the optimal (with respect to RMSE minimization) set of these parameters on the inherent structure of the sought abundance matrix. In this vein, five different types of abundance matrices are generated, each reflecting a specific combination of rank and sparsity level. Next, $K=9$ linearly mixed pixels are produced, corrupted with Gaussian i.i.d. noise and SNR=35dB. A number of 100 independent realizations is run for each of the five experiments, and the average RMSE is demonstrated as a function of $\tau$ and $\gamma$. As shown in Fig. \ref{fig:exp1b}, in the first case (Figs. \ref{4a} and \ref{4e}), which corresponds to solely low-rank abundance matrices (without any presence of sparsity), the sparsity promoting parameter $\gamma$ does not affect the estimation accuracy. In a similar manner, in the fourth experiment (Figs. \ref{4d} and \ref{4h}), where the abundance matrix is considered full-rank and sparse, the low-rank promoting parameter has no impact on the estimation performance. Notably, in the other two cases (columns 2 and 3) where both sparse and low-rank abundance matrices are considered, RMSE is minimized for non-zero values of both $\tau$ and $\gamma$. Such a result is consistent with the fundamental premise of our algorithms, which is the improvement in the abundance matrix estimation  by simultaneously exploiting  sparsity and low-rankness. 

Moreover, the above results indicate that the optimal choice of $\tau,\gamma$ depends on the particular structure (sparse and/or low-rank) of the abundance matrix. Thus, a proper selection of these parameters shall involve fine-tuning schemes, which are commonplace when it comes to algorithms dealing with regularized inverse problems.

%
\begin{figure*}[t!]
\centering
\includegraphics[width=\textwidth,height=0.5\textwidth]{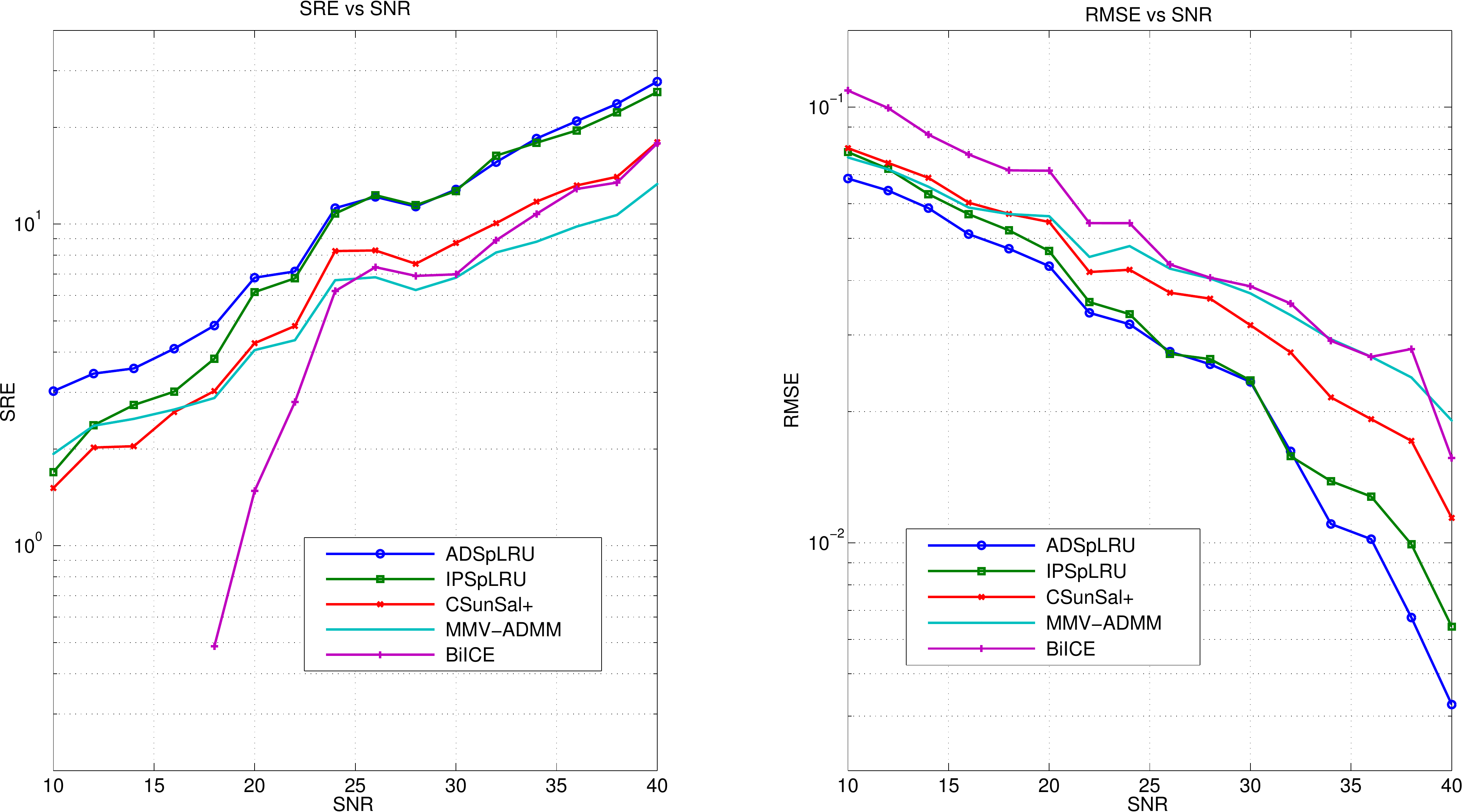}
\caption{Robustness to white noise (SRE \& RMSE).}
\label{fig:exp3a}
\end{figure*}
\begin{figure*}
\centering
\includegraphics[width=\textwidth,height=0.5\textwidth]{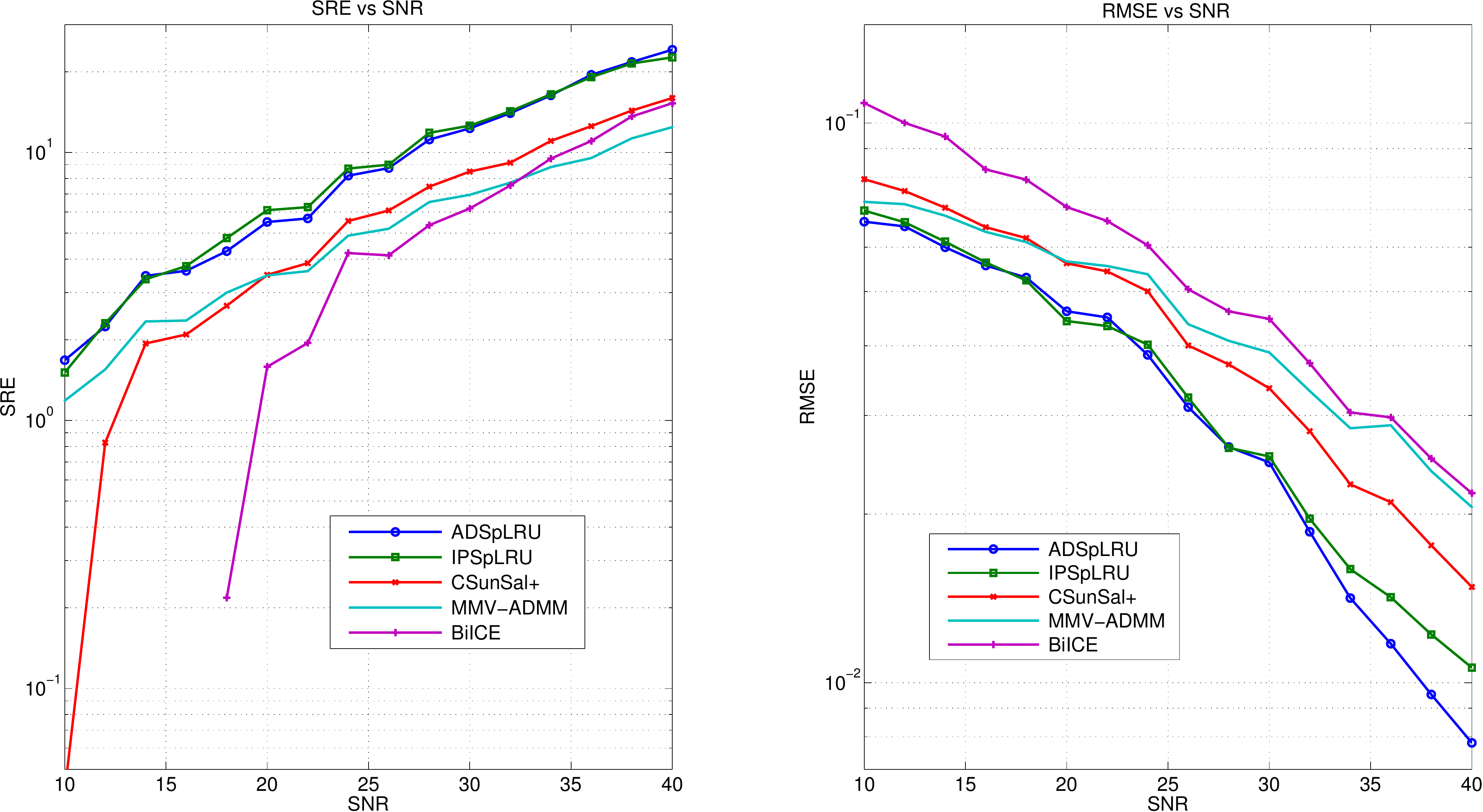}
\caption{Robustness to colored noise (SRE \& RMSE).}
\label{fig:exp3b}
\end{figure*}
\subsubsection{Robustness to noise}\label{expb}
In this experiment we aim at exhibiting the robustness of the proposed algorithms to white and correlated noise corruption. To this end, we stick with a specific simultaneously sparse and low-rank abundance matrix $\mathbf{W}$ of sparsity level 20\% and rank 3. Based on this $\mathbf{W}$, $K=9$ linearly mixed pixels are generated, in the same way as described above. Then, depending on the case, white or colored Gaussian noise contaminates the data. 16 SNR values are considered ranging from 10 to 40 dB, while 100 realizations are run for each SNR value, and the mean of the RMSE and SRE metrics is calculated.
\begin{itemize}
\item {\it White Gaussian Noise:}
Fig. \ref{fig:exp3a} shows the RMSE and SRE curves obtained for the proposed IPSpLRU, ADSpLRU and the three competing algorithms, namely, CSUnSAL, MMV-ADMM and BiICE. It is easily seen that both IPSpLRU and ADSpLRU attain remarkably better results comparing to CSUnSAL, MMV-ADMM and BiICE in all the examined SNR values. Additionally, we note that ADSpLRU performs slightly better as compared to IPSpLRU, especially for SNR values greater than $32$dB. The price to be paid is that the computational complexity  per iteration of ADSpLRU is higher than that of IPSpLRU. It is hence shown that sparse and low-rank methods are robust to different levels of white noise. At the same time, IPSpLRU and ADSpLRU outperform the sparse only CSUnSAL and BiICE algorithms and the joint-sparse MMV-ADMM algorithm, provided that both sparsity and low-rankness characterizes the abundance matrix. 
\item
{\it Colored Gaussian Noise:}
Actually, in real hyperspectral images the noise that corrupts the data is rather structured than white. Thus, to assess the behavior of the proposed methods in such realistic conditions, we simulate correlated Gaussian noise that adds up to the linearly mixed pixels. Fig. \ref{fig:exp3b}  illustrates the effectiveness of the tested algorithms in terms of RMSE and SRE, for different SNR values. Therein as well, we can see that IPSpLRU and ADSpLRU achieve better results than their competing algorithms in the whole range of the examined SNRs. Furthermore, ADSpLRU performs better for high SNR values ($>32\mathrm{dB}$), as compared to IPSpLRU. As a result, the robustness of our proposed methods is also corroborated in the presence of correlated noise with different magnitude.
\end{itemize}
\begin{figure*}
\centering
\begin{tabular}{c c}
\begin{adjustbox}{valign=c}
\subfloat[Synthetic Image, 16 blocks of size $10\times10$ pixels each.]
{\label{synth_image}\includegraphics[width=0.5\textwidth,height=0.5\textwidth]{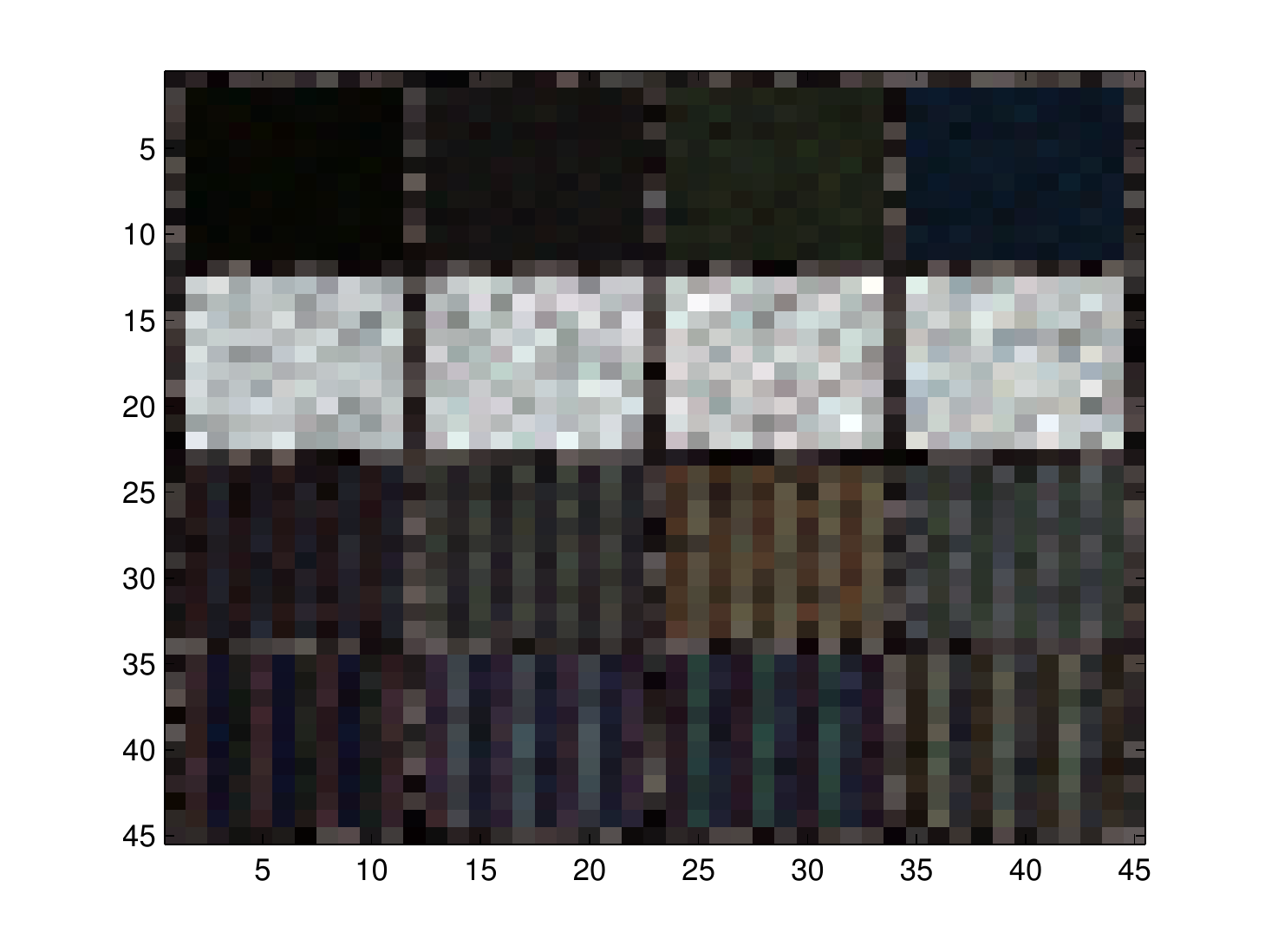}}
\end{adjustbox}  & 
\begin{tabular}{c}
\subfloat[Structure of $\mathbf{W}$ in each block of the synthetic image, each cell contains the pair : (sparsity-level\%,$\text{ rank}(\mathbf{W}).$)\label{synth_im_structure}]{\begin{adjustbox}{valign=c}
\vspace{-2cm}
\resizebox{0.5\columnwidth}{!}{
\begin{tabular}{|c || c | c | c | c |}
\hline
\multirow{2}{*}{row} &  \multicolumn{4}{c|}{column} \\ \cline{2-5} 
& $1^{st}$  & $2^{nd}$ & $3^{rd}$ & $4^{th}$ \\
\hline
\text{joint sparse} - $1^{st}$ & $(4,1)$ & $(8,2)$ & $(12,3)$ & $(16,4)$\\
\hline
\text{low-rank} - $2^{nd}$ & $(100,1)$ & $(100,2)$ & $(100,3)$ & $(100,4)$ \\
\hline
\text{sparse \& low-rank} - $3^{rd}$& $(4,2)$ & $(8,2)$ & $(12,2)$ & $(16,2)$\\
\hline
\text{sparse \& low-rank} - $4^{th}$& $(4,3)$ & $(8,3)$ & $(12,3)$  & $(16,3)$\\
\hline
\end{tabular}}
\end{adjustbox}}\\
\subfloat[RMSE ($10^{-2}$) and SRE (dB) results on synthetic image for each row.]{
\resizebox{0.5\columnwidth}{!}{
\label{synth_results}
\begin{tabular}{|c||c | c |c|c|c|c|c|c|}
\hline
\multirow{2}{*}{\bf{Algorithm}}& \multicolumn{2}{c|}{{$1^{st}$} row} &\multicolumn{2}{c|}{$2^{nd}$ row} &\multicolumn{2}{c|}{$3^{rd}$ row} &\multicolumn{2}{c|}{$4^{th}$ row}\\\cline{2-9}
 & RMSE  &  SRE& RMSE &  SRE& RMSE  & SRE & RMSE &  SRE  \\
\hline
 ADSpLRU & 0.009 & 28.96 & 0.078 & 16.62 & 0.032 & 18.71 & 0.029 & 19.62 \\\hline
 IPSpLRU & 0.008 & 28.39 & 0.081 & 16.41 & 0.026 & 21.01 & 0.030 & 19.81 \\\hline
 CSunSAL & 0.026 & 19.81 & 0.117 & 12.39 & 0.052 & 13.88 & 0.047 & 14.99  \\\hline
 MMV-ADMM & 0.030 & 18.00 & 0.105 & 12.99 & 0.061 & 12.32 & 0.056 & 13.16 \\\hline
 BiICE & 0.028 & 21.71 & 0.263 & 6.72 & 0.043 & 17.83 & 0.060 & 15.81 \\\hline
\end{tabular}}}
\end{tabular}
\end{tabular}
\caption{Structure of the synthetic image and results.} 
\vspace{-0.4cm}
\end{figure*}
\begin{figure*}[t]
\begin{tabular}{c c}
 \centering \subfloat[5th PC of the Salinas Valley scene.]{\label{salinas5pc}\includegraphics[width=0.5\textwidth,height=0.5\textwidth]{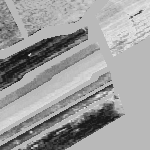}} \vspace{-8cm}  &
\begin{tabular}{c}
\subfloat[Rough ground truth information for a part of the Salinas valley scene under study.]{\label{salinasgroundtruth}\includegraphics[width=0.5\textwidth,height=0.3\textwidth]{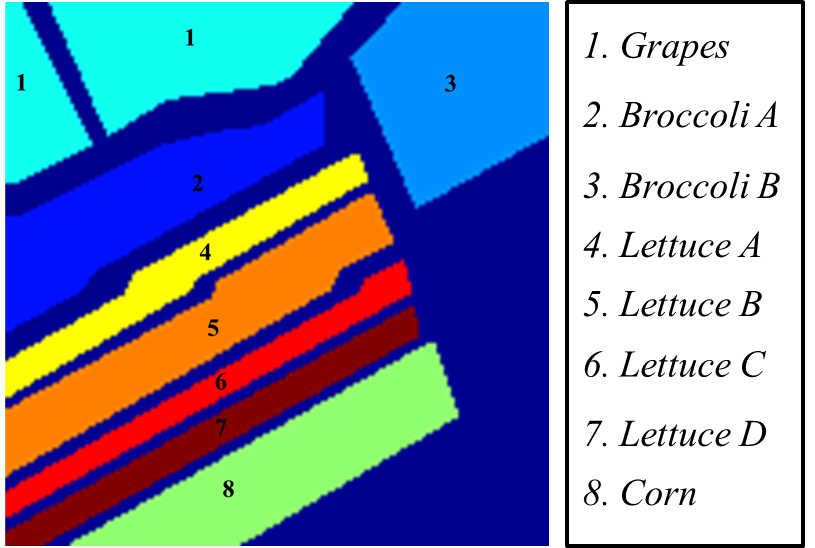}} \\
 \subfloat[Spectral signatures of the 17 endmembers manually selected as pure pixels.]{\label{dictionary}\includegraphics[width=0.5\textwidth,height=0.3\textwidth]{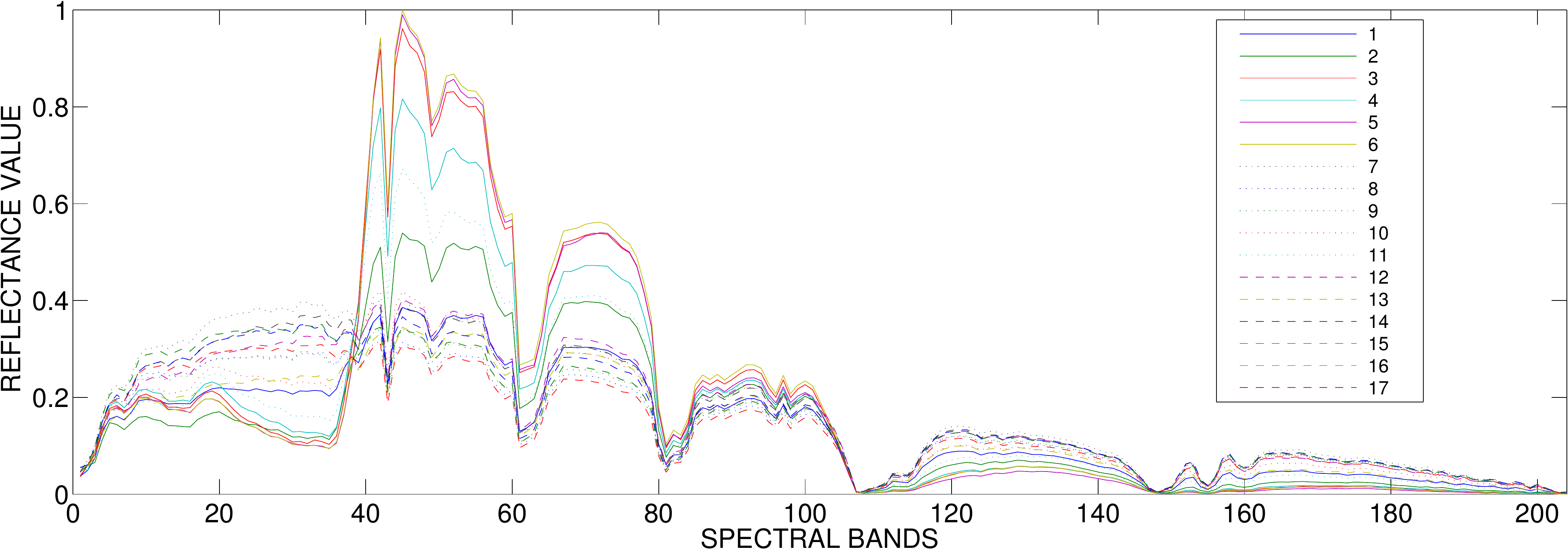}}\vspace{7.5cm}
\end{tabular}
\end{tabular}
\caption{Salinas valley image and endmembers' dictionary}
\label{salinas}
\end{figure*}
\subsubsection{Synthetic Image}\label{expc}
This experiment highlights the effectiveness of the proposed methods in estimating sparse, low-rank or both sparse and low-rank abundance matrices. Focused on this purpose we form a simulated hyperspectral image using the linear mixing model (1) and the same above-mentioned endmembers' dictionary $\boldsymbol{\Phi}$. As shown in Fig. \ref{synth_image}, the simulated hyperspectral image consists of 4 rows each consisting of 4 $10\times 10$ blocks of pixels. Each of the ``block rows'' is generated by abundance matrices of a distinct structure. To be more specific, the first row is generated by joint-sparse $\mathbf{W}$'s, the second by solely low-rank $\mathbf{W}$'s, while rows 3 and 4 are produced by simultaneously sparse and low-rank abundance matrices. The pixels in each block correspond to abundance matrices of a particular combination of sparsity level and rank. The detailed description of these structures is depicted in the table of Fig. \ref{synth_im_structure}. The linearly mixed pixels are corrupted by white Gaussian i.i.d. noise such that SNR = 30dB.
%
The table in Fig. \ref{synth_results} contains the obtained RMSE and SRE for all algorithms tested. It is worth  pointing out that our introduced IPSpLRU and ADSpLRU algorithms outperform their rivals, not only in the ``both sparse and low-rank'' rows 3 and 4, but also in rows 1 and 2 that correspond to either sparse only or low-rank only $\mathbf{W}$'s. 	
\begin{figure*}
\begin{center}
\begin{tabular}{ccccc}
grapes & \hspace{0cm} brocolli\_a &\hspace{0cm} brocolli\_b & \hspace{0cm} corn \\
\centering
\begin{subfloat}
\centering
\includegraphics[width=0.2\linewidth]{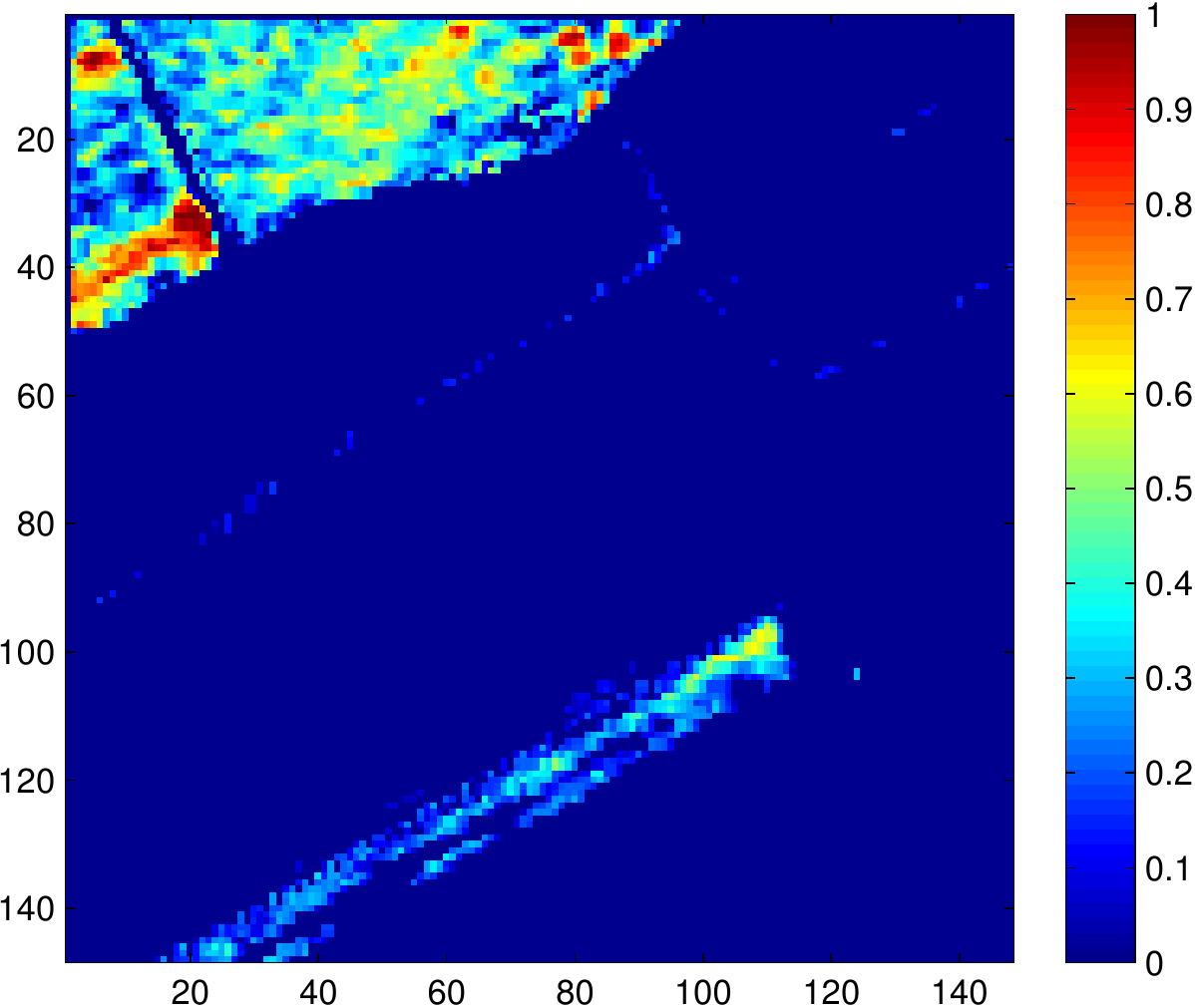}
\end{subfloat} &
\begin{subfloat}
\centering
{\includegraphics[width=0.2\linewidth]{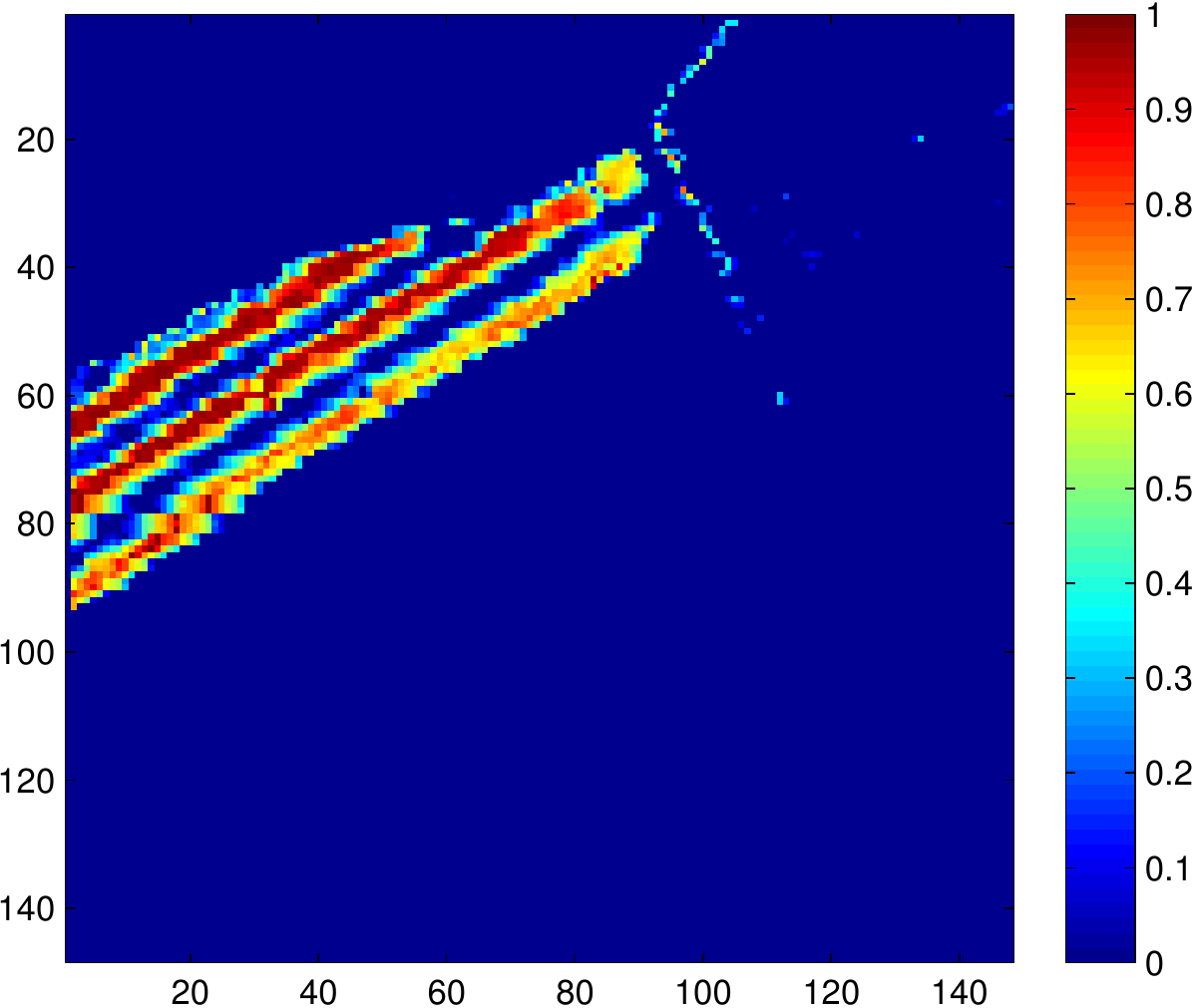}}
\end{subfloat} &
\begin{subfloat}
\centering
{\includegraphics[width=0.2\linewidth]{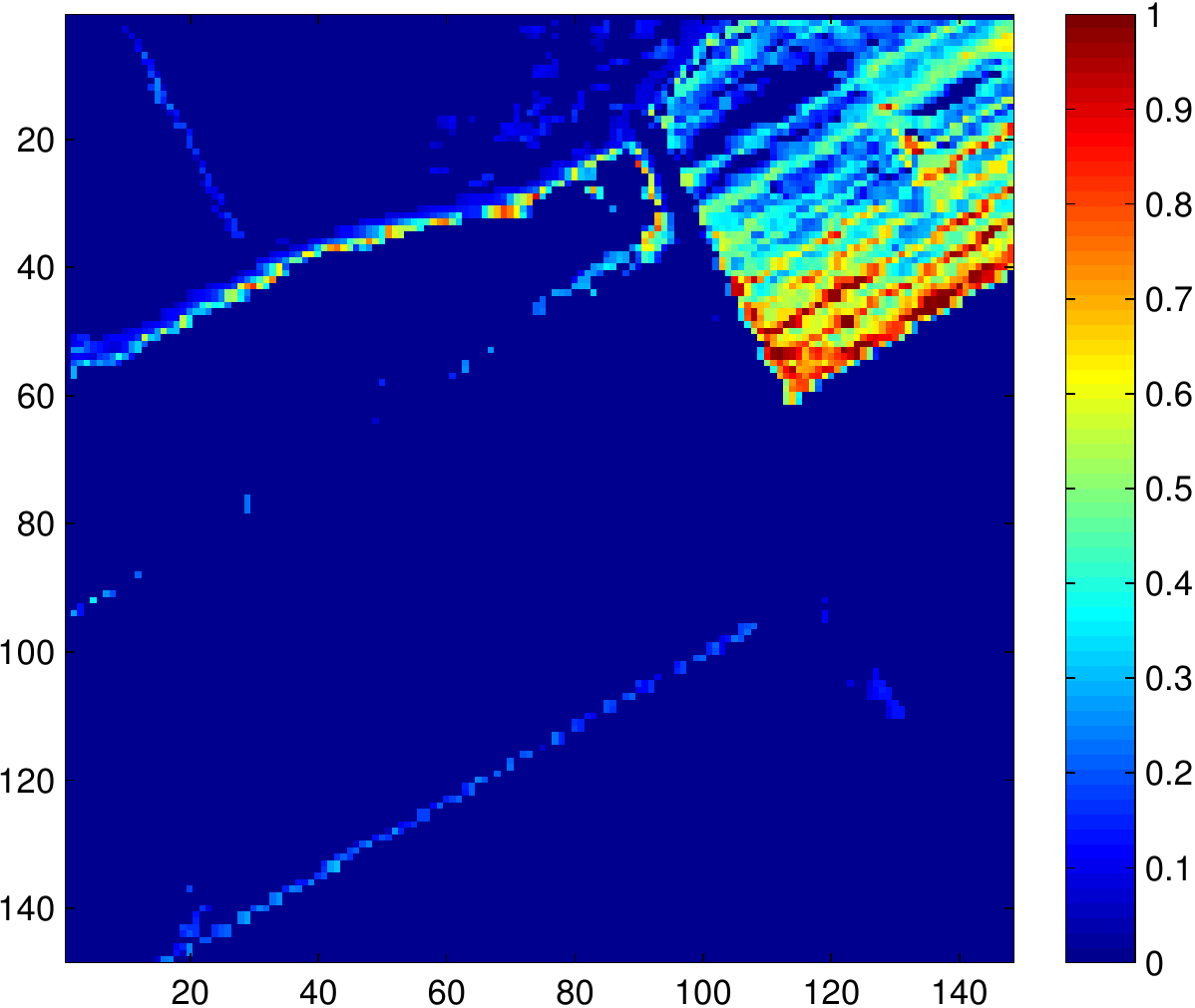}}
\end{subfloat} &
\begin{subfloat}
\centering
{\includegraphics[width=0.2\linewidth]{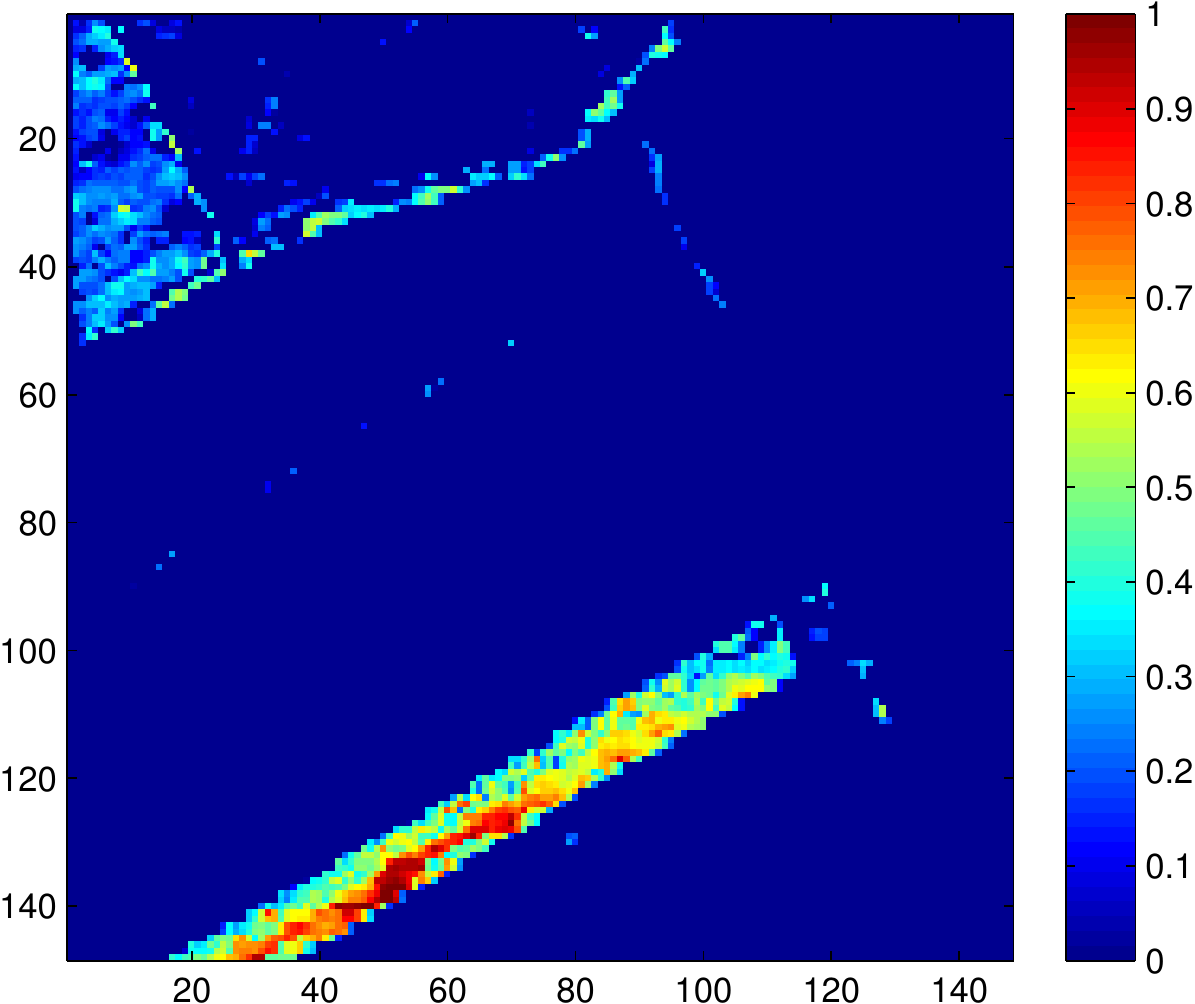}}
\end{subfloat} 
\end{tabular}
\vspace*{-0.3cm}
\caption*{(a) IPSpLRU}
\begin{tabular}{ccccc}
\centering
\begin{subfloat}
\centering
\includegraphics[width=0.2\linewidth]{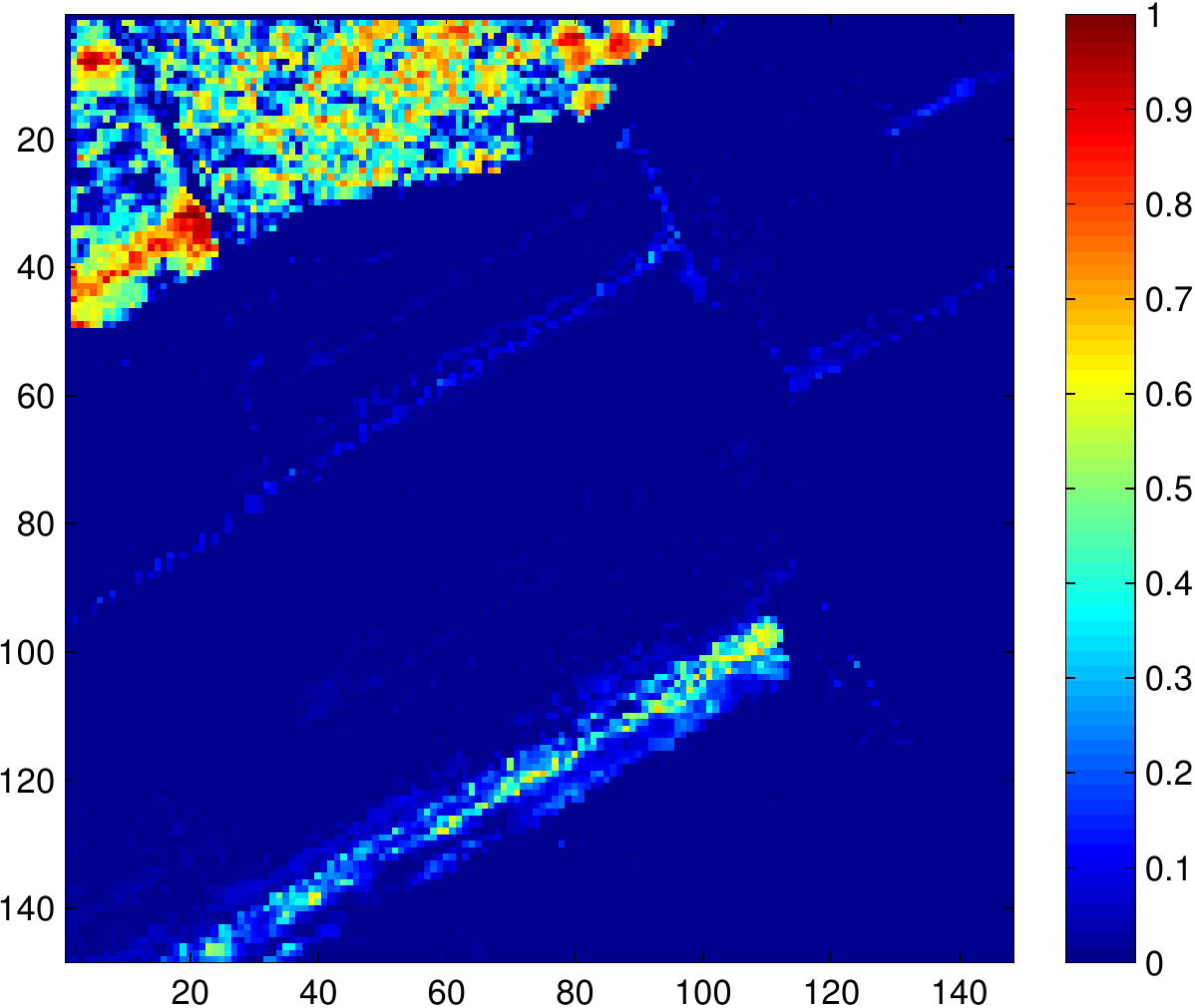}
\end{subfloat} &
\begin{subfloat}
\centering
{\includegraphics[width=0.2\linewidth]{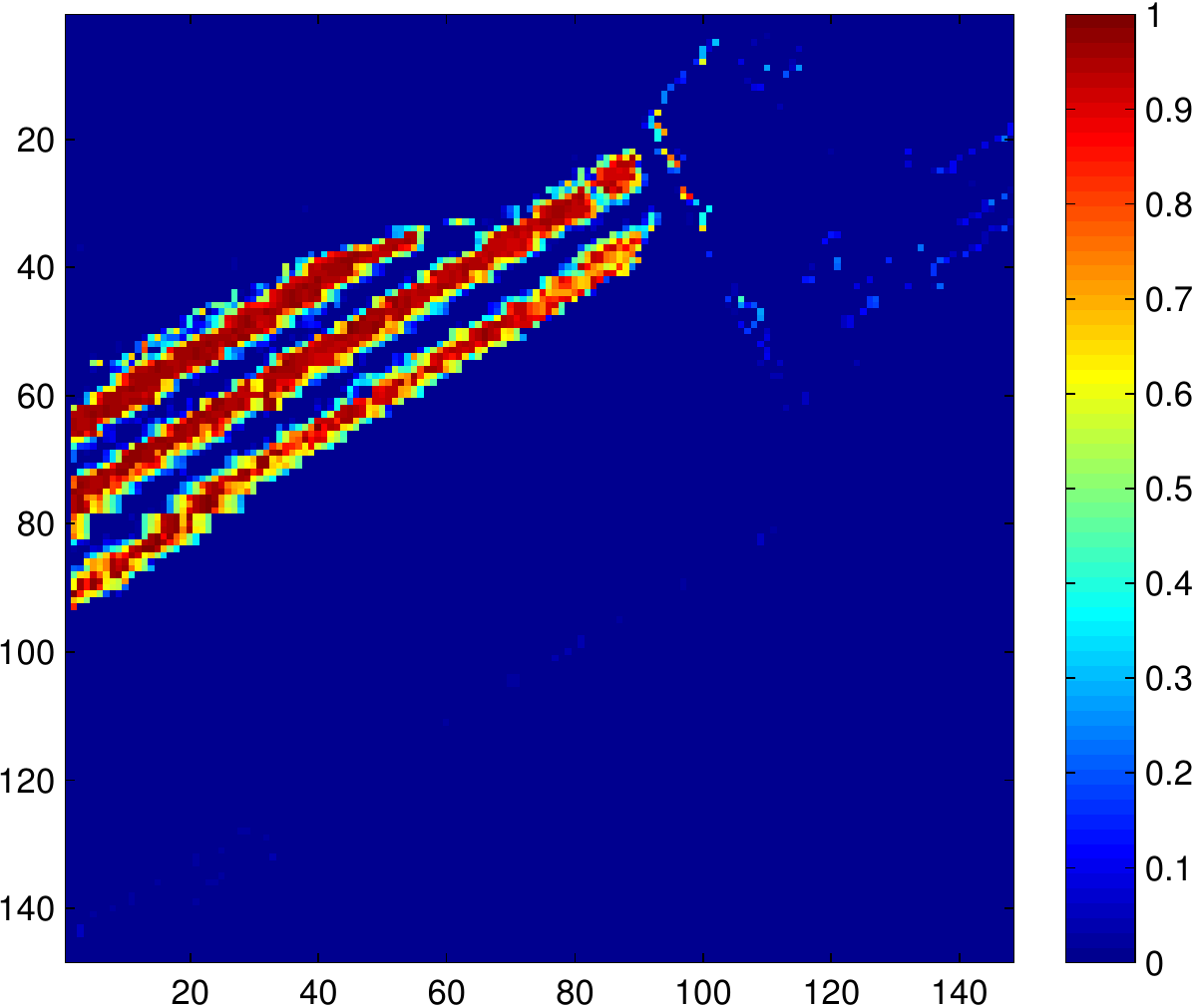}}
\end{subfloat} &
\begin{subfloat}
\centering
{\includegraphics[width=0.2\linewidth]{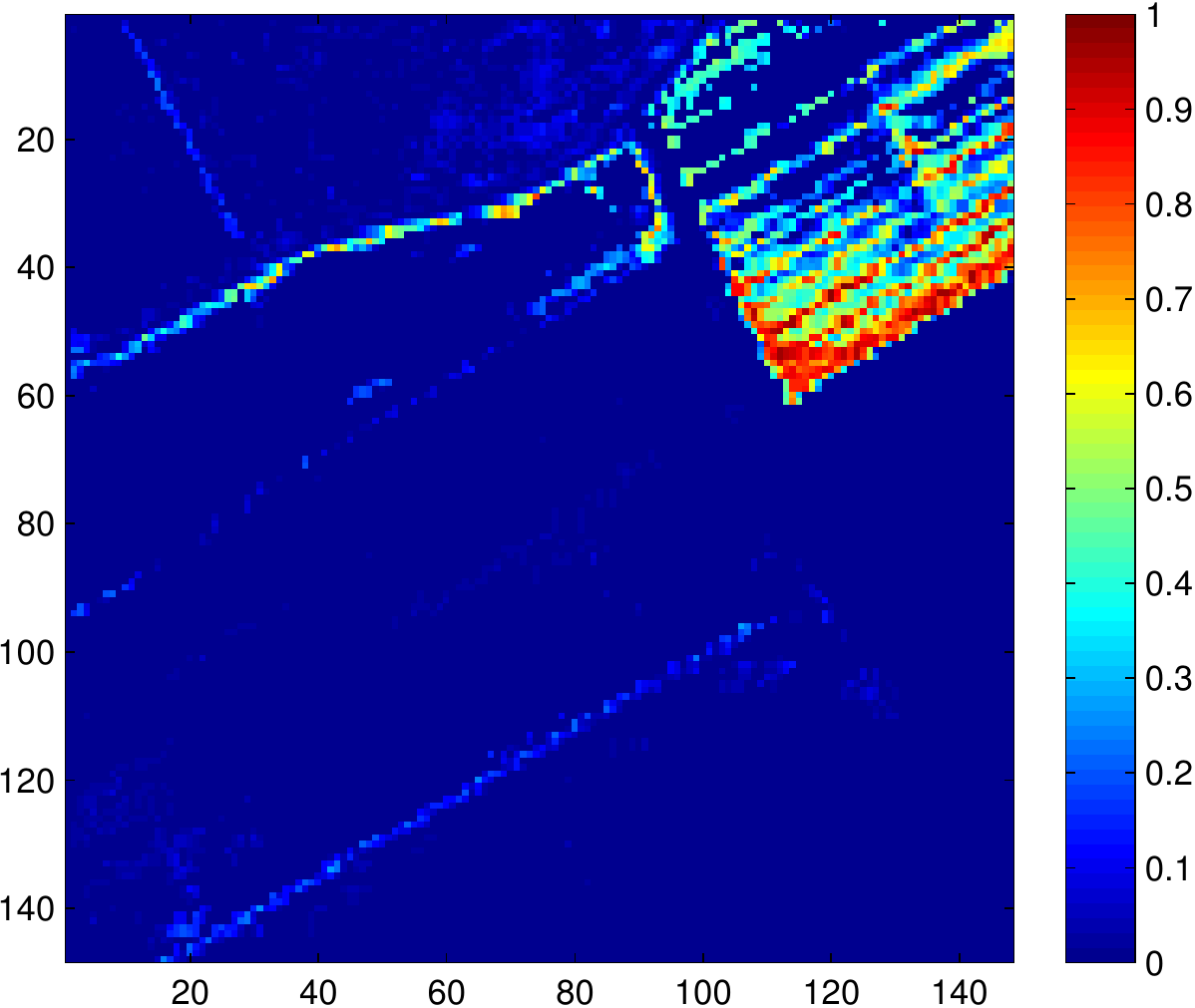}}
\end{subfloat} &
\begin{subfloat}
\centering
{\includegraphics[width=0.2\linewidth]{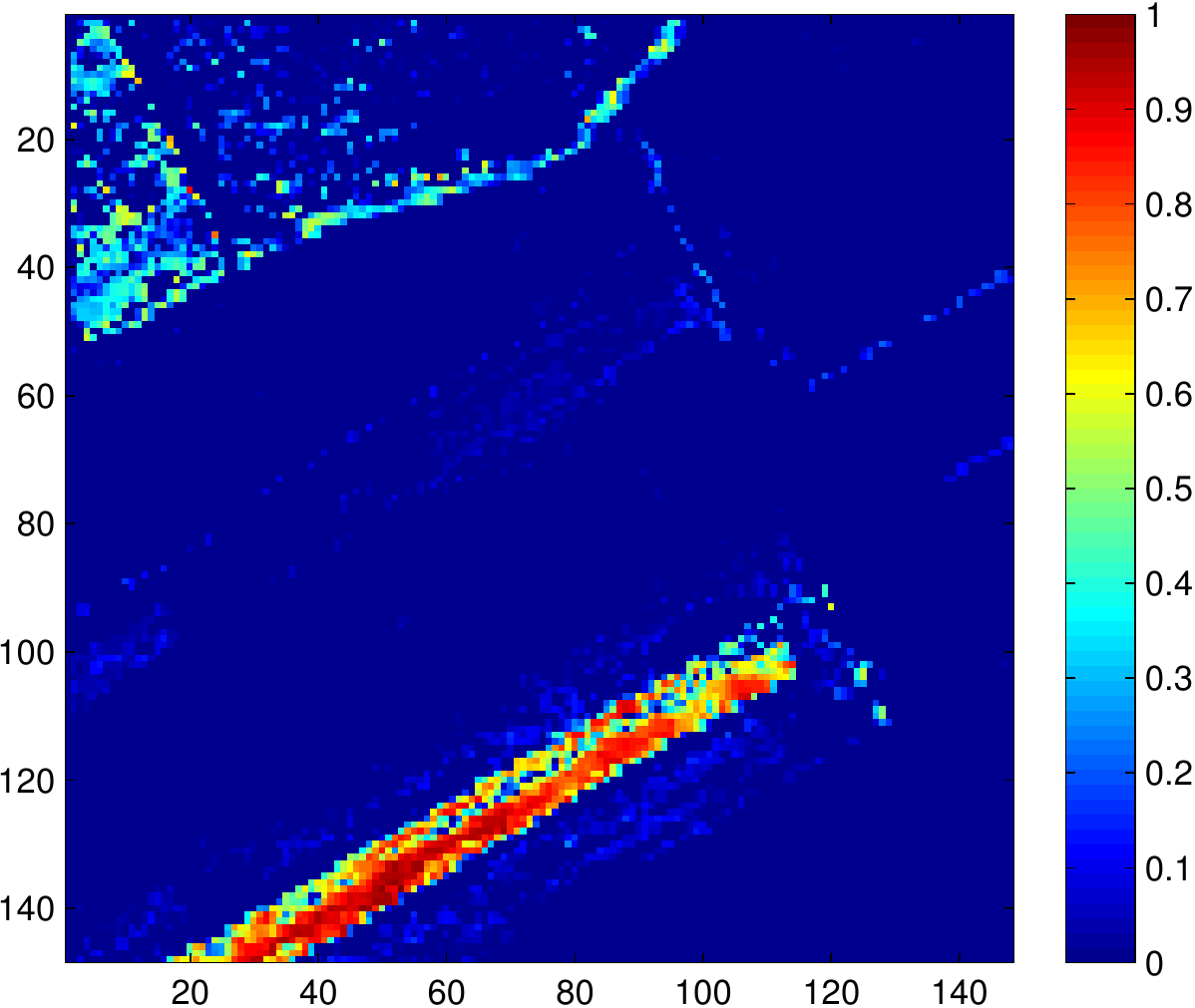}}
\end{subfloat}
\end{tabular} 
\vspace*{-0.3cm}
\caption*{(b) ADSpLRU}
 \begin{tabular}{c c c cc}
\begin{subfloat}
\centering
{{\includegraphics[width=0.2\linewidth]{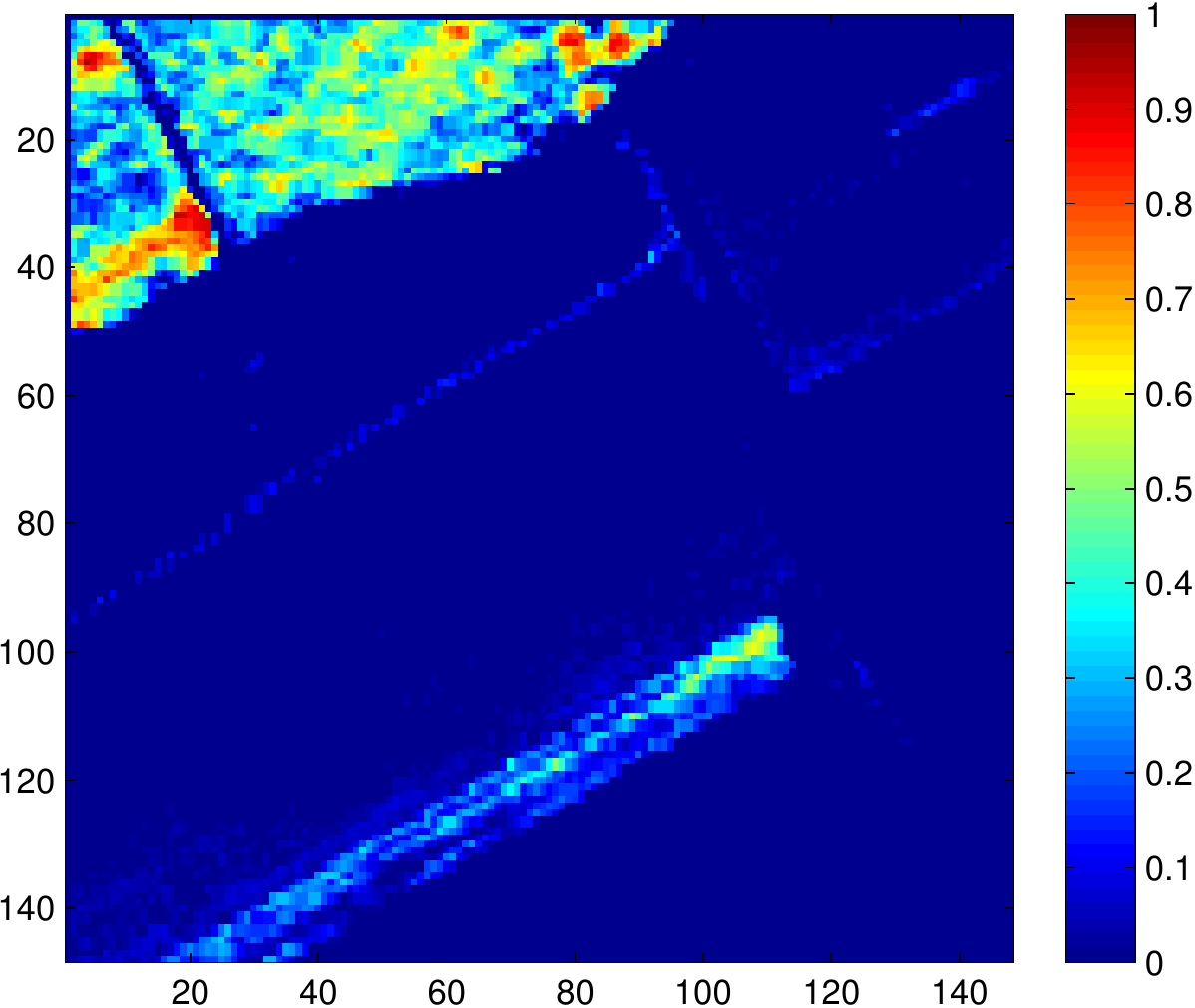}}}
\end{subfloat} 	&
\begin{subfloat}
\centering
{\includegraphics[width=0.2\linewidth]{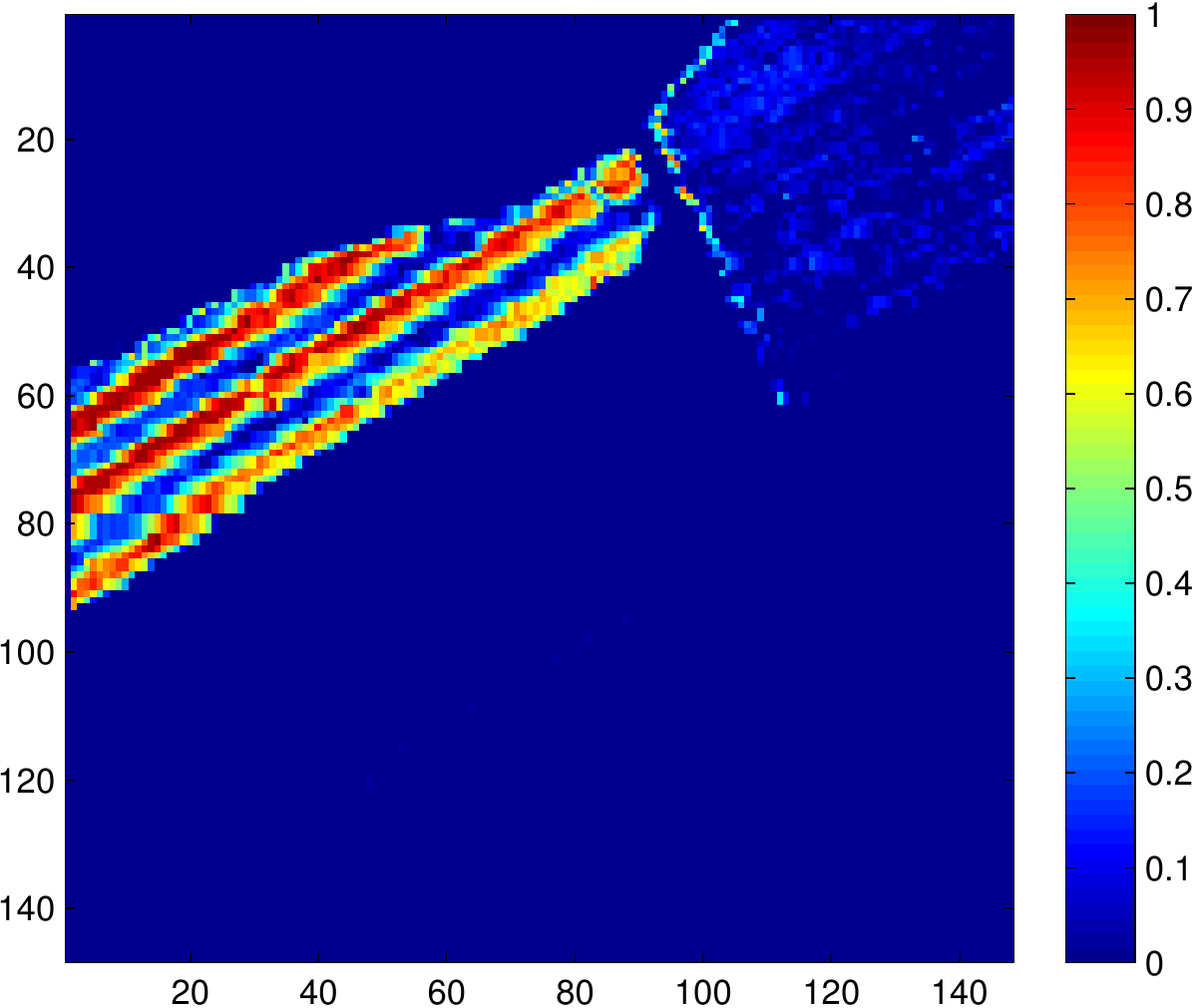}}
\end{subfloat} & 
\begin{subfloat}
\centering
{\includegraphics[width=0.2\linewidth]{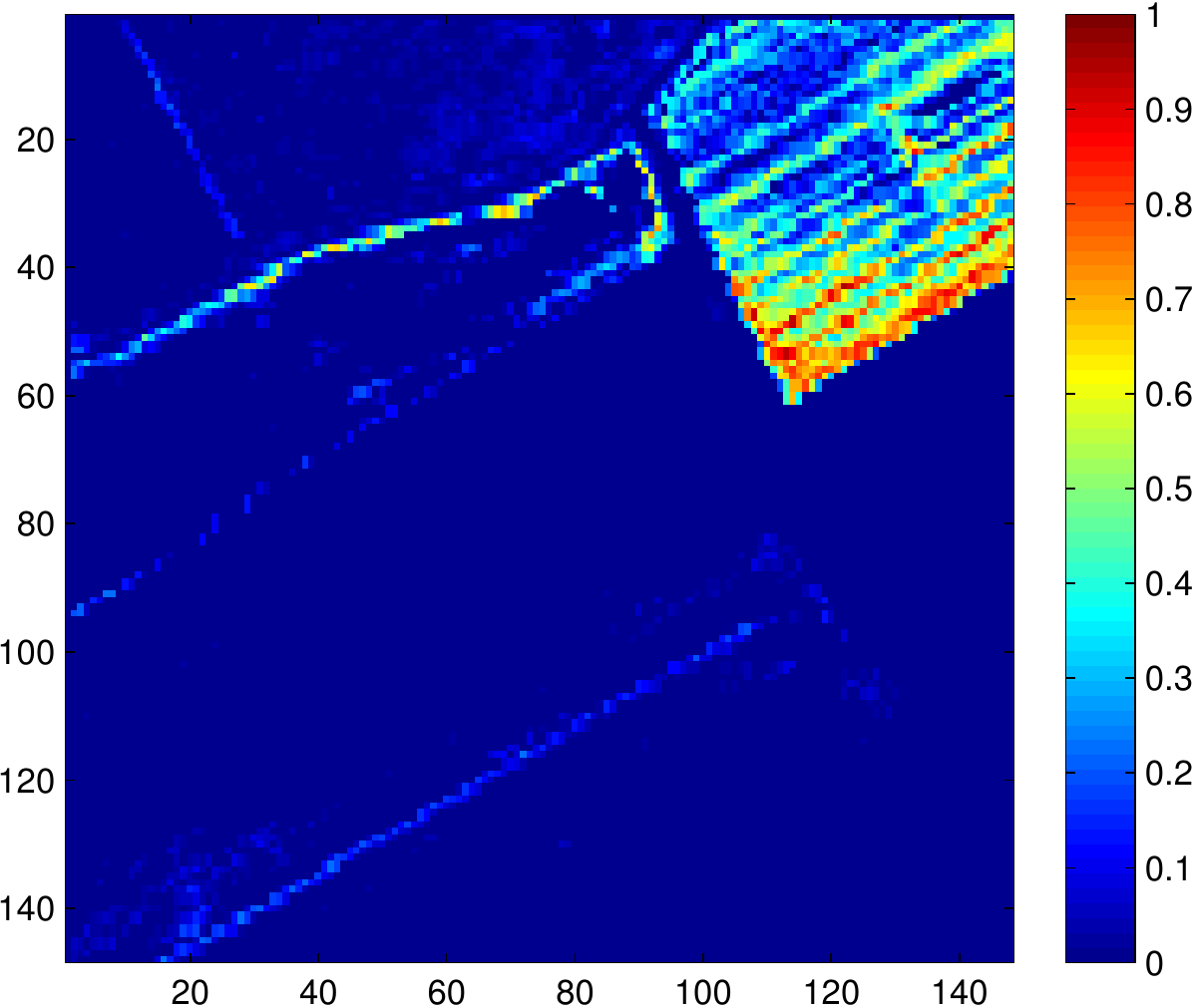}}
\end{subfloat}&
\begin{subfloat}
\centering
{\includegraphics[width=0.2\linewidth]{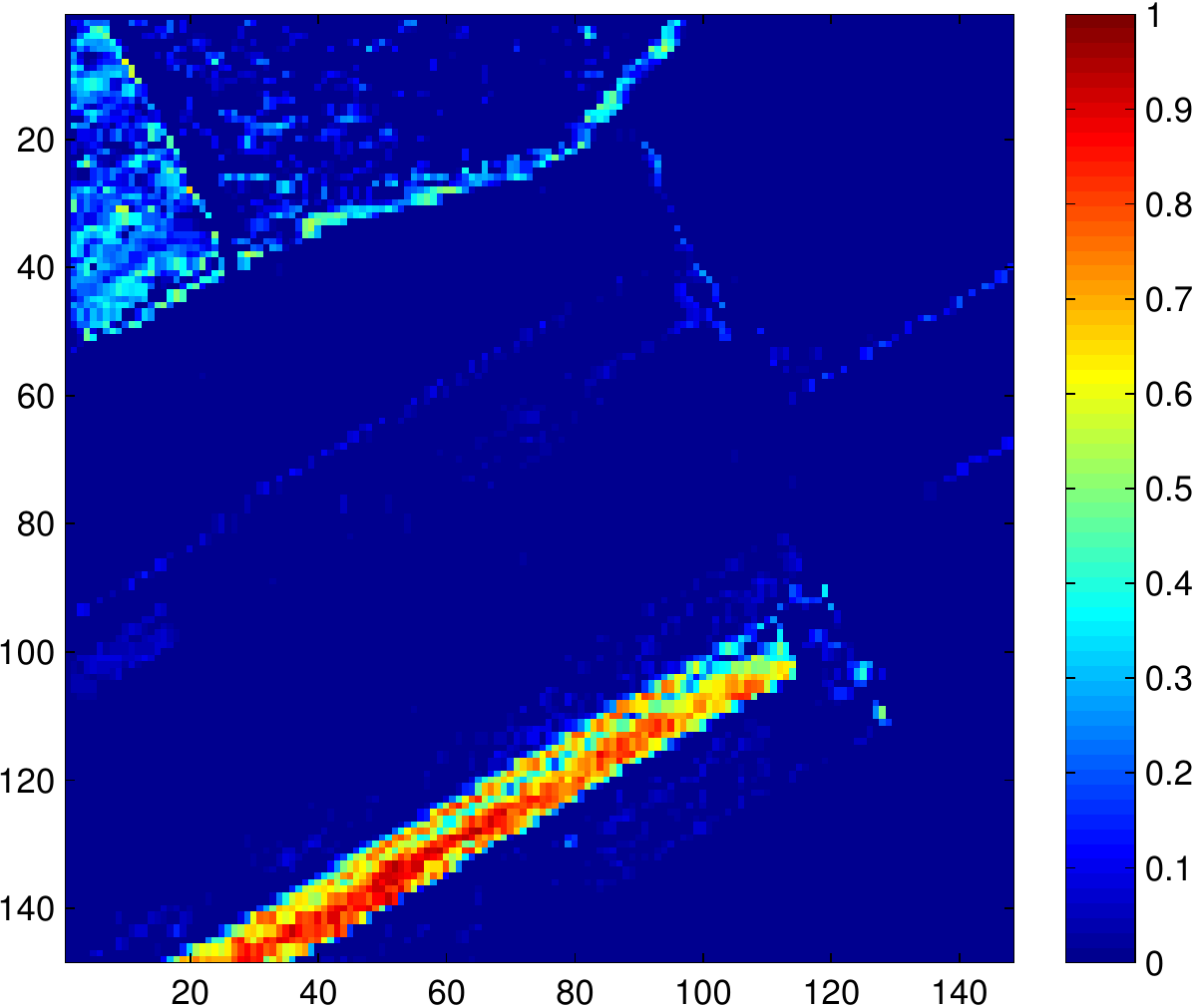}}
\end{subfloat} 
\end{tabular}
\vspace*{-0.3cm}
\caption*{(c) CSUnSAL}
\begin{tabular}{c c c c}
\centering
 \begin{subfloat}
\centering
{\includegraphics[width=0.2\linewidth]{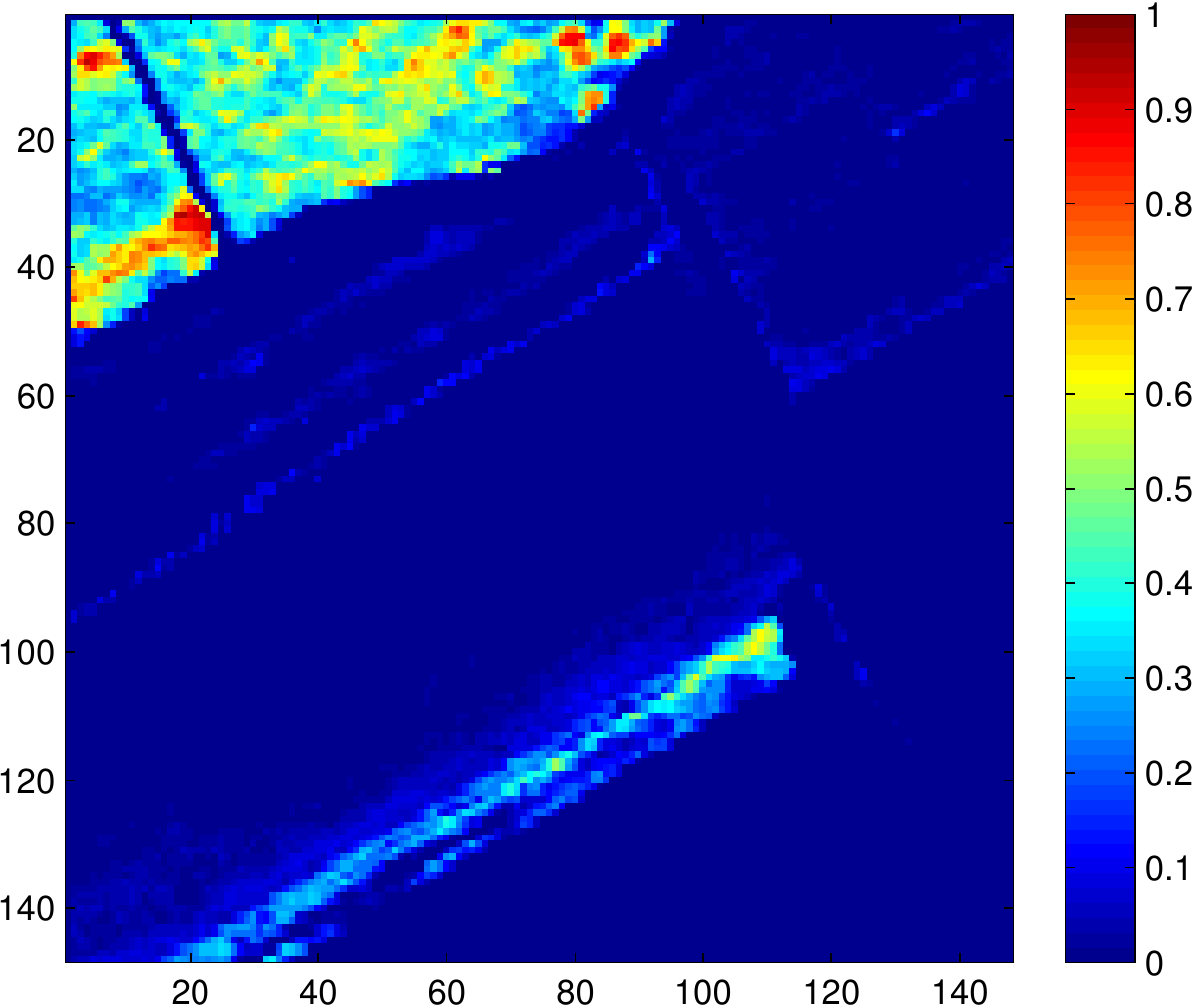}}
\end{subfloat}&
\begin{subfloat}
\centering
{\includegraphics[width=0.2\linewidth]{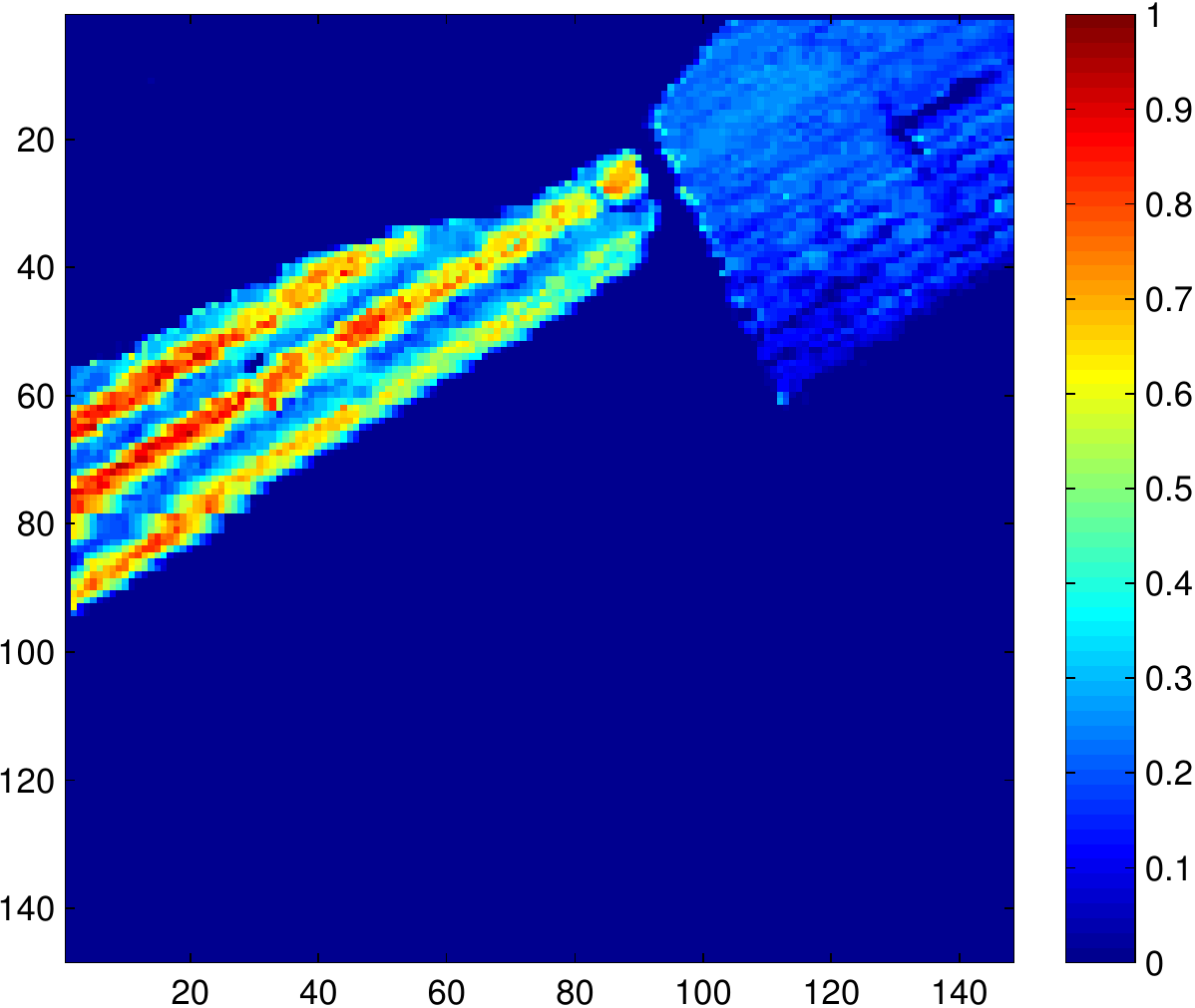}}
\end{subfloat}&
\begin{subfloat}
\centering
{\includegraphics[width=0.2\linewidth]{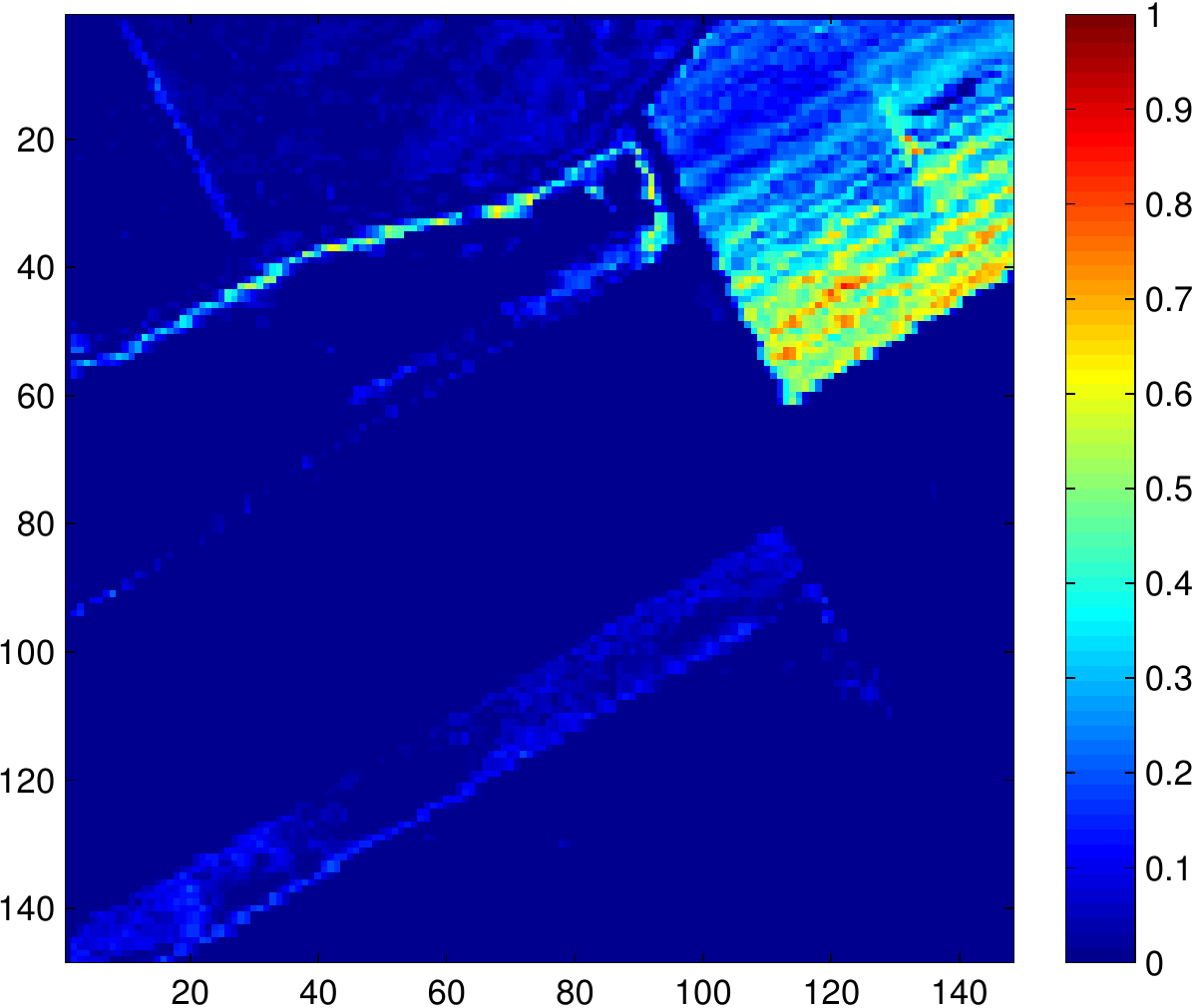}}
\end{subfloat}&
\begin{subfloat}
\centering
{\includegraphics[width=0.2\linewidth]{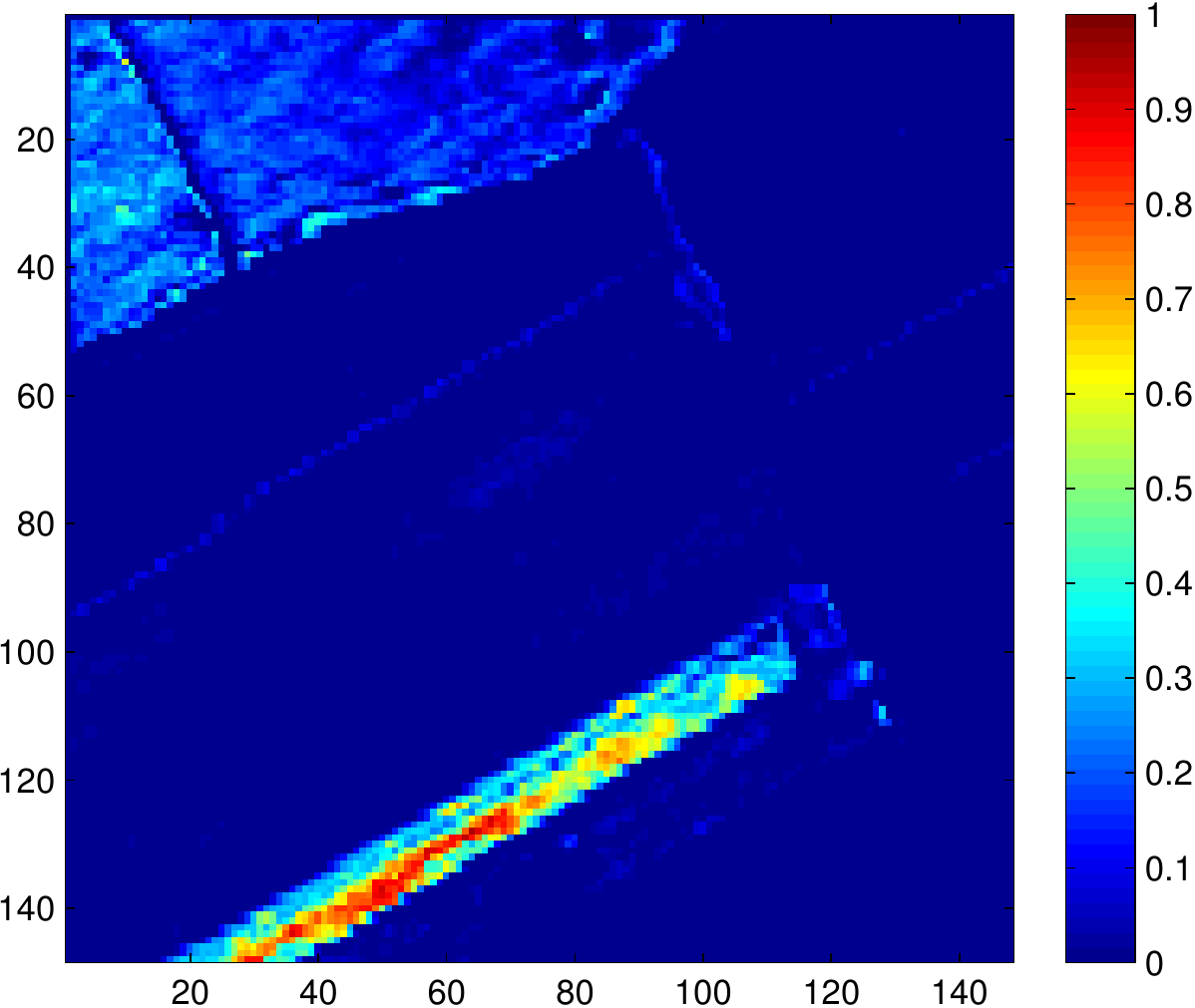}}
\end{subfloat} 
\end{tabular}
\vspace*{-0.3cm}
\caption*{(d) MMV-ADMM}	
\begin{tabular}{c c c c}
\centering
 \begin{subfloat}
\centering
{\includegraphics[width=0.2\linewidth]{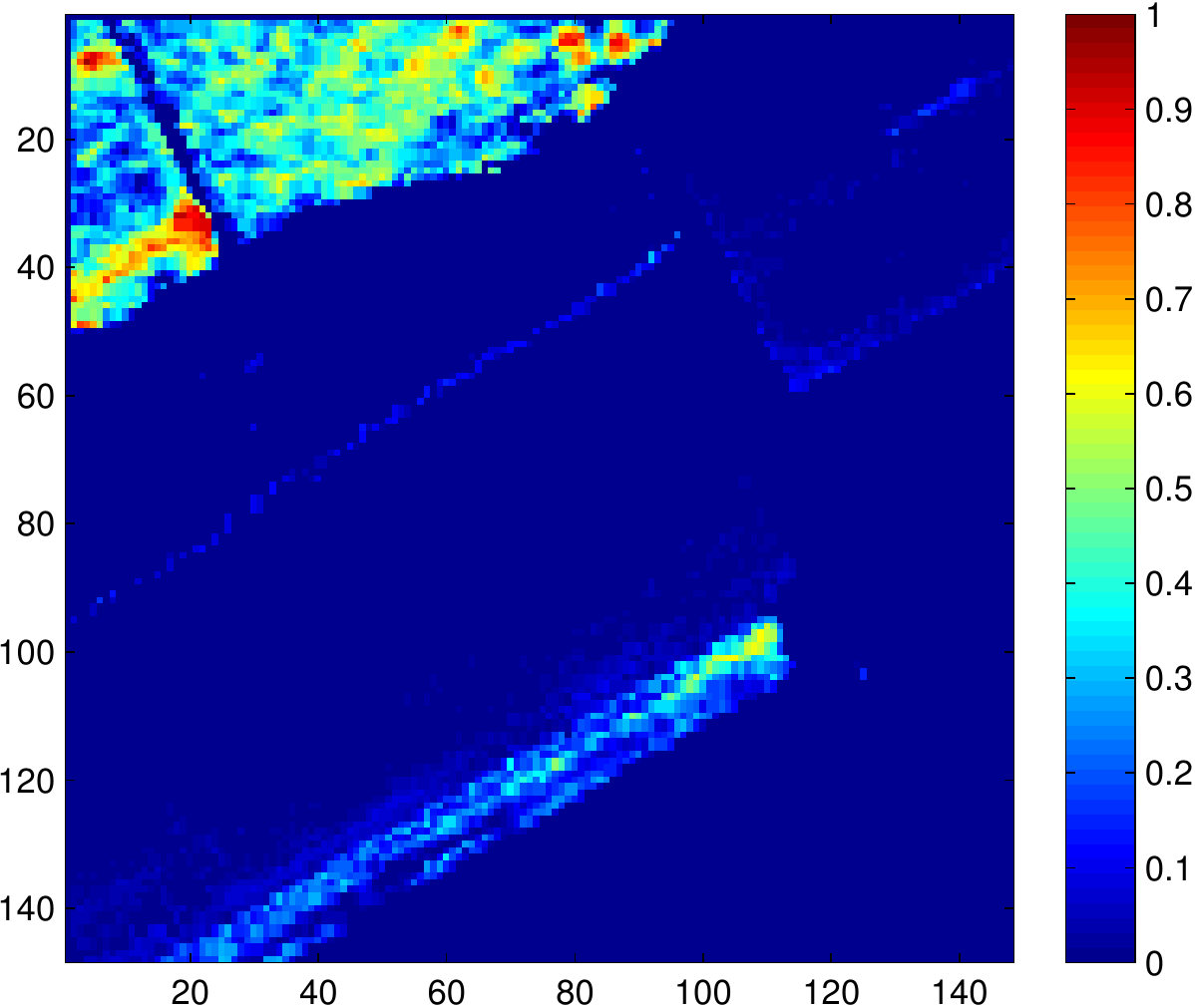}}
\end{subfloat}&
\hspace{0cm}\begin{subfloat}
\centering
{\includegraphics[width=0.2\linewidth]{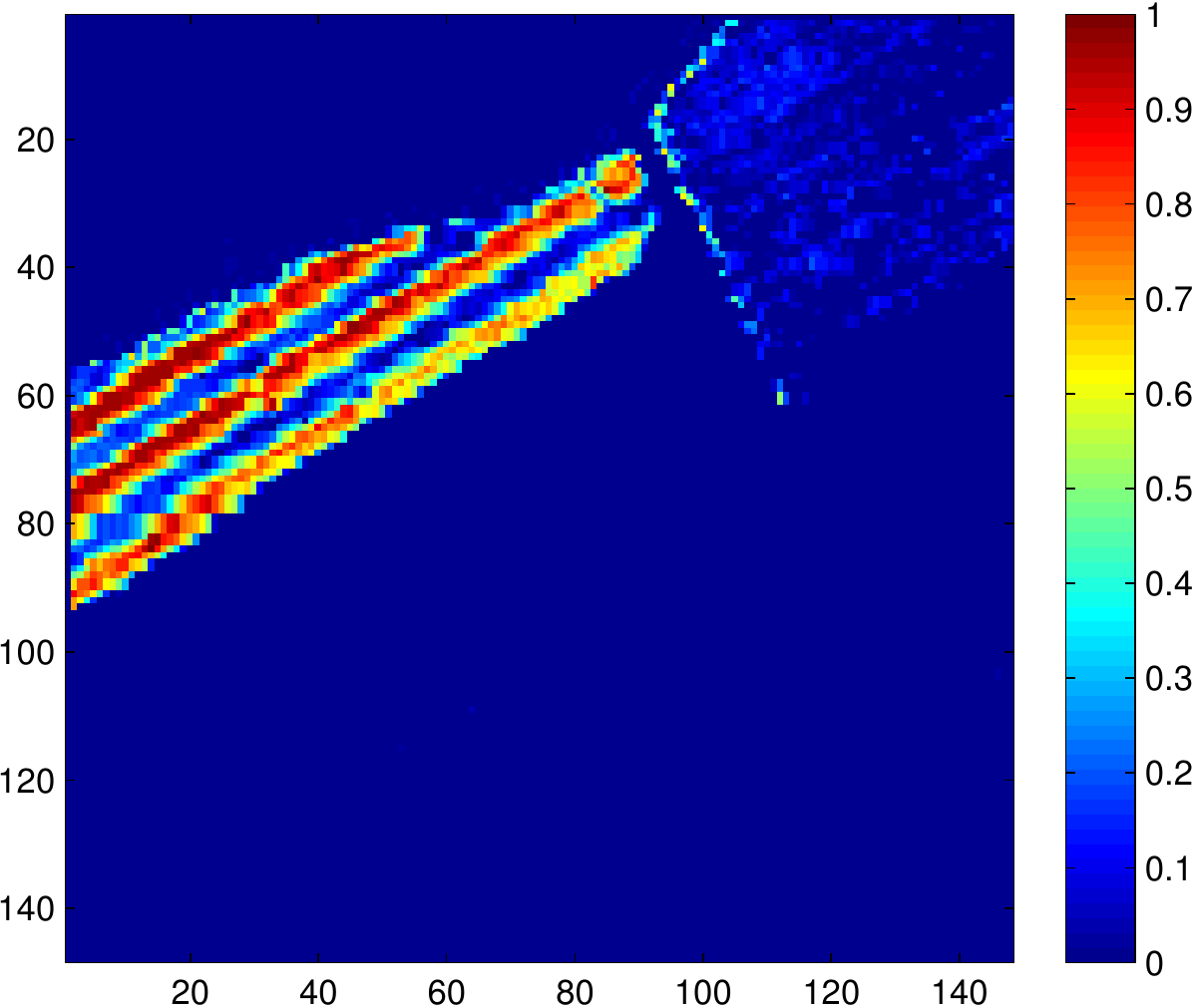}}
\end{subfloat}&
\hspace{0cm}\begin{subfloat}
\centering
{\includegraphics[width=0.2\linewidth]{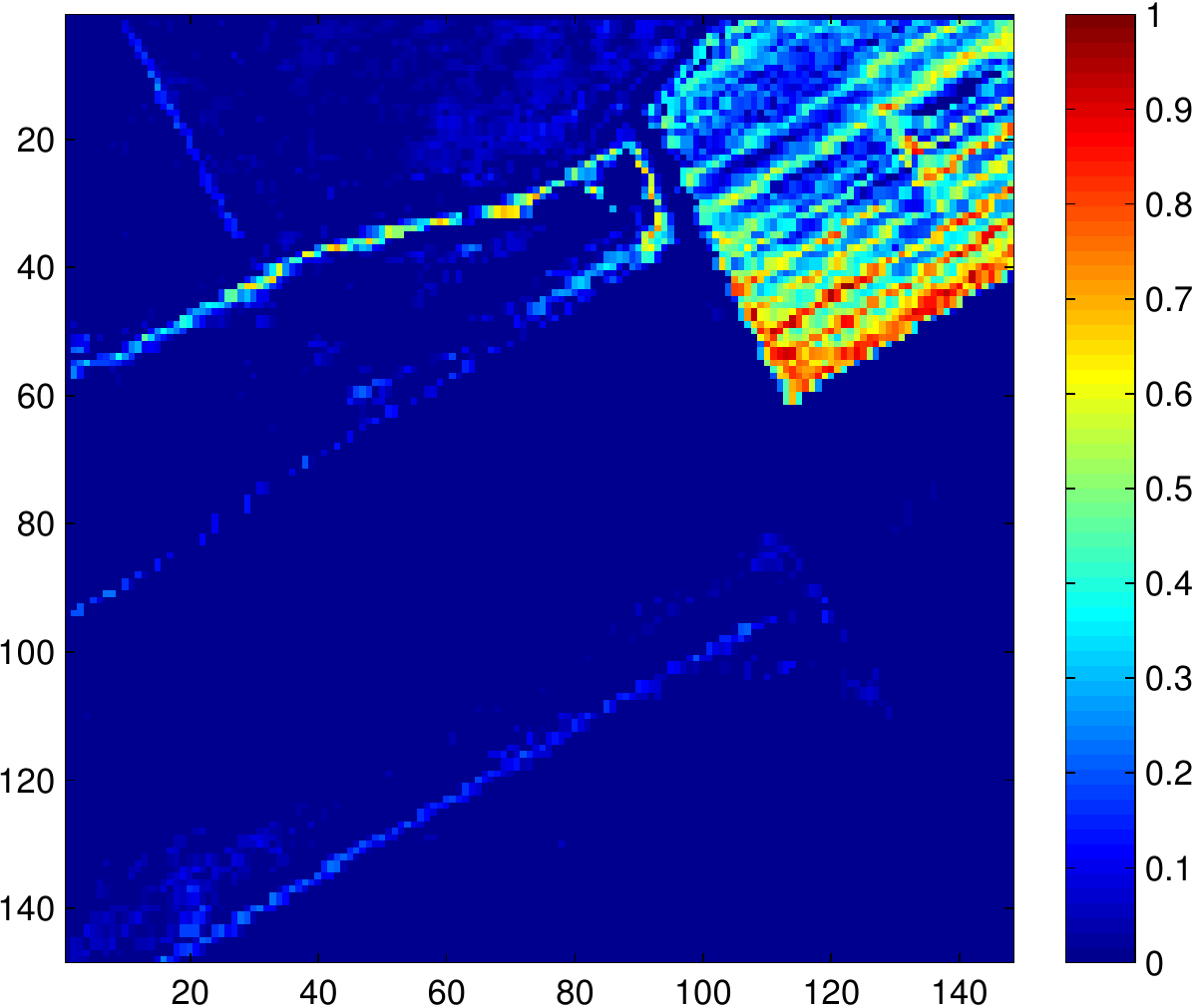}}
\end{subfloat}&
\hspace{0cm}\begin{subfloat}
\centering
{\includegraphics[width=0.2\linewidth]{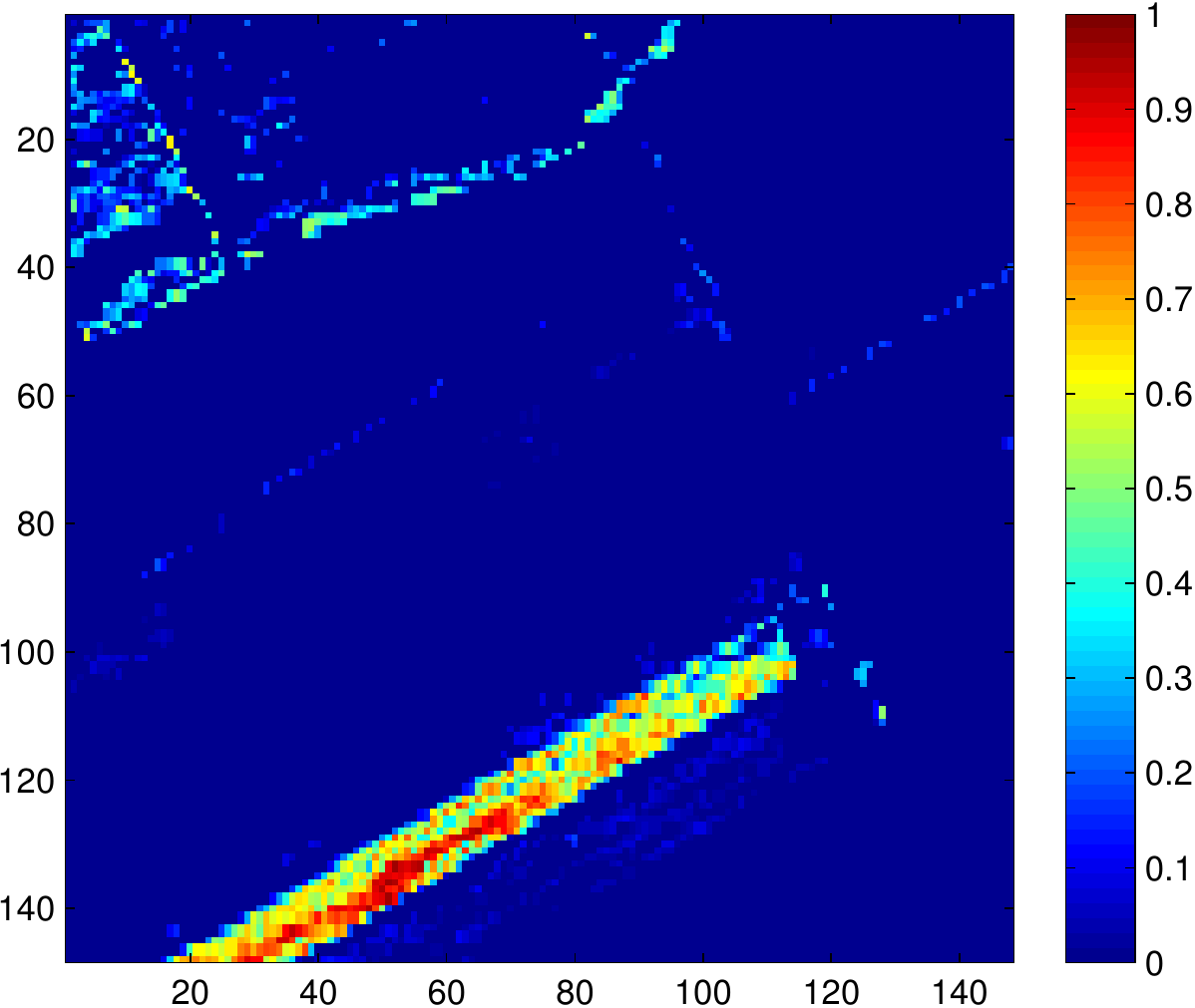}}
\end{subfloat} 
\end{tabular}
\vspace*{-0.3cm}
\caption*{(e) BiICE}	
\end{center}
\caption{Abundance maps of Salinas hyperspectral image.}
\label{abund_maps}
\end{figure*}
\subsection{Experiment on Real Data}\label{expd}
This section illustrates the performance of the proposed algorithms when applied on a real hyperspectral image. The hyperspectral scene under examination is a portion of the widely used Salinas vegetation scene acquired by AVIRIS sensor over Salinas Valley in California. This scene contains eight different vegetation species, namely grapes, brocolli\_A, brocolli\_B, lettuce\_a, lettuce\_b, lettuce\_c, lettuce\_d, corn, as shown in Fig. \ref{salinasgroundtruth}. Salinas hyperspectral image consists of $L = 204$ spectral bands and its spatial resolution is 3.7 meters. Taking the principal components (PCs) of the image, it can be seen that only the first 6 of them contain meaningful information. Focusing on them, we can see that the first PCs give more rough information about the formation of the vegetation, while less significant PCs give more refined information about the vegetation formation, \cite{theodoridis2008pattern}. Fig. \ref{salinas5pc} shows the $5$th principal component (PC) of the scene under study, where most of the vegetation is depicted. The endmembers dictionary $\boldsymbol\Phi$ is composed of 17 pure pixels' spectral signatures, which have been selected manually, as in \cite{emylona} and depicted in Fig. \ref{dictionary}. 



Fig. \ref{abund_maps} shows the abundance maps corresponding to the region of interest, as obtained by the proposed IPSpLRU, ADSpLRU and the three state-of-the-art competing algorithms namely CSUnSAL and MMV-ADMM and BiICE for $\gamma = 10^{-3}$, $\tau = 10^{-4}$, $\lambda=0.5$ and $\mu = 10^{-2}$. Four different maps are depicted, corresponding to four vegetation species, namely: grapes, brocolli\_a, brocolli\_b and corn. It is worth pointing out that since detailed ground truth information is not available, the evaluation is done in qualitative terms. From a careful visual inspection of the generated maps, we can see that the abundances obtained by IPSpLRU and ADSpLRU present patterns which are closer to those revealed by the first five principal components of the hyperspectral image provided in \cite{emylona}. This is particularly clear for the maps corresponding to brocolli\_a and brocolli\_b. More specifically, it is shown that the presence of these two species, which is mainly located in two distinct regions, is better emphasized by the proposed algorithms. In addition, the erroneous detection of these vegetation types is eliminated more effectively by IPSpLRU and ADSpLRU, as also verified by Figs \ref{salinas5pc} and \ref{abund_maps}. Hence, it is corroborated that the exploitation of the inherent spatial correlation existing in hyperspectral images, can lead us to qualitatively better results, thus verifying the significance of our approach. 

\section{Conclusions and Future Directions}
In this paper we presented a novel approach for performing hyperspectral image unmixing exploiting simultaneously sparsity and spatial correlation. A novel cost function was first introduced comprising a least squares proximity component regularized by a linear combination of the weighted $\ell_1$ norm and the weighted nuclear norm of the latent abundance matrix. The unmixing problem was thus treated as a sparse reduced-rank regression problem. Two different algorithms were then developed for solving it, namely an incremental proximal type algorithm called IPSpLRU, and an ADMM based strategy called ADSpLRU. Extensive simulations on both synthetic and real data corroborate the effectiveness of the proposed approach and algorithms, compared to other related state-of-the-art unmixing schemes. The derivation of more computationally efficient schemes alleviating the need for SVD is under current investigation. Another relevant future research direction is the exploitation of the specific structure or pattern of sparsity in the abundance matrices imposed implicitly by the low-rankness property, which could further improve estimation performance. This is also a topic of interest in the framework of a future work. 
\bibliographystyle{IEEEtran}
\bibliography{IEEEabrv,refs_report}
\end{document}